%% file: spark.tex
\documentclass[conference]{IEEEtran}
\usepackage{times}

\usepackage{times}
\renewcommand{\thetable}{\arabic{table}}
\usepackage{subcaption}
\usepackage{amsmath}
\usepackage{amssymb}
\usepackage{amsthm}
\usepackage{array}
\newcolumntype{C}[1]{>{\centering\arraybackslash}p{#1}}

\usepackage[numbers]{natbib}
\usepackage{multicol}
\usepackage[bookmarks=true]{hyperref}
\usepackage{hyperref}
\usepackage{cleveref}
\usepackage{booktabs}
\usepackage{graphicx} 
\usepackage{tabularx}
\usepackage{array}
\newcolumntype{Y}{>{\centering\arraybackslash}X}
\usepackage{array}
\usepackage{multirow}
\usepackage{makecell}
\let\labelindent\relax 
\usepackage{enumitem}
\usepackage{listings}
\usepackage{xcolor}
\usepackage{xspace}
\usepackage{tikz}
\usepackage[english]{babel}
\usepackage[utf8]{inputenc}
\usepackage{algorithm}
\usepackage[noend]{algpseudocode}
\usepackage{mathtools}
\usepackage{authblk}
\usepackage{bm}
\usepackage{wrapfig}
\usepackage{tikz}
\usepackage{pgfplots}
\pgfplotsset{compat=newest}
\usepackage[font={footnotesize}]{caption}
\usepackage{verbatim}
\usepackage{flushend}
\usepackage{rotating} 
\usepackage{listings}
\input{math_macro}

\newcommand{\spark}{\texttt{SPARK}\xspace}
\usepackage{pifont}

\begin{document}

\title{\spark: Safe Protective and Assistive Robot Kit}



\author{
    Yifan Sun$^{1}$, Rui Chen$^{1}$, Kai S. Yun$^{1}$, Yikuan Fang$^{1}$, Sebin Jung$^{1}$\\
    \vspace{0.3cm}
    Feihan Li$^{1}$, Bowei Li$^{1}$, Weiye Zhao$^{1}$, and Changliu Liu$^{1}$ \\
    \vspace{0.3cm}
    \textit{$^{}$Robotics Institute, Carnegie Mellon University} \\
    \vspace{0.1cm}
    \textit{\{yifansu2, ruic3, sirkhooy, yikuanf, sebinj, feihanl, boweili, weiyezha, cliu6\}@andrew.cmu.edu} \\
    \vspace{0.2cm}
    \href{https://intelligent-control-lab.github.io/spark/}{\texttt{https://intelligent-control-lab.github.io/spark/}}\\
    {*A short version of this paper has been presented at IFAC Symposium on Robotics in 2025.}
}

\twocolumn[{%
\renewcommand\twocolumn[1][]{#1}%
\maketitle
\begin{center}
    \centering
    \captionsetup{type=figure}
    \vspace{4cm}
    \begin{tikzpicture}[transform canvas={xshift=0cm}]
        \def\imgwidth{4.5cm}  
        \def\xgap{0}  

        \node at (-\xgap, 0) 
        {\includegraphics[width=1.0\linewidth]{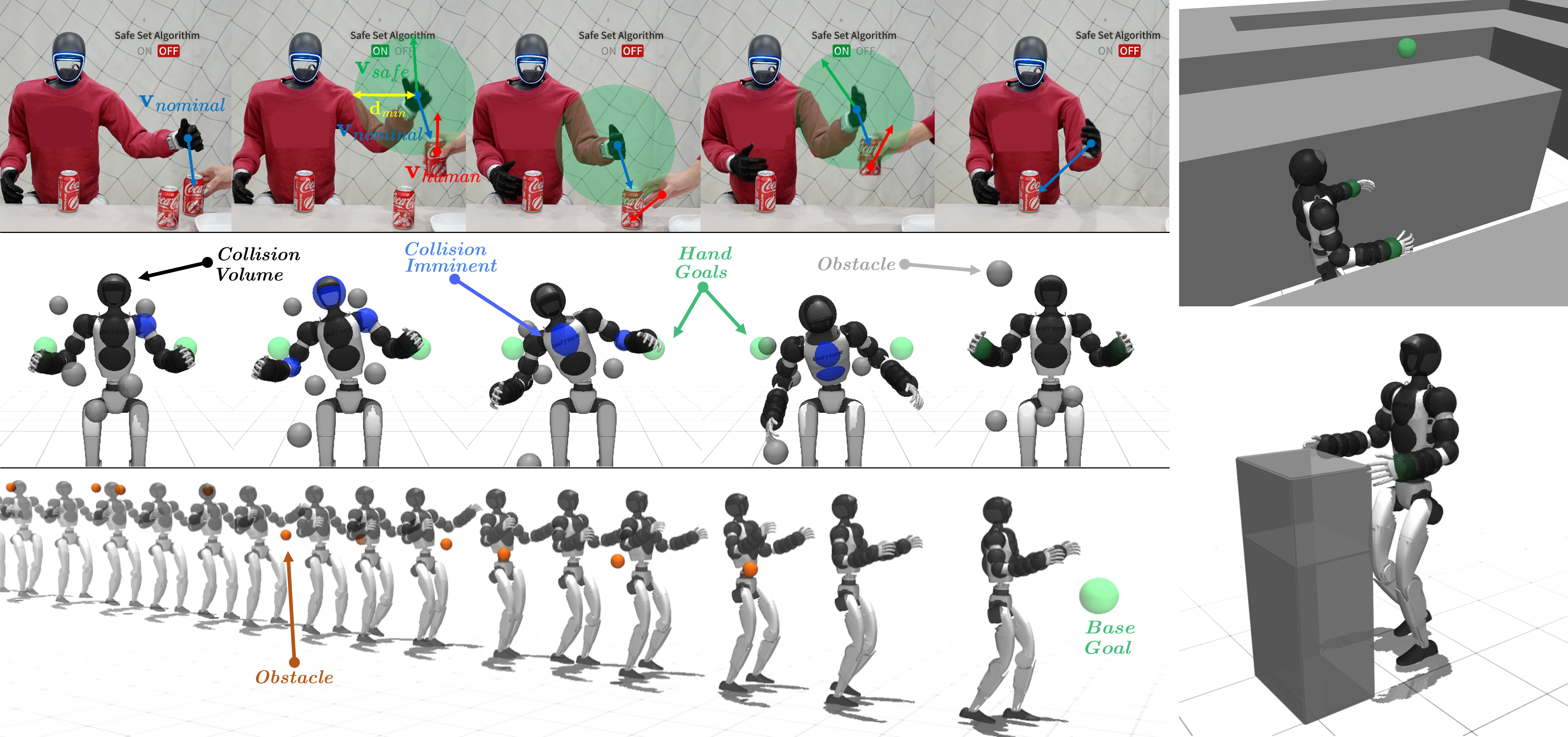}};
    \end{tikzpicture}
    \vspace{4.0cm}
    \captionof{figure}{Scenarios of safe humanoid control achieved with \textbf{\spark}. \textbf{Left top figure:} A real Unitree G1 humanoid robot avoiding human hands to ensure safe human-robot interaction. \textbf{Left middle figure:} A fixed-base humanoid robot safely reaching a target hand position in a crowded environment. \textbf{Left bottom figure:} A simulated humanoid robot navigating to a target position while avoiding obstacles. \textbf{Right top figure:} An example of a user-defined scenario for safe robot navigation in a maze. \textbf{Right bottom figure:} A daily-life scenario where a humanoid robot, under teleoperation, picks up objects from a cabinet in simulation. }
    \label{fig: title_figure}
\end{center}%
}]

\input{sections/0_abstract}

\IEEEpeerreviewmaketitle

\input{sections/1_introduction_v3_1}
\input{sections/2_related_v1}
\input{sections/3_system_v2}
\input{sections/4_test_suite_v1}

\input{sections/5_usecase_benchmark}
\input{sections/8_usecase_simulation_teleoperation}
\input{sections/6_usecase_hardware_autonomy}    
\input{sections/7_usecase_hardware_teleoperation}         
\input{sections/9_conclusion}
\input{sections/10_acknowledgement}







\input{spark.bbl}
\input{sections/appendix}

\end{document}

%% file: math_macro.tex
\newcommand{\ub}{\mathbf{u}}

\newcommand{\xb}{\mathbf{x}}

\newcommand{\Xb}{\mathbf{X}}

\newcommand{\cX}{\mathcal{X}}

\newcommand{\RR}{\mathbb{R}}

\newcommand{\minimize}{\mathop{\mathbf{min}}}

\newcommand{\minimizewrt}[1]{\mathop{\underset{#1}{\minimize}}}

\newcommand{\defeq}{\vcentcolon=}

\newcommand{\st}{\mathop{\mathbf{s.t.}}}

\newtheorem{problem}{Problem}

\newtheorem{remark}{Remark}[problem]
\newtheorem*{remark*}{Remark}

\colorlet{revision_color}{black}

%% file: sections/0_abstract.tex
\begin{abstract}
    This paper introduces the Safe Protective and Assistive Robot Kit (\spark), a comprehensive benchmark designed to ensure safety in humanoid autonomy and teleoperation.
    Humanoid robots pose significant safety risks due to their physical capabilities of interacting with complex environments.
    The physical structures of humanoid robots further add complexity to the design of general safety solutions.
    To facilitate safe deployment of complex robot systems, \spark can be used as a toolbox that comes with state-of-the-art safe control algorithms in a modular and composable robot control framework.
    Users can easily configure safety criteria and sensitivity levels to optimize the balance between safety and performance.
    To accelerate humanoid safety research and development, \spark provides simulation benchmarks that compare safety approaches in a variety of environments, tasks, and robot models.
    Furthermore, \spark allows quick deployment of synthesized safe controllers on real robots.
    For hardware deployment, \spark supports Apple Vision Pro (AVP) or a Motion Capture System as external sensors, while offering interfaces for seamless integration with alternative hardware setups at the same time.
    This paper demonstrates \spark's capability with both simulation experiments and case studies with a Unitree G1 humanoid robot.
    Leveraging these advantages of \spark, users and researchers can significantly improve the safety of their humanoid systems as well as accelerate relevant research.
    The open source code is available at \href{https://github.com/intelligent-control-lab/spark}{{https://github.com/intelligent-control-lab/spark}}. 
\end{abstract}

%% file: sections/1_introduction_v3_1.tex
\section{Introduction}
\label{sec: introduction}

Safety is an essential element for robotic systems not only in industrial settings but also in everyday life.
From an algorithmic perspective, safe control ensures a system operates within user-specified safety constraints to minimize risks and harm while achieving objectives. 
Recognizing the vital importance of maintaining these constraints, numerous safe control approaches have been developed and successfully applied to a wide range of robotic systems.

Among these approaches, the general pathway is to formulate the safe control problem as a constrained optimization problem, where the objective is to accomplish a given task while satisfying the safety constraints imposed on the robot.
Analytical methods can transform this optimization problem into a quadratic programming (QP) formulation, which can then be solved efficiently by existing solvers. Alternatively, data-driven methods leverage black-box policies to generate control solutions directly.
However, regardless of the success of the approaches, the fundamental challenge remains: \textit{how to allow non-experts to efficiently synthesize and deploy safe controllers in diverse applications?} The efficient synthesis and deployment require easy specification of both the objective and constraints that ensure safety while fulfilling task requirements, informed selection of the best safe control approaches, and tuning-free deployment on the real hardware. 

As model-free data-driven approaches have not been successful in safety assurance in high-dimensional systems, and the formal verification of these methods still requires the system model~\cite{fulton2018safe}, 
this paper focuses on efficient synthesis and deployment of model-based approaches. 

Synthesizing a model-based safe controller for simple scenarios is relatively straightforward.
Consider a differential-drive 2D autonomous vehicle tracking a trajectory with an upperbound velocity constraint for safety. 
It is not difficult to design an appropriate energy-like function that ensures the vehicle does not go over this speed limit.
However, as the scenario grows more complex (e.g., more complex constraints and objectives,  more complex robot dynamics, or more complex environments), synthesizing model-based safe controllers for each case becomes increasingly challenging and requires significantly more time and resources.

Now, consider replacing the differential-drive vehicle with a humanoid robot delivering a package with the same upperbound velocity constraint.
Due to the different dynamics of a humanoid robot, such as its joint limits, the safe controller previously synthesized for the vehicle may no longer be valid.
Furthermore, once the humanoid robot successfully arrives at the package dropoff destination, e.g., a loading truck, the task transitions from navigation to upper body manipulation, requiring the robot to load the box onto the truck while avoiding collisions.
The humanoid dynamic can no longer be treated as a mobile robot and the upper body needs to be incorporated into the safety constraint.
Even for the same loading task, changing the control mode from autonomous operation to teleoperation or language-vision-based commands introduces new challenges for safe control.

Finally, even for similar safety constraints, the safe controller may need re-synthesis when the environment changes. 
For instance, consider a warehouse where a humanoid robot operates alongside human workers who are loading boxes. In this setting, the robot must avoid both static obstacles, e.g., a parked truck, and dynamic human participants.
Safe controller synthesis in this scenario must also ensure robust human safety, potentially achieved by integrating a human motion prediction model into the control stack.

Hence, synthesizing a safe controller from scratch on a case-by-case basis can quickly become time-consuming and inefficient, as each change in \textit{robot}, \textit{task}, or \textit{environment} requires re-synthesis by experts. 
At the same time, designing a single, universal safe controller capable of handling every possible scenario - varying tasks, diverse robot dynamics, ever-changing environments, and arbitrary safety constraints - is highly challenging, if not impossible. 
This inherent inefficiency hinders the practical development and deployment of safe controllers across real-world robotic applications.

To lower the barrier and increase the efficiency of safe controller synthesis and deployment, a plug-and-play modular framework is essential. 
To this end, we present the \textbf{Safe Protective and Assistive Robot Kit (\spark)}, as a modular safe control framework. 
With \spark, users can efficiently design, verify, and benchmark safe controllers for complex robot systems without starting from scratch. 
This approach reduces the effort of synthesis while promoting scalability and generalization of safe control methods across diverse robot configurations and task scenarios.
Guided by these principles, we designed \spark to be
\\
\noindent\textbf{Composable:}
\begin{itemize}
    \item It provides a set of modular components that users can assemble to create safe robotic control scenarios with custom task goals and safety requirements as shown in \Cref{fig: title_figure}.
    \item It offers predefined module options to facilitate large-scale benchmark scenario generation for evaluating various safe control approaches.
    \item It enables users to switch between similar scenarios by simply replacing the corresponding module.
\end{itemize}
\noindent\textbf{Extensible:}
\begin{itemize}
    \item It allows users to customize each module, such as developing their own safe controllers or modifying robot dynamics, using provided templates.
    \item It supports the integration of additional user-defined modules, including external sensors and task planners.
    \item It provides a general wrapper to seamlessly bridge \spark with other existing benchmarks.
\end{itemize}
\noindent\textbf{Deployable:}
\begin{itemize}
    \item It enables users to deploy synthesized safe control algorithms on real robotic hardware by wrapping the hardware SDK (Software Development Kit) with the \spark interface.
    \item It ensures compatibility with real-time middleware such as ROS (Robot Operating System) and DDS (Data Distribution Service).
    \item It maintains robustness across diverse real-world task settings, including human-robot interaction and teleoperation through Apple Vision Pro.
\end{itemize}


Ultimately, \spark alleviates challenges in safe controller synthesis and benchmarking, empowering robots to operate safely and reliably in diverse real-world scenarios.

%% file: sections/2_related_v1.tex
\section{Related Work}  

In this section, we briefly survey previous work closely tied to our objective of enabling efficient, modular safe controller synthesis and benchmarking for robotic systems. We mainly focus on safe control for humanoids as humanoids are one of the most complex platforms in robotics.

\paragraph{Modular Safe Control Toolboxes}  
Safe control toolboxes aim to provide modular and general safe control algorithms that can be integrated into various robotic systems. The Benchmark of Interactive Safety (BIS)~\citep{wei2019safe} implements a set of energy-function-based safe control algorithms, such as the Potential Field Method (PFM)~\citep{khatib1986real}. Similarly, \citep{schoer2024control} introduces a toolbox focused specifically on Control Barrier Functions (CBF). However, these toolboxes do not offer the modularity required for seamless compatibility with other frameworks and lack tasks tailored for complex robotic systems.  

When it comes to complex robotic systems such as humanoids, specialized safe control methods have been proposed to address their high-dimensional dynamics. For example, CBF~\cite{ames2019control} is used in a case by case manner to address locomotion safety of bipedal robots~\citep{nguyen2015footstepCBF,peng2024real}, safe humanoid navigation~\citep{agrawal2017dcbfnavigation}, self-collision avoidance for humanoids~\citep{khazoom2022humanoid, huang2024whole}, and humanoid whole-body task space safety~\citep{paredes2024wholebodysafe}.
Besides, ARMOR~\citep{kim2024armor} employs a novel egocentric perception system to learn data-driven safe motion planning policies for humanoid robots.
Despite these advancements, no existing benchmark provides standardized tasks for comparing current methods or supports the development of new safe control approaches in a consistent and modular way.  

\paragraph{Benchmark Platforms for Safety-critical Tasks}  

As safe reinforcement learning (RL) has emerged as a powerful framework for learning safe control policies from data, such as Constrained Policy Optimization (CPO)~\citep{achiam2017cpo} and State-wise CPO (SCPO)~\citep{zhao2023state}, many of the existing benchmark platforms for safety-critical tasks are tailored for safe RL. For instance, GUARD~\citep{zhao2023guard} offers a comprehensive set of environments featuring safe RL tasks alongside state-of-the-art safe RL algorithms. Similarly, Safety-Gymnasium~\citep{ji2023safety} integrates a wide range of safe RL environments into a unified benchmark. Safe Control Gym~\citep{yuan2022safe} provides a collection of safe control environments to facilitate model-free safe control policy learning.  
However, these platforms do not cater to tasks that invovle high-dimensional robotic systems, such as humanoid robots. In addition, they place limited emphasis on model-based methods, which remain essential for addressing safe control challenges due to their analytical rigor, composability, and interpretability.  

\paragraph{Benchmark Platforms for Humanoid Robots}  
Humanoid robots are among the most complex robotic systems, posing significant challenges~\cite{gu2025humanoid} in control. Learning-based methods have shown impressive results in humanoid control, and various open-source humanoid benchmarks have been developed. HumanoidBench~\citep{sferrazza2024humanoidbench} provides a benchmark for humanoid whole-body tasks, while Humanoid-Gym~\cite{gu2024humanoid} enables zero-shot sim-to-real humanoid locomotion training.  Mimicking-Bench~\cite{liu2024mimicking} establishes a benchmark for humanoid-scene interaction learning using large-scale human animation data, while MS-HAB~\cite{shukla2024maniskill} accelerates in-home manipulation research with a GPU-optimized benchmark and scalable demonstration filtering. BiGym~\cite{chernyadev2024bigym} is a benchmark for bi-manual robotic manipulation with diverse tasks, human demonstrations, and multi-modal observations. Based on these benchmarks, researchers can evaluate state-of-the-art algorithms on a variety of tasks such as humanoid teleoperation and whole-body control. H2O~\cite{he2024learning} and OmniH2O~\cite{he2024omnih2o} introduced reinforcement learning-based whole-body humanoid teleoperation systems, with H2O using an RGB camera and OmniH2O extending to multimodal control and autonomy. HumanPlus~\cite{fu2024humanplus} enables humanoids to learn motion and autonomous skills from human data via a combination of RL and behavior cloning. ExBody~\cite{cheng2024expressive} and ExBody2~\cite{ji2024exbody2} develop RL-based whole-body controllers, where ExBody focuses on generating expressive motions, and ExBody2 enhances generalization and fidelity using a privileged teacher policy. In addition,~\cite{zhuang2024humanoid} and~\cite{zhang2024wococo} demonstrate pure RL approaches for humanoid locomotion. HOVER~\cite{he2024hover} unifies a wide range of humanoid control tasks through multi-mode policy distillation, enabling seamless transitions across modes without the need for retraining.

Nevertheless, none of these benchmarks specifically focus on testing the safety of humanoid robots. Achieving safety for humanoids requires not only model-free~\cite{zhao2021model,yang2023model} approaches but also model-based methods~\cite{berkenkamp2017safe,yun2024safe}, which offer valuable analytical insights and interpretability in addition to their control capabilities.  

%% file: sections/3_system_v2.tex
\section{\spark's Framework for Testing, Benchmarking, Development and Deployment}

In this section, we begin by summarizing the key components involved in synthesizing a robotic control system, along with the dependencies among them throughout the synthesis process. We then introduce the modular framework of \spark, which inherently follows this natural synthesis pipeline. Finally, we demonstrate how the structure of \spark aligns with the principles established in Section~\ref{sec: introduction}.

\subsection{Components of a Safety-Critical Robotic Scenario}
\label{sec: syetem_component}

Before introducing the \spark framework, we first review the natural process that users typically follow when synthesizing control systems for robots. In practical scenarios, the synthesis process often begins with defining the \textit{system state}—such as joint positions and velocities in the case of a robotic manipulator—followed by identifying the corresponding \textit{system measurements} obtained from joint encoders and external sensors. Besides, the user specifies \textit{system objectives} such as end-effector tracking and collision avoidance. Then, \textit{system dynamics} are modeled either analytically or through data-driven approaches to capture how control inputs affect the robot's behavior. The \textit{system controller} is finally synthesized using these components to generate control commands that achieve the task objectives while ensuring safety. Another example is for mobile robot navigation, the \textit{system state} includes the robot’s pose and surrounding obstacles, estimated from \textit{system measurements} such as LiDAR and odometry; \textit{system objectives} like goal-reaching and obstacle avoidance are then specified; and \textit{system dynamics} describe how the robot evolves under motion commands. The \textit{system controller} integrates all of this information to compute safe navigational actions. In the examples, the synthesis process follows a consistent dependency structure: the controller is grounded in dynamics, guided by objectives, and informed by measurements, which in turn depend on the underlying state. These synthesis patterns reveal the essential components and dependencies that system designers must consider, forming the key motivation behind the structured architecture of the \spark framework.

Below, we formally introduce our definitions of \textit{system state}, \textit{system dynamics}, \textit{system objectives}, \textit{system measurements}, and \textit{system controller} in the context of control system synthesis.
While some of these definitions echo conventional control theory, others have been adapted to more precisely capture the context of safe controller synthesis procedure.

\paragraph{\textbf{System State}} includes the robot's internal state, such as a humanoid's joint positions and locomotion velocity, as well as the external state, which encompasses information about obstacles and human participants. 
The \textit{system state} can be obtained from either a simulated environment or the real physical world.

\paragraph{\textbf{System Dynamics}} encompass models for both the robot's internal dynamics and the external dynamics that influence its operation. 
It is important to note that the \textit{system dynamics} defined here are user-defined models rather than the true dynamics of the real world. 
In practice, capturing the complete complexity of real-world dynamics is infeasible.
These models, whether formulated analytically or derived from data, approximate the real world and are tailored for use by the \textit{system controller}, though they may differ from the actual dynamics.

\paragraph{\textbf{System Objectives}} encompass both task-specific targets, such as following a trajectory or reaching a target point, and safety constraints, such as ensuring obstacle avoidance. 
These \textit{system objectives} can be derived from pre-coded autonomy programs and user-defined safety criteria. 
Additionally, they may originate from user inputs, such as teleoperator gestures, or be generated by a Large Language Model (LLM)-based task planner.

\paragraph{\textbf{System Measurements}} provide the information about the system state to the controller.
By interfacing with external sensors, \textit{system measurements} capture details such as obstacle shape, number, and location, along with the goal positions.

\paragraph{\textbf{System Controller}} encompasses robot algorithms that optimize the control inputs for the robot based on the \textit{system dynamics}, \textit{system objectives}, and the \textit{system measurements}. 
It ensures that the robot operates efficiently while maintaining safety and achieving task-specific goals.

\begin{figure}[htbp]
    \centering
    \vspace{2cm}  
    \begin{tikzpicture}[transform canvas={xshift=0cm}] 
        \def\imgwidth{8cm}  

        \node at (0, 0) {\includegraphics[width=\imgwidth]{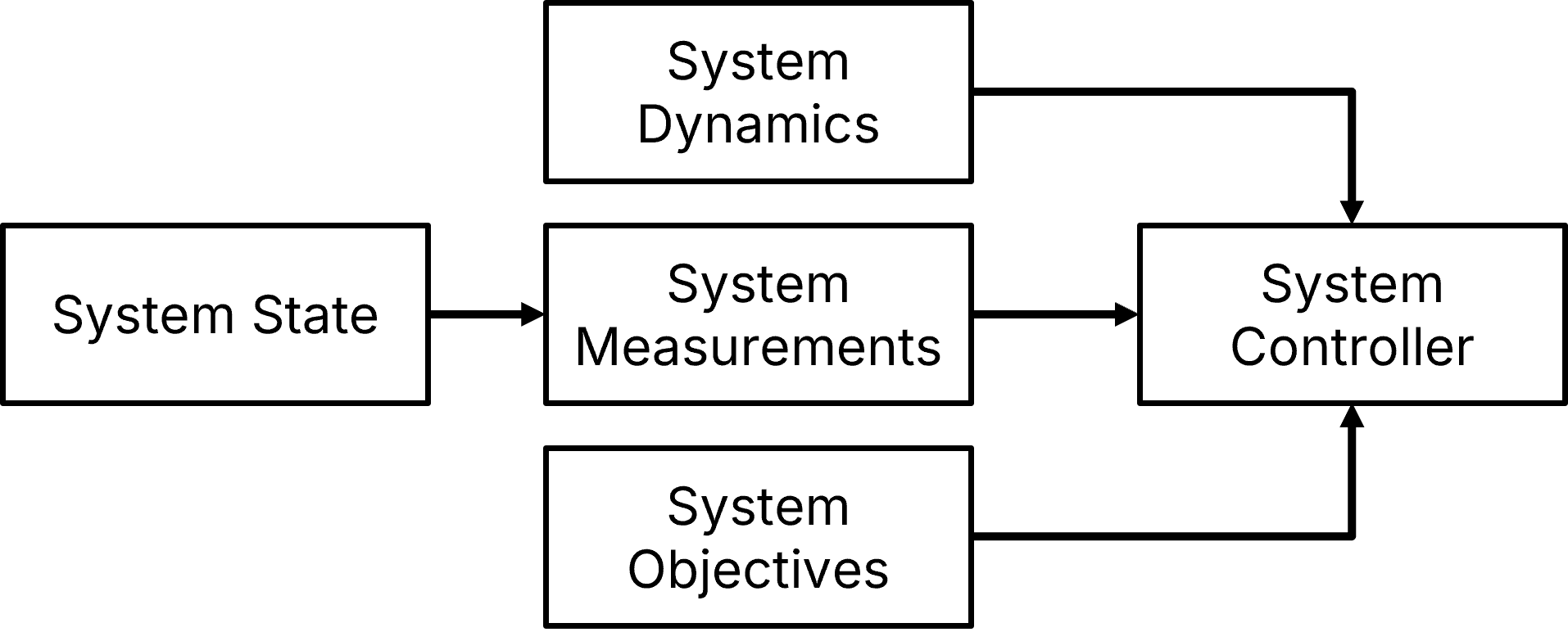}};
    \end{tikzpicture}
    \vspace{2cm}
    \caption{System components of a safety-critical robotic scenario and their interdependencies. $A\rightarrow B$ translates to ``$B$ depends on $A$''.}
    \label{fig: system_components}
\end{figure}

Synthesizing safe controllers for specific robot scenarios can be challenging due to the complex interdependencies among the system components as described above and shown in Figure~\ref{fig: system_components}. 
The \textit{system controller} heavily depends on the \textit{system measurements} to estimate the \textit{system state}, the \textit{system objectives} to inform its optimization process, and the \textit{system dynamics} to model the effect of its control actions.
\textit{System measurements} are intrinsically dependent on the \textit{system state}, aiming to approximate it through sensor data.
By contrast, the \textit{system state}, \textit{system objectives}, and \textit{system dynamics} are relatively self-contained and do not depend on other components.

Recognizing these dependencies motivates a structured approach to reducing modeling complexity, thereby enabling uniform and modular synthesis of safe controllers across diverse robotic scenarios. By explicitly capturing inter-component dependencies, the \spark framework promotes a transparent and extensible control synthesis process—an essential foundation for deploying reliable, safety-critical robotic systems.



\subsection{Framework of \spark}\label{sec: spark_framework} 

\begin{figure}[htbp]
    \centering
    \vspace{2.5cm}  
    \begin{tikzpicture}[transform canvas={xshift=0cm}] 
        \def\imgwidth{\linewidth}  

        \node at (0, 0) {\includegraphics[width=\imgwidth]{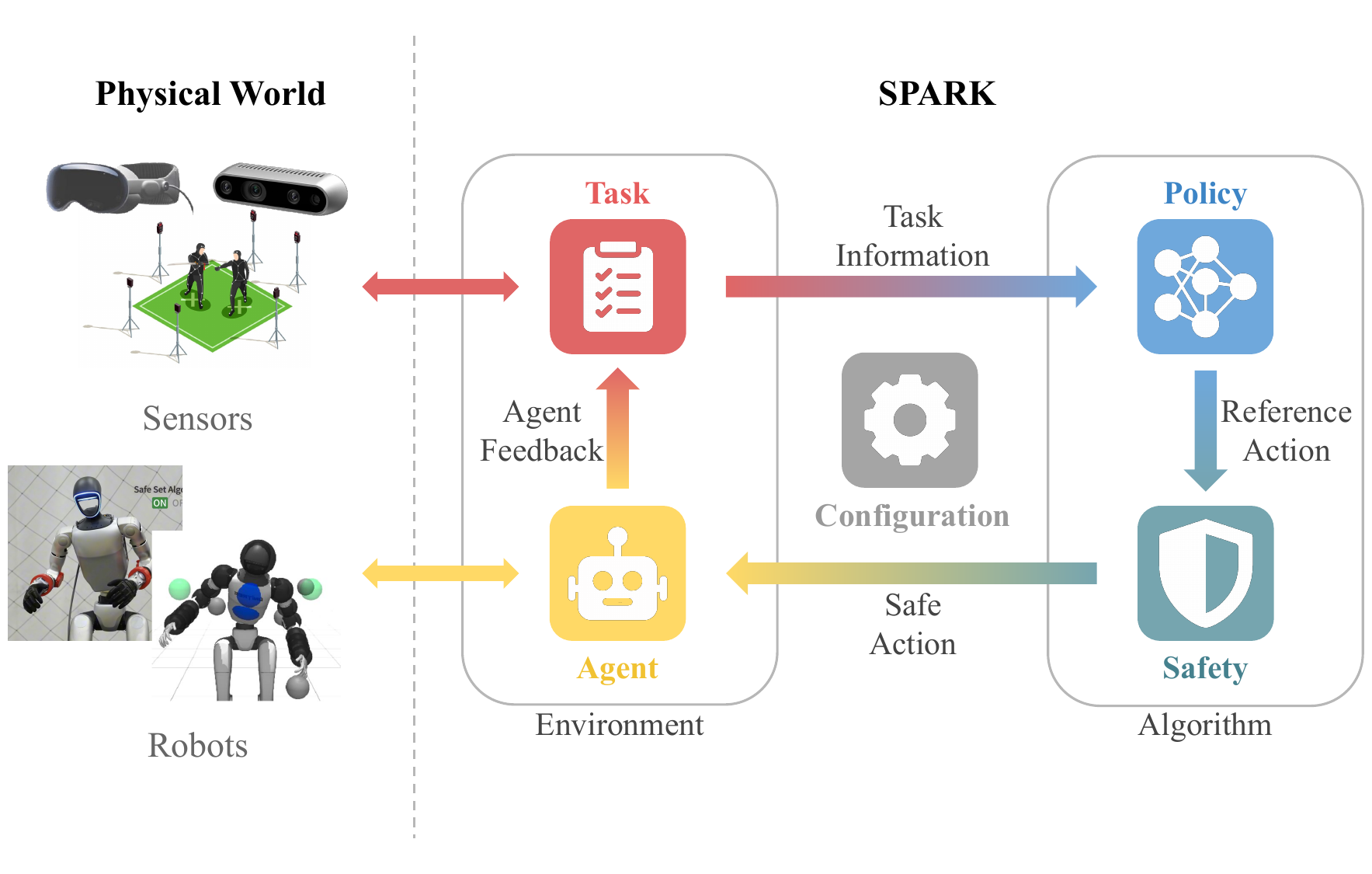}};
    \end{tikzpicture}
    \vspace{2cm}
    \caption{\spark system framework.}
    \label{fig:system_framework}
\end{figure}

To facilitate the decomposition and integration of the components discussed in \Cref{sec: syetem_component}, we present the \spark framework, as shown in Figure~\ref{fig:system_framework}, offering users a collection of modular Python class templates designed to streamline the synthesis, testing, benchmarking, development, and deployment of safe controllers for robotic systems. 

The \spark framework is structured around the decomposition of system components into distinct modules that allow efficient handling of system states, measurements, objectives, dynamics, and controllers. 
This modularity, inspired by the component relationships described in \Cref{sec:  syetem_component}, facilitates the synthesis of safe controllers that are composable, extensible, and deployable. The detailed software framework is described in \Cref{appendix: software_framework}, and Python examples are provided in \Cref{appendix: python_example} to facilitate a quick start.

We first introduce the \textbf{Configuration} module, which supports all other modules.

\paragraph{\textbf{Configuration Module}} houses \textit{system dynamics}, encapsulating robot-specific configurations such as degrees of freedom, motor interfaces, and system models. 
It provides essential context for other modules and enables the incorporation of robot-specific details, making it essential for defining and customizing dynamics within \spark. This module is typically implemented by modeling the physical robot system through system identification or model simplification.

\textbf{Environment}, composed of the \textbf{Agent} and \textbf{Task} modules, serves as the ``front-end'' of \spark. It is responsible for modeling the physical system by defining the relationship between the \textit{system measurements} and the \textit{system state}, as well as specifying the \textit{system objectives} to be achieved by the ``back-end'' modules. In a traditional control system, this process typically involves state estimation and observer design.

\paragraph{\textbf{Agent Module}} interacts with the physical robot or its simulation. 
It receives control commands from the system controller and manipulates the robot’s state by applying these commands to the robot’s actuators or simulated system. 
Consequently, the \textbf{Agent} module’s primary role is to influence and modify the robot’s \textit{system state}, reflecting its position in the overall framework as an executor of system control. 
\textbf{Agent} module houses the interfaces for both simulated and real robots, allowing \spark to be deployable for various tasks.

\paragraph{\textbf{Task Module}} is responsible for: (1) receiving the system state from the environment, such as current obstacle positions; (2) receiving the robot state from the \textbf{Agent} module, including joint positions and velocities; and (3) receiving user-specified task goals, such as teleoperation commands or planned trajectories. After collecting all relevant information for control, the \textbf{Task} module organizes it into a unified data structure and forwards it to the Algorithm module.

Traditionally, when synthesizing a controller, the specific inputs—whether system measurements or goal states—must be tailored case by case, making controllers difficult to generalize or reuse. In \spark, this information gathering and integration process is fully decoupled from the algorithm and is handled exclusively by the \textbf{Task} module, enabling modularity and ease of reuse.


We now turn to the ``back-end'' of the framework, which forms the core of the controller in a traditional control system synthesis pipeline. In \spark, where ensuring the safety of both the robot and its interaction with the environment is a priority, we adopt a modular and hierarchical structure for the back-end, referred to as the \textbf{Algorithm} module. This module consists of two submodules: \textbf{Policy} and \textbf{Safety}. The Policy module is responsible for generating the nominal control commands, while the Safety module modifies or overrides these commands to enforce safety constraints. This decoupled design allows users to easily test and integrate different safety algorithms without requiring detailed knowledge of the underlying nominal control policy.


\paragraph{\textbf{Policy Module}} processes the \textbf{Task} information to generate reference control actions that aim to achieve performance-oriented objectives, such as reaching a goal location, without considering safety constraints. 

\paragraph{\textbf{Safety Module}} refines the control actions given by the \textbf{Policy} module to ensure compliance with the safety constraints while attempting to follow the original reference control actions as closely as possible. 

Both \textbf{Policy} and \textbf{Safety} modules allow users to incorporate either model-based or data-driven controllers, preserving \spark's core principles of composability and extensibility.


\subsection{Why decomposing system components into the \spark framework is valuable}

As described in Section~\ref{sec: introduction}, the \spark framework is grounded in three core principles: \textit{composability}, \textit{extensibility}, and \textit{deployability}. By decomposing the system components into distinct, modular units, \spark maximizes flexibility and scalability for various robotic platforms and tasks. This design fosters the following advantages.

\begin{itemize}
    \item \textit{Composability:} Users can rapidly mix and match built-in or custom modules to create safe robotic control scenarios. 
    Predefined module options enable large-scale benchmarking, while module swapping allows easy scenario variation without re-engineering the entire synthesis process.
    \item \textit{Extensibility:} Each module can be independently customized or replaced, such as implementing novel safe control algorithms in the \textbf{Safety} module, integrating additional sensors in the \textbf{Task} module, testing new reinforcement learning algorithms in the \textbf{Policy} module, and incorporating latest humanoid robot in the \textbf{Configuration} module.
    This flexibility supports rapid development, wide-ranging experimental setups, and seamless adoption of new research innovations.
    \item \textit{Deployability:} The unified interface within the \textbf{Agent} module bridges simulation and physical hardware, simplifying the transition from prototyping to real-world experiments. 
    By supporting middleware like ROS and DDS in the \textbf{Agent} module, \spark ensures robust real-time performance, whether in controlled lab settings or complex human-robot interactions.
\end{itemize}

This clear separation of responsibilities across \spark's modules not only accelerates testing and benchmarking of safe control methods but also streamlines development and real-world deployment. Researchers can quickly iterate on scenarios and algorithms, and then confidently transfer their solutions to hardware platforms with minimal friction, all within one coherent framework.

%% file: sections/4_test_suite_v1.tex
\section{\spark Suite Options}
\label{sec: suite_options}

\begin{figure}[htbp]
    \centering
    \includegraphics[width=\linewidth]{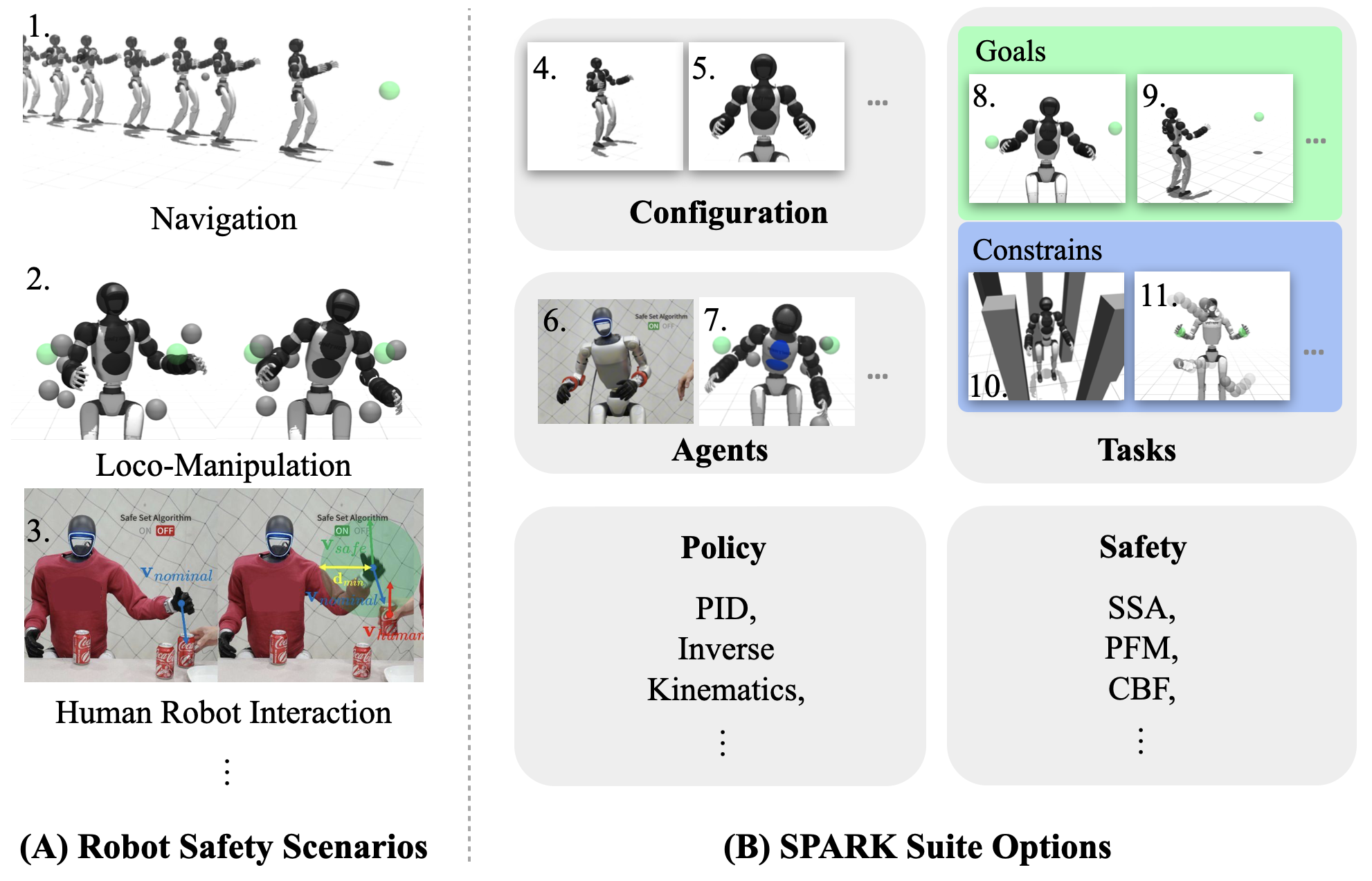}
    \caption{\spark Suite Options.}
    \label{fig:Suite}
\end{figure}
In addition to the safe control library, \spark offers users a comprehensive testing option suite to evaluate the performance of various safe controllers, which is shown in \Cref{fig:Suite}. 
\subsection{Configuration Options} 
\label{sec: configuration options}

As the humanoid robot represents a highly nonlinear and complex system, \spark provides users with predefined robot configurations based on the Unitree G1 humanoid robot. 
The robots are defined using MuJoCo XML files, and their reference frames are specified in \Cref{appendix: frame}.

The testing suite includes \textbf{8} types of robot configurations in the benchmark environments. Each configuration follows the format \texttt{\{Robot\}\_\{Dynamic\}}, where \texttt{\{Robot\}} specifies the base model and degrees of freedom, and \texttt{\{Dynamic\}} indicates the dynamics order.

The \texttt{{Robot}} component can be selected from the following \textbf{4} types:

\textbf{G1RightArm:} This robot configuration features the right arm of a humanoid robot with \textbf{7} degrees of freedom (DoFs), offering a simple single-manipulator model that can be easily adapted to other single-arm setups.

\textbf{G1FixedBase:} This robot configuration consists of \textbf{17} DoFs, including 7 DoFs for each arm and 3 DoFs for the waist. The pelvis of the robot is fixed relative to the world frame. The setup is designed to help users analyze the performance of safe controllers specifically for the robot manipulators.  

\textbf{G1MobileBase:}
This configuration includes \textbf{20} degrees of freedom (DoFs), consisting of 17 DoFs for the upper body and 3 additional DoFs for base motion. The base is modeled as a floating base in the world frame, with the 3 DoFs representing velocity along the x-axis, velocity along the y-axis, and yaw rotational velocity relative to the robot’s base frame. This can be interpreted as a wheeled mobile dual-arm manipulator. The setup is designed to evaluate safe controller performance for humanoid robots with both locomotion and manipulation capabilities. The floating base abstraction decouples upper-body motion from lower-body locomotion, enabling whole-body safety analysis without disturbances from locomotion dynamics.

\textbf{G1SportMode:}
This configuration shares the same DoFs as \textbf{G1MobileBase}. However, instead of using a floating base abstraction, the base motion in \textbf{G1SportMode} is realized through a learned policy that maps high-level 3-DoF base motion commands to joint-level control for the lower body. This provides a more realistic bipedal humanoid behavior, allowing users to evaluate safe control strategies under full-body dynamic coupling.

The \texttt{{Dynamics}} component can be selected from the following \textbf{2} types:

\textbf{Dynamic1:}
In this model, each controllable joint is treated as a single integrator. The system state includes only the joint positions, and the control inputs are the joint velocities. For \textbf{G1MobileBase} and \textbf{G1SportMode}, the state is defined in the world frame, while the control velocities are expressed in the robot’s base frame.

\textbf{Dynamic2:}
This model treats each controllable joint as a double integrator, where the system state includes both joint positions and velocities, and the control inputs are the joint accelerations. This dynamic setting poses greater challenges for safe control algorithms due to its increased complexity.

The details of the robot configurations are listed in \Cref{appendix: robot_config}

\begin{figure}[htbp]
    \centering
    \vspace{2cm}  
    \begin{tikzpicture}[transform canvas={xshift=0cm}] 
        \def\imgwidth{4.5cm}  

        \node at (0, 0) {\includegraphics[width=\imgwidth]{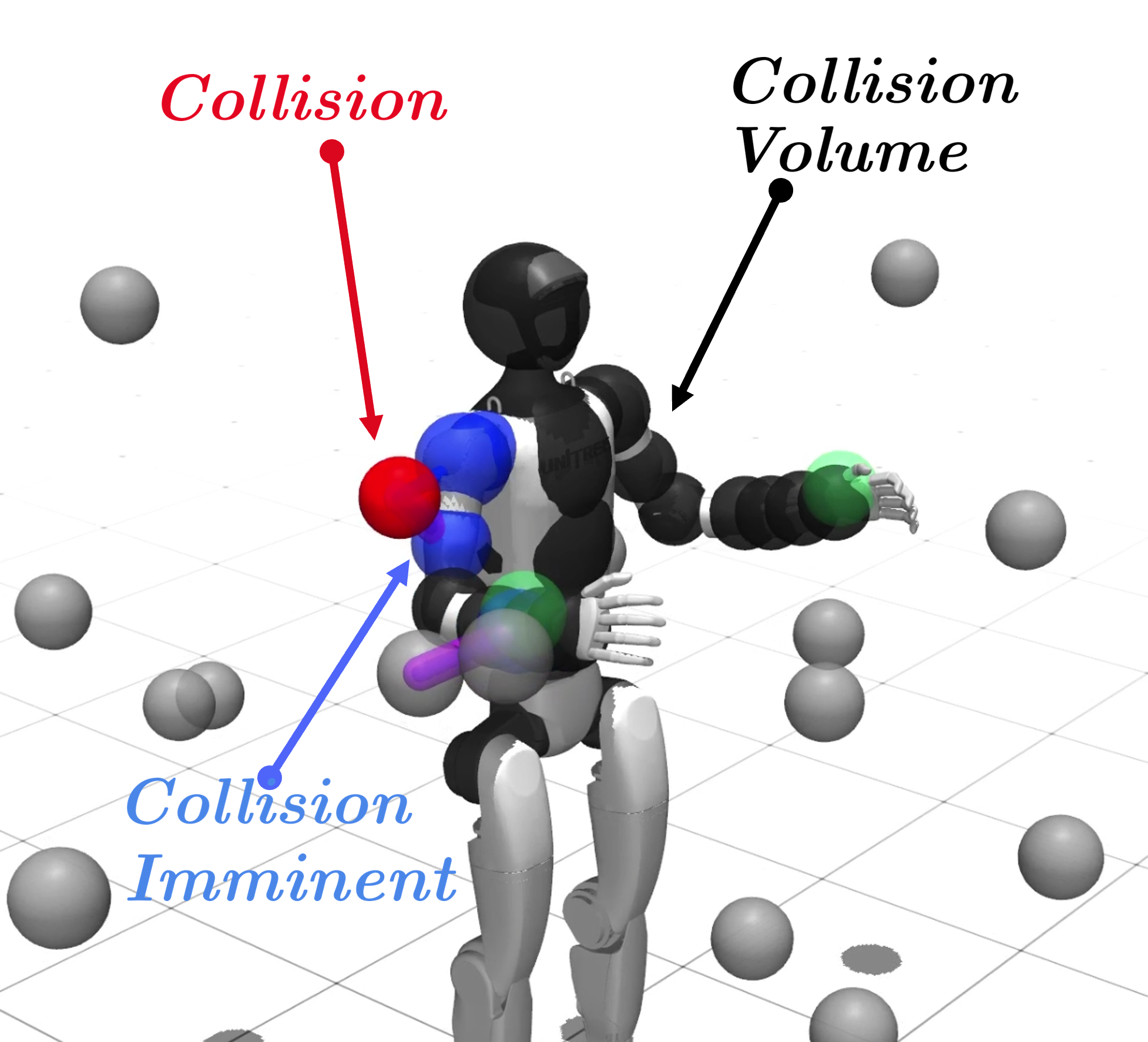}};
    \end{tikzpicture}
    \vspace{2.2cm}
    \caption{\spark whole body task environment.}
    \label{fig: whole_body}
\end{figure}

\subsection{Agent Options}
Our testing suite includes two types of agents: a simulation agent and a real robot agent, both of which support the configurations specified in the configuration options. We used a Unitree G1 humanoid robot, for which \spark provides an interface in the agent module. G1 humanoid physically features 29 DOFs, but we modeled it as a 20 DOF mobile dual-manipulator system by simplifying the locomotion.
For the upper body dynamics, each arm was treated as a general manipulator with seven DOFs, while the waist was equipped with three rotational joints (roll, pitch, yaw).
Regarding locomotion, three DOFs were considered: longitudinal, lateral, and rotational in the humanoid's body frame. 
Table~\ref{tb: tasks} in the Appendix shows the detailed configuration for the G1 humanoid.

\subsubsection{\textbf{Simulation Agent}}
As shown in \Cref{fig:Suite} (7). The simulation agent is implemented based on MuJoCo and serves as a benchmark tool for evaluating algorithm performance. Within the simulation environment, users can access the agent's position, joint positions, and other relevant states, facilitating the implementation of various control algorithms.
\subsubsection{\textbf{Real Robot Agent}}
As shown in \Cref{fig:Suite} (6). In the real robot agent, users can access low-level interfaces to read the robot's joint position data. Within the testing suite, we utilize Unitree's high-level control API to achieve raw pitch and yaw movements on a 2D plane.

\subsection{Task Options}  
The testing suite of \spark provides predesigned tasks to generate reference controls, enabling the evaluation of safe controllers under various scenarios.  
\subsubsection{Task Objectives Options}
To simulate manipulation and navigation tasks, \spark provides test suites with \textbf{two} types of goal configurations:


\textbf{Arm Goal:} 
As shown in \Cref{fig:Suite} (8). This goal is used to simulate the manipulation scenario. The task requires the robot to reach designated static 3D target positions with each hand while ensuring precise and safe movement.

\textbf{Base Goal:}
As shown in \Cref{fig:Suite} (9). This goal is used to simulate the navigation scenario. The task requires the robot to autonomously navigate to a specified 2D goal position while ensuring safe movement and avoiding obstacles in the environment.


In addition to the above two types of goals, users can also configure a \textbf{Goal Motion}, where the goal can be set to remain static or move dynamically at each time step. The goals for the arms and the robot base are marked in green spheres in \Cref{fig: whole_body}. 

\subsubsection{Task Constraint Options}
\label{subsec:task_constraints}

In \spark, \emph{task constraints} primarily address collision avoidance. Users can approximate the collision volumes of objects—such as obstacles and robot links—using the geometries provided by \spark. Each geometric primitive admits a signed distance function \(d\), allowing \spark to evaluate all pairwise distances in parallel with \(O(1)\) time complexity each pair, which adds minimal overhead for real-time safe control. The \textit{computational safety constraint} becomes $\min d\geq d_{min}$ where the minimization on the left is over all the pairwise distances $d$ over geometric primitives and the parameter $d_{min}$ is a user-specified distance threshold. The thresholds for self-collision and environmental collision could be different. The specific approach for handling these safety constraints is discussed in \Cref{subsubsec: safety_options}.

\textbf{Obstacle Geometries:}
Conventional collision checkers typically return only a binary collision flag~\cite{ericson2004real, van2003collision}. However, many safety controllers require the signed distance \(d\). To support these use cases, \spark includes two analytic primitives that allow for efficient minimum distance computation:
\textbf{Sphere}: $\mathcal{O}_{\text{sphere}}(\mathbf{p}, r)
            = \bigl\{\,
                \mathbf{x} \in \mathbb{R}^{3}
                \;\big|\;
                \|\mathbf{x}-\mathbf{p}\|_{2} \le r
              \bigr\}$,
where \(\mathbf{p}\in\mathbb{R}^{3}\) is the center and \(r>0\) is the radius. Sphere can be used to represent complex object in a simple but flexible way.
\textbf{Box}:$\
            \mathcal{O}_{\text{box}}(\mathbf{p},\mathbf{w})
            = \bigl\{\,
                \mathbf{x} \in \mathbb{R}^{3}
                \;\big|\;
                |x_i - p_i| \le \tfrac{w_i}{2},
                \; i \in \{x,y,z\}
              \bigr\},
          $
where \(\mathbf{p}\in\mathbb{R}^{3}\) is the center and \(\mathbf{w}=(w_x,w_y,w_z)^{\!\top}\) the edge lengths. Boxes efficiently model large, flat surfaces (e.g.\ tables) that would otherwise require many spheres.

Besides, each collision volume is stored together with a homogeneous transform giving its current pose, and spatial velocity for time-varying safety analyses.
This representation enables \spark to compute not only the signed distance
\(d\) but also its \emph{analytic gradient} which is also required by some safe controllers. Although numerical schemes such as centered finite differences could be used, they introduce extra evaluations and thus incur a significant performance penalty.

\textbf{Obstacle Motion:}
Obstacles can be either static or dynamic. Both types are represented as floating geometries that the robot must avoid. Trespassing incurs penalties. Static obstacles remain fixed, while dynamic obstacles move based on predefined patterns, such as Brownian motion.

\textbf{Obstacle Number:}
Users have the flexibility to control the number of obstacles within a task. For example, in our experiments described below, configuration v0 represents a scenario with 10 obstacles, whereas v1 corresponds to a denser environment with 50 obstacles.

The collision volumes of the robot and the environment obstacles are marked in \Cref{fig: whole_body}. \Cref{appendix: collision_config} reports the configuration of the robot collision volumes used in \spark.

\subsubsection{Task Interface Options}
Different agents can utilize distinct interfaces. For the Simulation Agent, users can control the position of obstacles using a keyboard, allowing for flexible and interactive environment adjustments within the simulation. In contrast, the Real Robot Agent leverages Apple Vision Pro to track human hand positions in real time, treating them as dynamic obstacles. This enables the real robot to perceive and react to obstacles in its environment, enhancing its ability to perform collision avoidance in real-world scenarios.

\subsection{Policy Options:}
We achieve humanoid locomotion tasks using PID control and implement manipulation tasks through a combination of PID control and inverse kinematics (IK). However, the choice of policy is not limited, as \spark supports user-defined policies, including data-driven approaches. Additionally, users can replace the standalone safety module with an end-to-end safe reinforcement learning policy, further demonstrating the extensibility of \spark. Since the model based safe controllers are the primary focus of \spark, we leave the integration of more complex control policies for future work.

\subsection{Safety Options:}
\label{subsubsec: safety_options}

The safe control of humanoid robots with respect to the computational safety constraints faces two major challenges: the presence of complex dynamics and the requirements for dexterous safety. Humanoids integrate mobile robotics and dual-arm manipulation, resulting in a high-dimensional, nonlinear system where the coupling of uncertain legged locomotion and high-DOF arm movements significantly increases control complexity. While some degrees of freedom (DOFs) operate independently, others—such as those affecting localization—impact the entire system, making precise tracking and safe motion in Cartesian space difficult. Beyond dynamic constraints, safety considerations further complicate control. Humanoids must navigate confined spaces, carefully adjusting their poses to avoid collisions with obstacles and themselves. Modeling at a limb level rather than a whole-body level is essential, but this introduces combinatorial safety constraints, creating a highly nonconvex safe state space that challenges real-time safety assurance.

While there has been no known method that can fully assure humanoid safety, there have been a few real-time safe control methods that successfully safeguarded lower-dimensional robotic systems. \spark implements these algorithms with the following purposes. First, users may directly use these algorithms to safeguard their humanoid robots in certain scenarios after some tuning. Second, users may compare new algorithms against these existing baselines. Finally, these algorithms are implemented as examples to show users how to modularize control algorithms so that they can be easily deployed with various configurations, agents, and tasks. 
It is worth mentioning that these algorithms are not the complete set of safety algorithms available nowadays, and we are leaving the integration of other state-of-the-art methods for future work such as safe control with learned certificates~\cite{dawson2023safe},  safe reinforcement learning based control~\cite{brunke2022safe} and safe learning for MPC~\cite{hewing2020learning}.
The implemented algorithms all belong to energy-function-based safety algorithms (or more broadly value-based safety methods) which first define a scalar energy function to characterize dynamic safety levels with its signs and then direct real-time control algorithms to maintain a certain energy level. With its simplicity and interpretability, we hypothesize that they have a high chance to achieve safety assurance for humanoid robots in complex scenarios in the future.

Consider the control-affine system dynamics:
\[
  \dot{\mathbf{x}} = \mathbf{f}(\mathbf{x}) + \mathbf{g}(\mathbf{x})\,\mathbf{u},
  \qquad
  \mathbf{u} \in \mathcal{U},
\]
where \(\mathbf{x}\) denotes the system state, \(\mathbf{u}\) the control input, and \(\mathcal{U}\) the admissible control set.
The objective of safe control is to synthesize a policy that produces a safe control input \(\mathbf{u}_{\mathrm{safe}}\), ensuring that the system state remains within a prescribed safe set $\Xb_\mathrm{s}$ while attempting to track a given reference control.

In \spark, users can select from a suite of baseline safe control algorithms.  
Each algorithm leverages an energy function \(\phi(\mathbf{x})\) to quantify the risk of leaving the safe set in the future. States for which \(\phi(\mathbf{x}) > 0\) are considered \emph{unsafe}.

A class of methods enforces safety by solving the following constrained optimization problem:
\begin{align}\label{eq:safe_control_problem}
\minimizewrt{\mathbf{u}}~~ & \|\mathbf{u} - \mathbf{u}_{\text{ref}} \|^2_{\mathbf{Q}_\mathbf{u}}   \\ \nonumber
\text{subject to}~~ & \dot{\mathbf{x}} = \mathbf{f}(\mathbf{x}) + \mathbf{g}(\mathbf{x})\mathbf{u} \\ \nonumber
& \text{Safety constraint based on } \phi(\mathbf{x}) \\ \nonumber
& \mathbf{u} \in \mathcal{U}
\end{align}
where \( \mathbf{u}_{\text{ref}} \) is the nominal reference control from a high-level planner, and \( \mathbf{Q}_{\mathbf{u}} \) is a diagonal matrix that defines the weights for the control cost.
The following three methods adopt the optimization-based control.

\subsubsection{Safe Set Algorithm}
Proposed by \cite{liu2014control}, Safe Set Algorithm (SSA) introduces a continuous, piecewise smooth energy function $\phi \defeq \cX \mapsto \RR$, or safety index, to quantify safety while considering the system dynamics.
A linear $n^\mathrm{th}$ ($n\geq 0$) order safety index $\phi_n$ has the following general form:
\begin{equation}\label{def:phi_recursive}
    \begin{aligned}
        \phi_n &= (1+a_1 s)(1+a_2 s)\dots(1+a_n s)\phi_0,
    \end{aligned}
\end{equation}
where $s$ is the differentiation operator.
\eqref{def:phi_recursive} can also be expanded as
$\phi_n \defeq \phi_0 + \textstyle\sum_{i=1}^{n}k_i \phi^{(i)}_0
$
where $\phi_0^{(i)}$ is the $i^\mathrm{th}$ time derivative of $\phi_0$.
$\phi_n$ should satisfy that (a) the characteristic equation $\prod_{i=1}^n(1+a_i s) = 0$ only has negative real roots to prevent overshooting of $\phi_0$ and (b) $\phi_0^{(n)}$ has relative degree one to the control input $\ub$. The safety index can also be shaped nonlinearly by replacing $\phi_0$ with another function $\phi_0^*$ that always has the same sign as $\phi_0$:
\begin{align}
    \phi_n \defeq \phi_0^* + \textstyle\sum_{i=1}^{n}k_i \phi^{(i)}_0.
\end{align}
To support both first-order and second-order dynamics described in \Cref{sec: configuration options}, the safety module includes predefined first-order and second-order safety indices, which can be computed using either numerical or analytical methods. 
For one pair-wise distance constraint, the computational safety specification can be written as
\[
\phi_0 = d_{\min} - d(\xb, \mathcal{O}),
\]
where \( d(\xb, \mathcal{O}) \) is the distance function described in \Cref{subsec:task_constraints}, and \( d_{\min} \geq 0 \) is the task-specific minimum safety distance that must be maintained. As the distance between the robot and the obstacle increases, the value of \( \phi_0 \) decreases, indicating that the state becomes safer.

The first-order safety index is then defined as
\[
\phi_1 = \phi_0 = d(\xb, \mathcal{O}) - d_{\min}.
\]

For the second-order safety index, \spark applies the nonlinear formulation to allow more design freedom:
\[
\phi_2 = d(\xb, \mathcal{O})^n - d_{\min}^n - k \, \dot{d}(\xb, \mathcal{O}),
\]
where \( n > 0 \) and \( k > 0 \) are user-defined parameters. When \( n = 1 \), this expression reduces to the linear form of the second-order safety index.

Nevertheless, to account for dexterous safety and allow modeling of the robot on the limb-level, we need to consider all pair-wise distance constraints. There are two strategies to account for all pair-wise distance constraints: one is to enforce the most critical constraint at every time step, and the other is to enforce all constraints at all times. It has been shown in \cite{chen2025dexterous} that the first approach leads to significant failures for humanoids in cluttered environments. 
\spark adopts the second approach and extends the conventional single-constrained safe control problem \cite{liu2014control} to handle multi-constrained cases. The targeted safe set of system states, $\Xb_\mathrm{s}$, is now defined by $M \geq 1$ energy functions $\phi$:
\begin{equation}
    \Xb_\mathrm{s} \coloneqq \{\xb \in \cX \mid \phi[i](\xb) \leq 0,~\forall i \in [M]\}.
\end{equation}  

Leveraging Nagumo's theorem, the safe control is selected to make the targeted safe set $\Xb_\mathrm{s}$ forward invariant. The generalized safe control problem solved by \spark is written as:  
\begin{align}\label{eq:safe_control_problem_spark}
    \minimize_{\mathbf{u}}~~ & \|\mathbf{u} - \mathbf{u}_{\text{ref}} \|^2_{\mathbf{Q}_\mathbf{u}} \\ \nonumber
    \st~~ & \dot{\mathbf{x}} = \mathbf{f}(\mathbf{x}) + \mathbf{g}(\mathbf{x})\mathbf{u}, \\ \nonumber
    & \forall i \in [M],~ \dot{\phi}[i](\xb, \ub) \leq -\eta_{ssa} \quad \text{if} \quad \phi[i](\xb) > 0, \\ \nonumber
    & \forall i \in [M],~ \dot{\phi}[i](\xb, \ub) \leq 0 \quad \text{if} \quad \phi[i](\xb) = 0.
\end{align}  
where $\eta_{ssa}$ is a positive constant. It is worth mentioning that the safety index $\phi$ needs to be properly synthesized to ensure that \eqref{eq:safe_control_problem_spark} always has a solution; but there has not been any known method that could obtain such a feasibility-guaranteed $\phi$. As a result, there is no guarantee that the targeted safe set  $\Xb_\mathrm{s}$ is forward invariant. While there are ongoing works studying the synthesis of $\phi$, various other approaches could be used to mitigate the infeasibility of \eqref{eq:safe_control_problem_spark} at run time, including adding slack variables~\cite{chen2025dexterous} or dropping certain constraints that led to infeasibility, etc. For now, \spark supports the slack variables to each constraints. These slack variables help ensure problem feasibility while striving to maintain safety as much as possible.  

\subsubsection{Control Barrier Function}

The Control Barrier Function (CBF) method \cite{ames2019control} differs from SSA in how the constraint on control is formed. It enforces safety constraints continuously by ensuring that  $\dot{\phi} < -\alpha(\phi)$,
where $\alpha: \mathbb{R} \to \mathbb{R}$ is a strictly increasing function with $\alpha(0) = 0$.  

In its simplest form, $\alpha$ can be chosen as a positive constant $\lambda_{cbf}$, leading to a straightforward implementation. When considering multiple safety constraints, the safe control problem solved using CBF can be formulated as:
\begin{align}\label{eq:safe_control_problem_cbf}
    \minimize_{\mathbf{u}}~~ & \|\mathbf{u} - \mathbf{u}_{\text{ref}} \|^2_{\mathbf{Q}_\mathbf{u}} \\ \nonumber
    \st~~ & \dot{\mathbf{x}} = \mathbf{f}(\mathbf{x}) + \mathbf{g}(\mathbf{x})\mathbf{u}, \\ \nonumber
    & \forall i \in [M],~ \dot{\phi}[i](\xb, \ub) \leq -\lambda_{cbf} \phi[i](\xb).
\end{align}  

As a result, the CBF may deviate from the reference control input $\mathbf{u}_{\text{ref}}$ even when the system is already safe (i.e., $\phi[i] < 0$). In such cases, the control input may fail to accurately track the nominal reference.

\subsubsection{Sublevel Safe Set Algorithm}

The Sublevel Safe Set (SSS) algorithm \cite{wei2019safe} combines the strengths of SSA and CBF to address their respective limitations. It solves the following safe control problem:  
\begin{align}\label{eq:safe_control_problem_sss}
    \minimize_{\mathbf{u}}~~ & \|\mathbf{u} - \mathbf{u}_{\text{ref}} \|^2_{\mathbf{Q}_\mathbf{u}} \\ \nonumber
    \st~~ & \dot{\mathbf{x}} = \mathbf{f}(\mathbf{x}) + \mathbf{g}(\mathbf{x})\mathbf{u}, \\ \nonumber
    & \forall i \in [M],~ \dot{\phi}[i](\xb, \ub) \leq -\lambda_{sss} \phi[i](\xb) \quad \text{if} \quad \phi[i](\xb) \geq 0.
\end{align}  

In the SSS algorithm, the control correction is only applied when $\phi[i](\xb) \geq 0$, allowing for a more efficient correction compared to CBF. Since the constraint is inactive when $\phi[i](\xb) < 0$, the robot achieves better performance while tracking the reference control input $\mathbf{u}_{\text{ref}}$. 



In addition to the optimization-based baselines, \spark{} also provides two projection methods that directly modify the reference input, thereby avoiding the need to solve constrained optimization problems:

\subsubsection{Potential Field Method}

PFM \cite{khatib1986real} computes the control input $\mathbf{u}$ indirectly by first deriving a Cartesian-space control input $\mathbf{u}_{\text{c}}$. By defining the function $\mathbf{c}_r = \mathbf{h}(\mathbf{x})$, which calculates the closest point on the robot to the obstacles, the Cartesian-space dynamics can be written as:
\begin{equation}
    \dot{\mathbf{c}}_r = \mathbf{u}_{\text{c}} := \mathbf{u}_{\text{c-ref}} + \mathbf{u}_c^*,
\end{equation}
where $\mathbf{u}_{\text{c-ref}}$ is the reference control in the Cartesian space transformed from $\mathbf{u}_{\text{ref}}$ via:
\begin{equation}
    \mathbf{u}_{\text{c-ref}} = \nabla{\mathbf{h}}(\mathbf{x}) \mathbf{f}(\mathbf{x}) + \nabla{\mathbf{h}}(\mathbf{x}) \mathbf{g}(\mathbf{x}) \mathbf{u}_{\text{ref}}.
\end{equation}
The term $\mathbf{u}_c^*$ represents a repulsive ``force" added to the reference $\mathbf{u}_{\text{c-ref}}$ to push the robot away from the obstacles whenever the safety constraint is violated. Specifically:
\begin{equation}
    \mathbf{u}_c =
    \begin{cases} 
        \mathbf{u}_{\text{c-ref}} - c_{pfm} \nabla \tilde{\phi} & \text{if } \tilde{\phi}(\mathbf{c}_r) \geq 0, \\
        \mathbf{u}_{\text{c-ref}} & \text{otherwise},
    \end{cases}
\end{equation}
where $c_{pfm} > 0$ is a tunable constant and $\tilde{\phi}(\mathbf{c}_r)$ is the energy function with respect to the Cartesian point $\mathbf{c}_r$. Finally, the equivalent control input $\mathbf{u}$ in the configuration space is derived from $\mathbf{u}_c$.

\subsubsection{Sliding Mode Algorithm}

SMA \cite{gracia2013reactive} ensures safety by maintaining the system state around a sliding layer defined by $\phi = 0$ whenever the safety constraint is violated. 

For a single safety index $\phi$, its gradient can be decomposed as:
\begin{align}
    \dot{\phi}(\mathbf{x}) &= \nabla \phi^\top(\mathbf{x}) \dot{\mathbf{x}} \\ \nonumber
    &= \nabla \phi^\top(\mathbf{x}) \big( \mathbf{f}(\mathbf{x}) + \mathbf{g}(\mathbf{x}) \mathbf{u} \big) \\ \nonumber
    &= \underbrace{\nabla \phi^\top(\mathbf{x}) \mathbf{f}(\mathbf{x})}_{L_\mathbf{f} \phi} + \underbrace{\nabla \phi^\top(\mathbf{x}) \mathbf{g}(\mathbf{x})}_{\mathbf{L}_\mathbf{g} \phi} \mathbf{u},
\end{align}
where the term $\mathbf{L}_\mathbf{g} \phi$ represents the sensitivity of the safety index $\phi$ to the control input $\mathbf{u}$.

To handle multiple constraints, \spark considers only the most unsafe safety index, denoted as $\phi_{\max}$, and corrects the reference control input as follows:
\begin{align}
    \mathbf{u} = 
    \begin{cases} 
        \mathbf{u}_{\text{ref}} - c_{sma} \mathbf{L}_\mathbf{g} \phi_{\max}^\top & \text{if } \phi_{\max} \geq 0, \\
        \mathbf{u}_{\text{ref}} & \text{otherwise},
    \end{cases}
\end{align}
where the constant $c_{sma} > 0$ is set sufficiently large to ensure that:
\begin{equation}
    \dot{\phi}_{\max} = L_\mathbf{f} \phi_{\max} + \mathbf{L}_\mathbf{g} \phi_{\max} \mathbf{u} - c_{sma} \|\mathbf{L}_\mathbf{g} \phi_{\max}\|^2 < 0.
\end{equation}

By doing so, SMA ensures that the system remains in a safe state while effectively handling multiple constraints.

The five safety control algorithms introduced above constitute the set of safety options provided by \spark to handle the task constraints described in \Cref{subsec:task_constraints}. 
Here, we briefly describe how \spark automates the computation of \( \dot{\phi} \) to efficiently implement the provided safety controllers. Detailed derivations and examples are presented in Appendix~\Cref{appendix: safety_index_intro}.

In \spark, \( \dot{\phi} \) is computed through a hierarchical pipeline. First, it is expressed in terms of the time derivative of the Cartesian state:
\begin{equation}
\dot{\phi} = \frac{\partial \phi(\mathbf{x}_c)}{\partial \mathbf{x}_c} \, \dot{\mathbf{x}}_c,
\end{equation}
which captures how the safety index evolves with respect to the motion of the robot and obstacle volumes. By applying the Cartesian dynamics, it can be further written in terms of the Cartesian control input:
\begin{equation}
\dot{\phi} = \frac{\partial \phi(\mathbf{x}_c)}{\partial \mathbf{x}_c} \, \mathbf{f}_c(\mathbf{x}_c) + \frac{\partial \phi(\mathbf{x}_c)}{\partial \mathbf{x}_c} \, \mathbf{g}_c(\mathbf{x}_c)\mathbf{u}_c.
\end{equation}
This formulation is directly usable by algorithms such as PFM, which operate in Cartesian space.

Next, the time derivative is mapped to the configuration space by applying the appropriate Jacobian:
\begin{equation}
\dot{\mathbf{x}}_c = \mathbf{J}(\boldsymbol{\theta}) \, \dot{\mathbf{x}}_q,
\end{equation}
relating the Cartesian state derivatives to the robot’s configuration space. Finally, by applying the configuration space dynamics or leveraging the relationship between Cartesian and configuration space control, \( \dot{\phi} \) can be expressed as a function of the configuration space control input $\mathbf{u}_q$. This enables compatibility with optimization-based methods such as SSA, SSS, and CBF. Moreover, the partial derivative \( \frac{\partial \dot{\phi}}{\partial \mathbf{u}_q} \) can be utilized by SMA to construct gradient-based safety filters.

In summary, a key contribution of \spark is the modularization of the value function computation and its time derivative, allowing users to focus solely on designing the safety index \( \phi \) and specifying their control law, such as a QP-based controller.

\subsection{Evaluation metrics}\label{subsec:evaluation_metrics}

\spark provides four metrics to assess the performance of various safe control methods across different benchmark scenarios, evaluating both their safety and efficiency:
\begin{itemize}
    \item Step-wise average arm goal tracking score $J_{arm}$.
    \item Step-wise average base goal tracking score $J_{base}$.
    \item Step-wise average self safety score $M_{self}$.
    \item Step-wise average environment safety score $M_{env}$.
\end{itemize}

Formally,

\begin{align}
    &J_{arm} =\frac{1}{T}\sum_{t=0}^T \mathcal{G}(\Delta d_{arm}, \sigma_{arm}), \Delta d_{arm} \geq 0\label{eq: arm goal score}\\ 
    &J_{base} =\frac{1}{T}\sum_{t=0}^T \mathcal{G}(\Delta d_{base}, \sigma_{base}), \Delta d_{base} \geq 0\label{eq: base goal score}\\
    &M_{self} =\frac{1}{T}\sum_{t=0}^T \mathcal{G}(\Delta d_{self},\sigma_{self}), \Delta d_{self} \leq 0\label{eq: self safety score}\\
    &M_{env} =\frac{1}{T}\sum_{t=0}^T \mathcal{G}(\Delta d_{env} ,\sigma_{env}), \Delta d_{env} \leq 0\label{eq: environment safety score}\\ \nonumber
\end{align}
where $\Delta d_{arm}$ is the distance from the robot's hands to the corresponding goals, $\Delta d_{base}$ is the distance from the robot's pelvis to the corresponding base goal, $\Delta d_{env}$ is the violated distance between the robot and obstacles, and $d_{self}$ is the violated distance between the robot's collision volumes. 

$\mathcal{G}$ is the goal tracking score function, which converts the distance into a score within the range $[0, 1]$. The closer the distance to $0$, the higher the score. 

The score function is defined as follows:
\begin{align}
    &\mathcal{G}(\Delta d, \delta) = 
    \exp\bigg[-\frac{\Delta d^2}{\sigma}\bigg]
\end{align}

In the following sections, we demonstrate the versatility of the \spark framework by deploying its safe control pipelines across various use cases. 
The agents range from a simulated humanoid robot to the real Unitree G1 humanoid robot, while the task inputs vary from autonomous target trajectories to teleoperation commands. Additionally, the environments span from static confined spaces to dynamic human-robot interaction scenarios. 
The various use cases are designed to serve multiple purposes, including benchmarking safe control algorithms \ref{sec: usecase_benchmark}, allowing users to teleoperate a simulated robot with visual feedback \ref{sec: usecase_safe_teleop_sim}, enabling safe teleoperation for human-robot interaction (HRI) scenarios \ref{sec: usecase_safe_teleop_real}, and deploying them on real robots \ref{sec: usecase_safe_auto_real}. The details of each use case, along with their corresponding sections, are outlined in \Cref{tab:config_agent_task_obstacle_control_safety}.

\input{table_template/tables/agent_task}

\input{table_template/tables/eva_full_table}

%% file: table_template/tables/agent_task.tex
\begin{table*}[ht]
\caption{Use Cases}
\centering
\begin{tabular}{lcccccc}
    \toprule
    \multirow{2}{*}{\centering Use Cases} & \multirow{2}{*}{\centering Configuration} & \multirow{2}{*}{\centering Agent} & \multicolumn{2}{c}{Task} & \multirow{2}{*}{\centering Policy} & \multirow{2}{*}{\centering Safety} \\
    \cmidrule(lr){4-5}
    & & & Objective & Constraints & & \\
    \midrule
    Benchmark \ref{sec: usecase_benchmark} & All & Simulation & All (Autonomy) & All & PID+IK & All \\
    Simulation teleoperation \ref{sec: usecase_safe_teleop_sim} & G1FixedBase & Simulation & Arm Goal (Teleoperation) & Static & PID+IK & SSA \\
    Real autonomy \ref{sec: usecase_safe_auto_real} & G1FixedBase & Real Robot & Arm Goal (Autonomy) & Dynamic & PID+IK & SSA \\
    Real teleoperation \ref{sec: usecase_safe_teleop_real} & G1FixedBase & Real Robot & Arm Goal (Teleoperation) & Dynamic & PID+IK & SSA \\
    \bottomrule
\end{tabular}
\label{tab:config_agent_task_obstacle_control_safety}
\end{table*}

%% file: table_template/tables/eva_full_table.tex
\begin{table*}[ht]
\centering
\captionsetup{width=0.8\linewidth}
\caption{Performance metrics for different tasks. The table presents the scores obtained by each algorithm using its optimal parameters. The final scores are computed based on these selections. \textbf{Bold}: The highest score among different algorithms for the same task and metric. \textcolor{blue}{\textbf{Blue}}: The lowest score among different algorithms for the same task and metric.
}
\label{tab:alg_metric_task}

\scalebox{1.0}{
\begin{tabular}{cc|cccccccc}
\toprule
\textbf{Algorithm} & \textbf{Metric} & \multicolumn{2}{c}{\textbf{G1MobileBase\_D1}} & \multicolumn{2}{c}{\textbf{G1MobileBase\_D1}} & \multicolumn{2}{c}{\textbf{G1FixedBase\_D1}} & \multicolumn{2}{c}{\textbf{G1FixedBase\_D1}} \\
\cmidrule(lr){3-4} \cmidrule(lr){5-6} \cmidrule(lr){7-8} \cmidrule(lr){9-10}
 & & \textbf{WG\_SO\_v0} & \textbf{WG\_SO\_v1} & \textbf{WG\_DO\_v0} & \textbf{WG\_DO\_v1} & \textbf{AG\_SO\_v0} & \textbf{AG\_SO\_v1} & \textbf{AG\_DO\_v0} & \textbf{AG\_DO\_v1} \\
\midrule
\multirow{4}{*}{SSA} & $J_{arm}$  & 0.8691 & 0.8318 & 0.8536 & 0.6780 & 0.6202 & 0.7090 & 0.5810 & 0.5680 \\
                     & $J_{base}$ & \textbf{0.5517} & 0.5805 & \textbf{0.5065} & \textbf{0.6221} & NA     & NA     & NA     & NA \\
                     & $M_{self}$ & \textbf{1.0000} & \textbf{1.0000} & \textbf{1.0000} & \textbf{1.0000} & \textbf{1.0000} & \textbf{1.0000} & \textbf{1.0000} & \textbf{1.0000} \\
                     & $M_{env}$  & 0.9010 & 0.8784 & 0.8876 & 0.7222 & 0.7590 & 0.6815 & 0.3448 & 0.3745 \\
\midrule
\multirow{4}{*}{PFM} & $J_{arm}$  & \textcolor{blue}{\textbf{0.8535}} &  \textcolor{blue}{\textbf{0.6357}} &  \textcolor{blue}{\textbf{0.8482}} &  \textcolor{blue}{\textbf{0.6502}} &  \textcolor{blue}{\textbf{0.4019}} &  \textcolor{blue}{\textbf{0.3077}} &  \textcolor{blue}{\textbf{0.3143}} &  \textcolor{blue}{\textbf{0.3642}} \\
                     & $J_{base}$ &  \textcolor{blue}{\textbf{0.5062}} & 0.5584 & 0.4971 &  \textcolor{blue}{\textbf{0.5839}} & NA     & NA     & NA     & NA \\
                     & $M_{self}$ & \textbf{1.0000} & \textbf{1.0000} & \textbf{1.0000} & 0.9658 & \textbf{1.0000} & \textbf{1.0000} &  \textcolor{blue}{\textbf{0.6328}} & \textbf{1.0000} \\
                     & $M_{env}$  &  \textcolor{blue}{\textbf{0.8374}} &  \textcolor{blue}{\textbf{0.5390}} & 0.5977 & 0.3364 &  \textcolor{blue}{\textbf{0.4767}} &  \textcolor{blue}{\textbf{0.3649}} & 0.2794 &  \textcolor{blue}{\textbf{0.2299}} \\
\midrule
\multirow{4}{*}{CBF} & $J_{arm}$  & 0.8761 & 0.8290 & 0.8662 & 0.7275 & 0.6742 & 0.5726 & 0.5538 & 0.5705 \\
                     & $J_{base}$ & 0.5277 & \textbf{0.5910} & 0.5060 & 0.6193 & NA     & NA     & NA     & NA \\
                     & $M_{self}$ & \textbf{1.0000} & \textbf{1.0000} & \textbf{1.0000} & 0.9472 & \textbf{1.0000} & \textbf{1.0000} & \textbf{1.0000} & \textbf{1.0000} \\
                     & $M_{env}$  & 0.9819 & 0.9596 &  \textcolor{blue}{\textbf{0.3674}} & 0.8995 & \textbf{0.8687} & \textbf{0.8631} &  \textcolor{blue}{\textbf{0.2387}} & 0.2512 \\
\midrule
\multirow{4}{*}{SMA} & $J_{arm}$  & \textbf{0.8808} &  \textbf{0.8649} & \textbf{0.8816} & \textbf{0.8443} & 0.6276 & 0.7028 & 0.5720 & \textbf{0.5915} \\
                     & $J_{base}$ & 0.5116 &  \textcolor{blue}{\textbf{0.5268}} &  \textcolor{blue}{\textbf{0.4824}} & 0.6059 & NA     & NA     & NA     & NA \\
                     & $M_{self}$ & \textbf{1.0000} & \textbf{1.0000} & \textbf{1.0000} &  \textcolor{blue}{\textbf{0.6619}} & \textbf{1.0000} & \textbf{1.0000} & \textbf{1.0000} & \textbf{1.0000} \\
                     & $M_{env}$  & 0.8582 & 0.6254 & 0.4925 &  \textcolor{blue}{\textbf{0.2863}} & 0.6994 & 0.5876 & 0.5345 & \textbf{0.4155} \\
\midrule
\multirow{4}{*}{SSS} & $J_{arm}$  & 0.8761 & 0.8306 & 0.8732 & 0.7253 & \textbf{0.6752} & \textbf{0.7179} & \textbf{0.5839} & 0.5634 \\
                     & $J_{base}$ & 0.5281 & 0.5907 & 0.4997 & 0.6170 & NA     & NA     & NA     & NA \\
                     & $M_{self}$ & \textbf{1.0000} & \textbf{1.0000} & \textbf{1.0000} & 0.9497 & \textbf{1.0000} & \textbf{1.0000} & \textbf{1.0000} & \textbf{1.0000} \\
                     & $M_{env}$  & \textbf{0.9844} & \textbf{0.9685} & \textbf{0.9380} & \textbf{0.9026} & 0.8618 & 0.6728 & 0.3538 & 0.3749 \\
\bottomrule
\end{tabular}
} 
\end{table*}

%% file: sections/5_usecase_benchmark.tex
\section{Use Case 1: \\Benchmarking Safe Control Algorithms} \label{sec: usecase_benchmark}

To demonstrate the capability of SPARK as a benchmarking toolbox for safe control, we use it to systematically evaluate a variety of algorithms under diverse constraints and objectives. By leveraging \spark’s suite functionalities, we ensure fair comparisons and gain insights into algorithmic performance, safety-efficiency trade-offs, and the impact of task complexity. In this use case, we utilize \spark’s composability to extract two configurations of the Unitree G1 robot from the Configuration Module and retrieve the corresponding simulation agent from the Agent Module. Within the Task Module, we combine different goals and constraints to generate benchmark scenarios. The Policy Module is then employed to execute control, and the Safety Module is used to analyze the performance and safety characteristics of different algorithms. Through this experimental setup, we aim to address the following research questions:

\textbf{Q1:}  
What are the overall benchmark results?  

\textbf{Q2:}  
How do various safe control algorithms balance the trade-off between safety and efficiency?  

\textbf{Q3:}  
How do different types of constraints affect algorithm performance?  

\textbf{Q4:}  
How does task complexity influence algorithm performance?  

\textbf{Q5:}  
How does the success rate of each algorithm compare?

\subsection{Experimental Setup}  
\paragraph{Experiment tasks}
By combining the robot, task, and constraint options introduced in \Cref{sec: suite_options}, $\textbf{8}$ benchmark tests are designed for the evaluation and comparison of algorithm performance. These predefined benchmark testing suites adhere to the standardized format: \texttt{\{Robot\}\_\{Dynamic\}\\\_\{Task\}\_\{Constraint\}\_\{Version\}}. The configuration of all experiment tasks is introduced in \Cref{appendix: task_config}. In this format, \textbf{WG} represents \textit{Whole Body Goal}, \textbf{AG} stands for \textit{Arm Goal}, \textbf{SO} denotes \textit{Static Obstacle}, and \textbf{DO} refers to \textit{Dynamic Obstacle}. In particular, this benchmark does not provide WG as a standalone task; instead, WG indicates a combination of both \textit{Arm Goal} and \textit{Base Goal}. In addition to obstacles, self-collision is also taken into consideration. For more details, please refer to \Cref{appendix: self_collision}. In our experiments, we analyze and compare the performance of these algorithms using the metrics provided in \Cref{subsec:evaluation_metrics}.
We use $\sigma_{arm} = 0.002$, $\sigma_{base} = 0.05$, and $\sigma_{env} =\sigma_{self} = 0.0002$ to scale the scores to fit the magnitude of the raw distance.

\paragraph{Comparison Group}  
To evaluate the performance of the \spark safe control library, we define a comparison group consisting of various alternative approaches. The methods in this group include all the techniques currently available in the safe control library \spark(version 0.1):
(i) Safe Set Algorithm (SSA)~\citep{liu2014control},  
(ii) Control Barrier Function (CBF)~\citep{ames2019control},  
(iii) Sublevel Safe Set (SSS)~\citep{wei2019safe},  
(iv) Potential Field Method (PFM)~\citep{khatib1986real}, and
(v) Sliding Mode Algorithm (SMA)~\citep{gracia2013reactive}.

\subsection{Overall Benchmark Results}
\Cref{tab:alg_metric_task} presents the overall benchmark performance results, evaluating different safety controllers in terms of safety and efficiency metrics. The experimental results indicate significant variations in the performance of different methods across various tasks and robotic setups. Specifically, in terms of motion accuracy and environmental safety, PFM exhibited the weakest overall performance, whereas optimization-based methods such as SSA, CBF, and SSS demonstrated a superior balance between safety and efficiency. Notably, despite not relying on optimization for safety control, SMA still managed to achieve a commendable trade-off between safety and efficiency.

PFM consistently showed low motion accuracy and environmental safety scores across all tasks, with particularly poor performance in fixed-base robotic tasks, where both its safety and execution efficiency were significantly lower than those of other methods. This suggests that PFM struggles to ensure feasibility in complex environments and is more susceptible to environmental constraints, leading to a higher risk of collisions. The primary issue with PFM is its reliance solely on repulsive forces in low-dimensional Cartesian space, making it ineffective in handling complex constraints in high-dimensional joint space, thereby limiting its safety control capabilities.

In contrast, SSA exhibited stable environmental safety across all tasks, demonstrating strong adaptability to external constraints. However, in fixed-base robotic tasks, SSA's motion accuracy was somewhat reduced, indicating that it may encounter challenges when dealing with scenarios where movement degrees of freedom are restricted.

CBF excelled in environmental safety, significantly outperforming PFM and showcasing its advantage in minimizing environmental collisions. However, compared to SSA, CBF exhibited slightly lower motion accuracy, suggesting that in certain tasks, it may sacrifice some execution efficiency to achieve higher safety.

As an optimization-based method, SSS demonstrated stable safety and execution efficiency across multiple tasks. For example, in some mobile robot tasks, its motion accuracy was comparable to SSA and CBF, while its environmental safety performance exceeded that of other methods. Additionally, in fixed-base robotic tasks, SSS achieved higher motion accuracy than other optimization methods, indicating that its optimization strategy effectively adapts to task requirements. 

Unlike optimization-based methods, SMA does not rely on optimization for safety control; instead, it refines control signals by decomposing the gradient of the safety index, thereby preserving effective task execution capabilities. The experimental results reveal that SMA generally achieved higher motion accuracy than other methods in most tasks, while also maintaining high stability in environmental safety. This suggests that SMA can reduce environmental collision risks while ensuring smooth task execution.

\subsection{Trade-off between Safety and Efficiency}

\begin{figure}[htbp]
    \centering
    \begin{tikzpicture} 
        \def\imgwidth{4cm}  
        \def\xgap{4.5}  

        \node at (0, 0) {\includegraphics[width=\imgwidth]{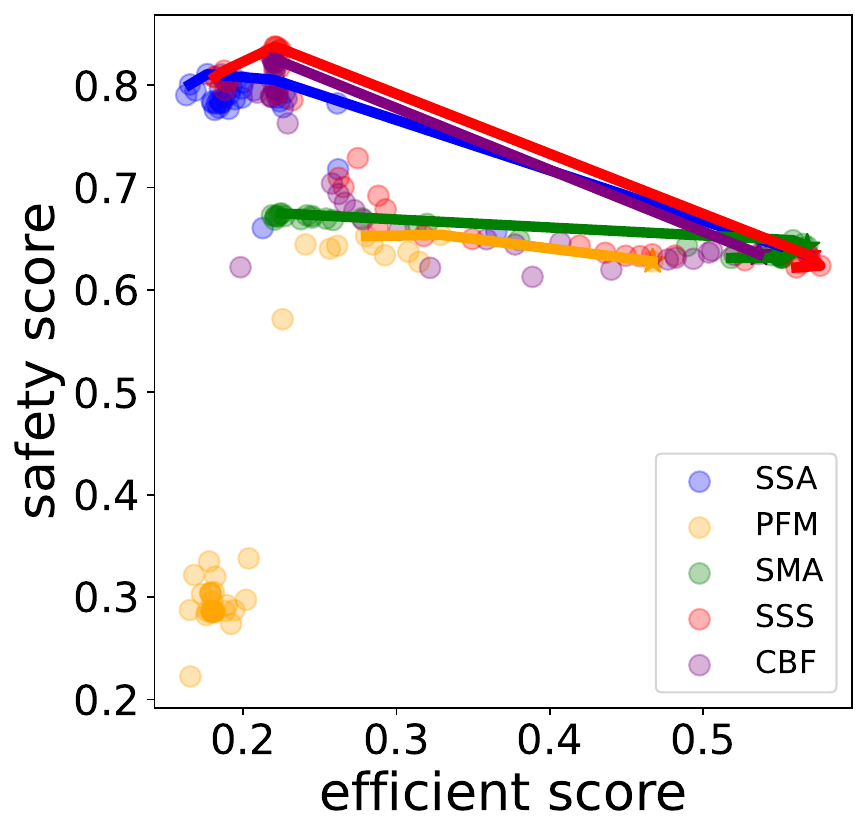}};
        \node[below, align=center, font=\small] at (0, -2) {(a) 
        G1FixedBase\_D1\\\_AG\_DO\_v1};

        \node at (\xgap, 0) {\includegraphics[width=\imgwidth]{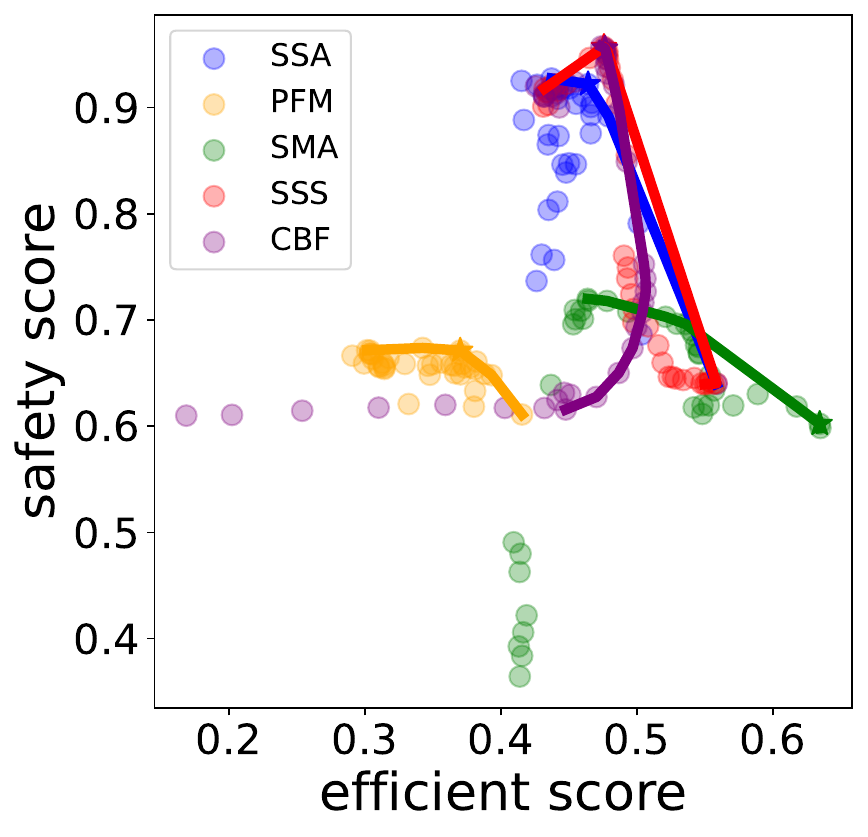}};
        \node[below, align=center, font=\small] at (\xgap, -2) {(b) G1MobileBase\_D1\\\_WG\_DO\_v1};

    \end{tikzpicture}

    \caption{Comparison of trade-off curves between safety and efficiency for different robot configurations.}
    \label{fig: param_tuning_subset}
\end{figure}

Balancing performance between safety and efficiency is a key challenge for every safe controller. To explore this trade-off, we tuned the parameters of each safe control algorithm ($\eta_{\text{ssa}}, \lambda_{\text{cbf}}, \lambda_{\text{sss}}, c_{\text{pfm}}, c_{\text{sma}}$) across a range of scales: from 0 to 1 in steps of 0.1, from 1 to 10 in steps of 1, from 10 to 100 in steps of 10, and from 100 to 1000 in steps of 10. These experiments were conducted under different tasks, enabling us to plot safety and efficiency performance on the same graph. By extracting the convex hull of the sampled parameters, We generated the trade-off curves and selected two representative scenarios, as shown in \Cref{fig: param_tuning_subset}, which illustrate how each controller balances safety and efficiency; the full set of results is provided in Appendix~\Cref{app: param_tuning_app}.
The efficiency is defined as $\frac{J_{arm} + J_{base}}{2}$ and is plotted on the x-axis, while the average safety score ($\frac{M_{self}+ M_{env}}{2}$) is plotted along the y-axis.

The results indicate that as parameters vary, safety and efficiency may adjust in opposition to each other. On average, SMA and PFM achieve better balance, while SSA, SSS and CBF excel at achieving higher safety scores without collisions. This behavior arises because, when their respective parameters \(c_{sma}\) and \(c_{pfm}\) approach zero, SMA and PFM cease correcting the reference control and revert to nominal controllers. However, the yellow curves in \Cref{fig: param_tuning_subset} shows that PFM exhibits a less favorable trade-off curve compared to SMA due to its incompatibility with high-dimensional robotic systems and environments with multiple constraints.

Unlike PFM and SMA, which project the reference control onto a safe control along a fixed direction, optimization-based methods search for an optimal safe control input within the admissible control space constrained by the safety conditions.
These methods prioritize ensuring safety over minimizing control deviations from the reference control. Unlike other optimization-based methods, CBF simultaneously impacts both safety and efficiency. This is due to its control law, which enforces that \(\dot{\phi}\) remains below an upper bound.

When the CBF parameter \(\lambda_{cbf}\) is too small, the robot may reduce the safety index unnecessarily even in safe environments, negatively impacting efficiency. Additionally, a small \(\lambda_{cbf}\) limits the safety constraints' ability to react rapidly in unsafe situations. Conversely, a large \(\lambda_{cbf}\) imposes overly strict safety constraints, even for minor safety violations, which can hinder efficiency. From \Cref{fig: param_tuning_subset}, it can be observed that CBF (purple) has more data points concentrated near the origin compared to the other two optimization-based methods, SSA and SSS. These points represent parameter settings that fail to achieve either safety or efficiency.

In the other comparison experiments, the parameter for each algorithm is picked to achieve the best trade-off balancing. The optimal parameters for each algorithm and task are marked in \Cref{fig: param_tuning} and \Cref{appendix: hyperparameters} reports the parameters used for the rest experiments.

\subsection{Impacts of Constraint Types}

\begin{figure}[htbp]
    \centering
    \begin{tikzpicture} 
        \def\imgwidth{4cm}  
        \def\xgap{4.5}  
        \def\ygap{-3.5} 

        
        \node at (0, 0) {\includegraphics[width=\imgwidth]{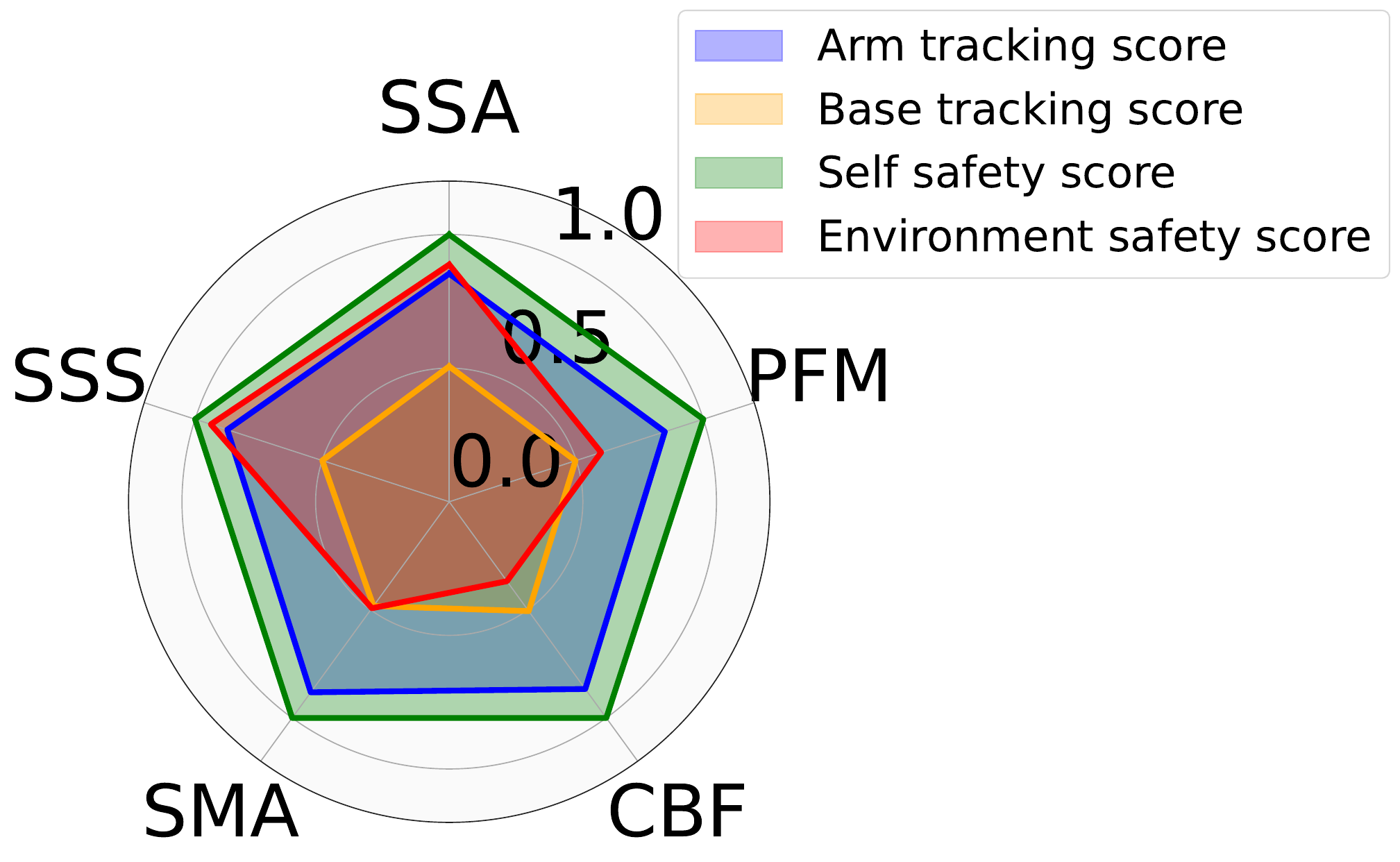}};
        \node[below, align=center, font=\small] at (-0.5, -1.5) {(a) G1MobileBase\_D1\\\_WG\_DO\_v0};

        \node at (\xgap, 0) 
        {\includegraphics[width=\imgwidth]{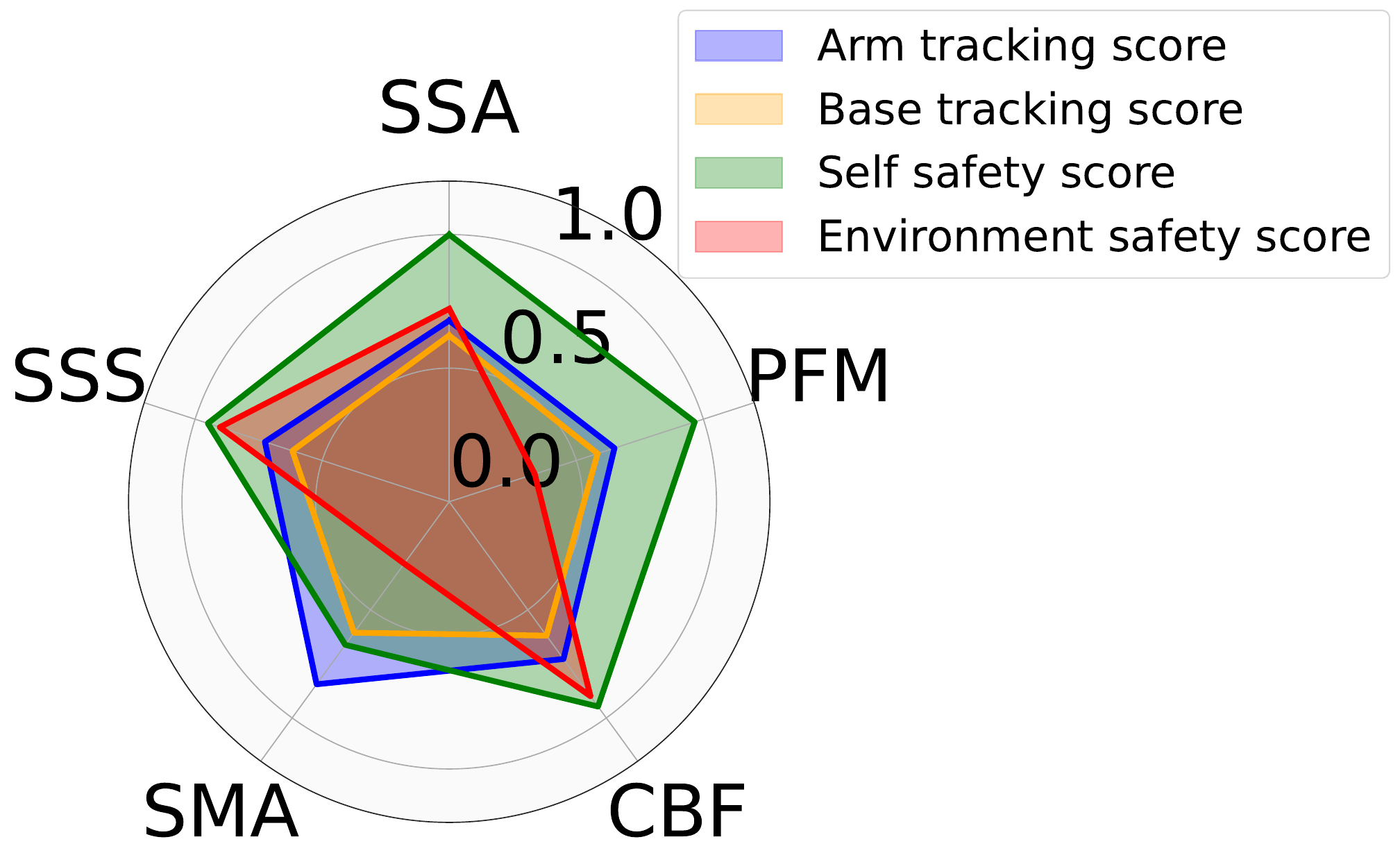}};
        \node[below, align=center, font=\small] at (\xgap-0.5, -1.5) {(b) G1MobileBase\_D1\\\_WG\_DO\_v1};

        \node at (0, \ygap) 
        {\includegraphics[width=\imgwidth]{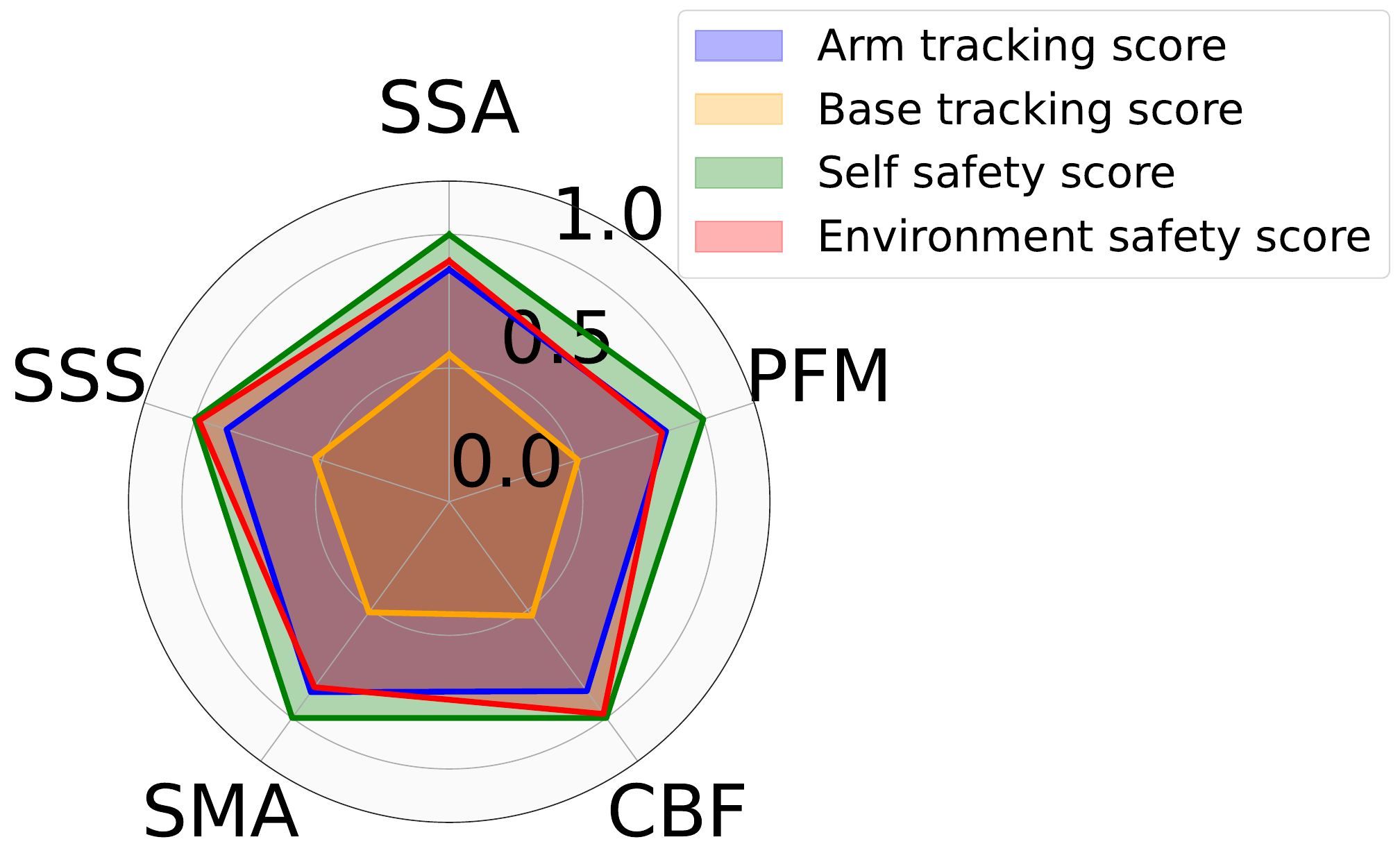}};
        \node[below, align=center, font=\small] at (-0.5, \ygap-1.5) 
        {(c) G1MobileBase\_D1\\\_WG\_SO\_v0};
        
        \node at (\xgap, \ygap) 
        {\includegraphics[width=\imgwidth]{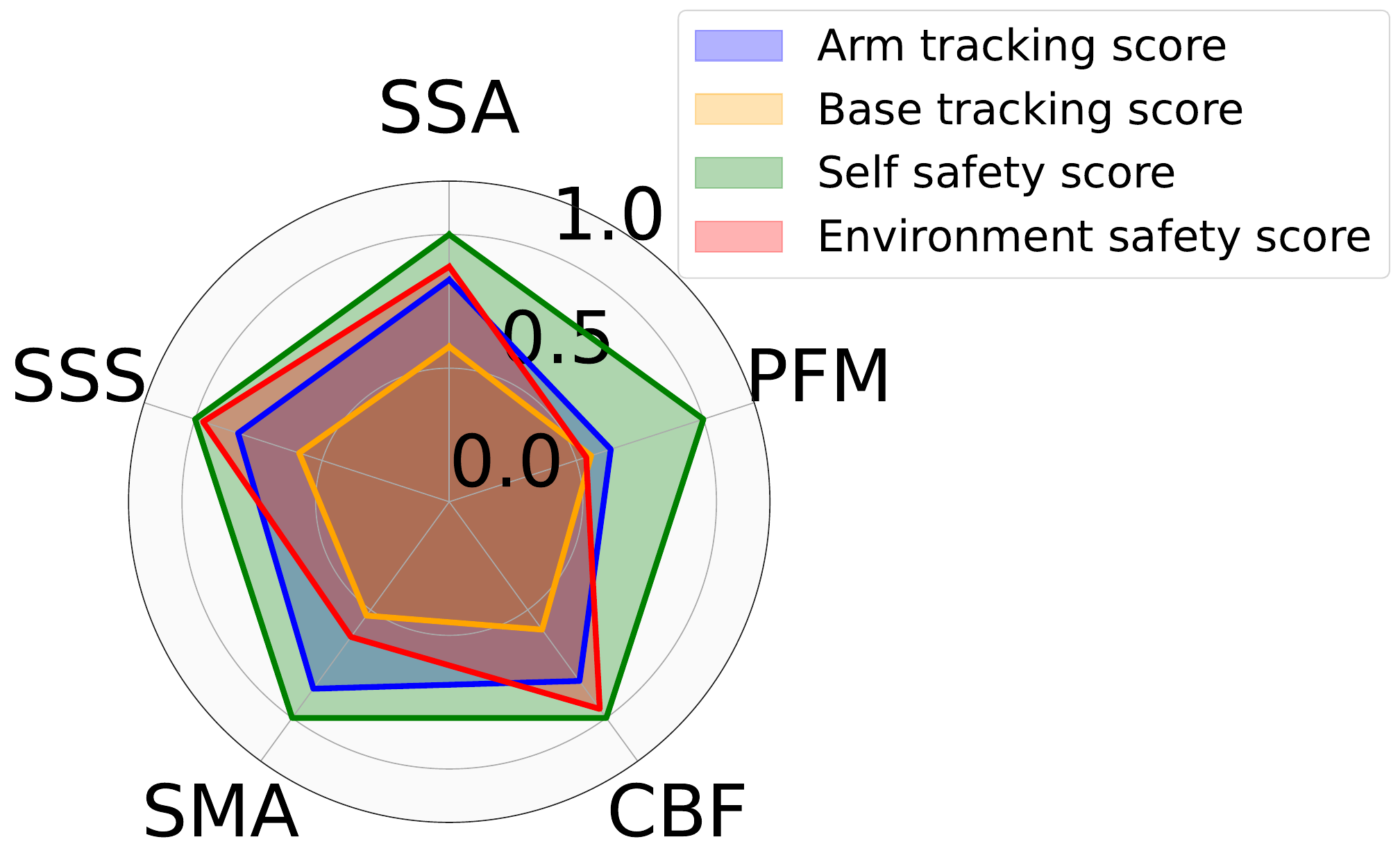}};
        \node[below, align=center, font=\small] at (\xgap-0.5, \ygap-1.5)
        {(d) G1MobileBase\_D1\\\_WG\_SO\_v1};

    \end{tikzpicture}

    \caption{Performance comparison of the benchmark.}
    \label{fig: benchmark_performance_fixed}
\end{figure}
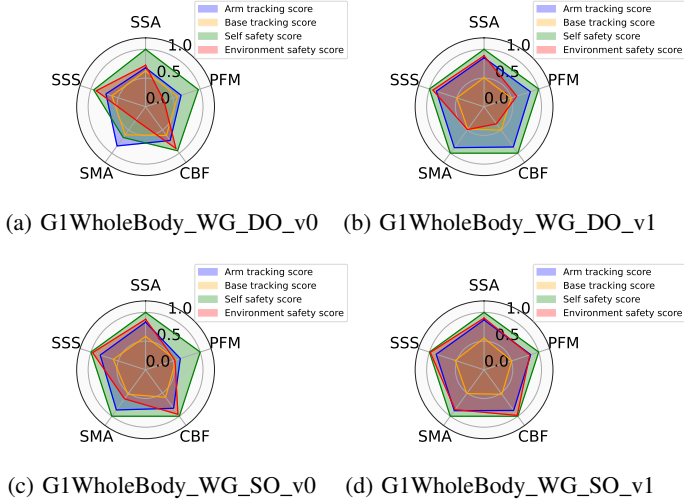

As shown in \Cref{fig: benchmark_performance_fixed}, obstacle motion and obstacle number both significantly impact algorithm performance. Dynamic obstacles introduce greater safety challenges compared to static ones. The radar charts illustrate that when obstacles are in motion, the corresponding \textbf{Arm Tracking Score} and \textbf{Environment Safety Score} tend to decrease. Furthermore, a comparison between v0 and v1 reveals that v1, which has fewer obstacles, achieves higher \textbf{Arm Tracking Score} and \textbf{Environment Safety Score}. This observation underscores the fact that a greater number of obstacles exerts a more pronounced negative impact on performance.

The other benchmark performance comparison is provided in \Cref{app: benchmark_performance_app} and \Cref{tab:alg_metric_task}.

\subsection{Influence of Task Difficulty}  
From \Cref{fig: param_tuning_subset}, it is evident that task difficulty significantly affects performance. For instance, when the robot is required to track both arm and base goals using whole-body movement, its performance is lower compared to tasks where a fixed base allows it to focus solely on arm goal tracking. This is because tracking the robot's base goal may require sacrificing arm goal tracking to avoid collisions and reach the base target.

\subsection{Success Rate across various Algorithms}
In addition to evaluating average step-wise performance, it is crucial to assess whether the safe controllers can successfully complete tasks trajectory-wise. A trajectory is defined to be successful if the robot reaches the goal without collisions within a maximum number of steps. In this evaluation, 50 feasible environment settings were generated and each algorithm was tested under these settings. The maximum number of steps was set to 200.  
\input{table_template/tables/success_rate}
The results of each algorithm are listed in \Cref{tab: success_rate}, which shows the conditional success rate of the algorithms. Detailed results are provided in \Cref{appendix: success_rate}. From these results, it can be observed that SSS achieves the highest success rate for tasks involving whole-body motion, The second is CBF. For tasks with a fixed base, SMA attains the highest success rate, while SSS and SSA also perform well. This suggests that optimization-based methods, such as SSS and SSA, are better equipped to handle multiple constraints in complex environments. Conversely, SMA's safe correction is more efficient for relatively simple tasks. Among all algorithms, PFM exhibits the lowest success rate. 

\definecolor{myPurple}{rgb}{1.0, 0.0, 0.0}
\def\dataA{{
{1.0000, 0.5000, 0.5250, 0.9500, 0.9500},
{1.0000, 1.0000, 0.7500, 0.9500, 1.0000},
{0.9545, 0.6818, 1.0000, 0.9545, 0.9545},
{0.8085, 0.4043, 0.4468, 1.0000, 0.7872},
{0.9744, 0.5128, 0.5385, 0.9487, 1.0000}
}}

\def\dataB{{
{1.0000, 0.6977, 0.6279, 1.0000, 1.0000},
{0.9677, 1.0000, 0.7742, 0.9677, 0.9677},
{0.9643, 0.8571, 1.0000, 0.9643, 0.9643},
{0.9149, 0.6383, 0.5745, 1.0000, 0.9149},
{0.9773, 0.6818, 0.6136, 0.9773, 1.0000}
}}

\def\dataC{{
{1.0000, 0.4419, 0.7907, 0.9767, 1.0000},
{1.0000, 1.0000, 0.9474, 1.0000, 1.0000},
{1.0000, 0.5294, 1.0000, 0.9706, 1.0000},
{0.8750, 0.3958, 0.6875, 1.0000, 0.8958},
{0.9556, 0.4222, 0.7556, 0.9556, 1.0000}
}}

\def\dataD{{
{1.0000, 0.6200, 1.0000, 1.0000, 1.0000},
{1.0000, 1.0000, 1.0000, 1.0000, 1.0000},
{1.0000, 0.6200, 1.0000, 1.0000, 1.0000},
{1.0000, 0.6200, 1.0000, 1.0000, 1.0000},
{1.0000, 0.6200, 1.0000, 1.0000, 1.0000}
}}

\def\dataE{{
{1.0000, 0.6250, 0.9750, 0.5000, 0.9750},
{0.8621, 1.0000, 0.8966, 0.5862, 0.8966},
{0.8864, 0.5909, 1.0000, 0.4773, 0.9773},
{0.8696, 0.7391, 0.9130, 1.0000, 0.9130},
{0.8667, 0.5778, 0.9556, 0.4667, 1.0000}
}}

\def\dataF{{
{1.0000, 0.9592, 0.9184, 1.0000, 1.0000},
{0.9792, 1.0000, 0.9167, 0.9792, 1.0000},
{1.0000, 0.9778, 1.0000, 1.0000, 1.0000},
{1.0000, 0.9592, 0.9184, 1.0000, 1.0000},
{0.9800, 0.9600, 0.9000, 0.9800, 1.0000}
}}

\def\dataG{{
{1.0000, 0.5957, 0.9574, 0.6596, 0.9574},
{0.9333, 1.0000, 0.9333, 0.8000, 0.9333},
{0.9783, 0.6087, 1.0000, 0.6739, 1.0000},
{1.0000, 0.7742, 1.0000, 1.0000, 1.0000},
{0.9783, 0.6087, 1.0000, 0.6739, 1.0000}
}}

\def\dataH{{
{1.0000, 0.9592, 1.0000, 0.9796, 1.0000},
{1.0000, 1.0000, 1.0000, 1.0000, 1.0000},
{0.9800, 0.9400, 1.0000, 0.9600, 1.0000},
{1.0000, 0.9792, 1.0000, 1.0000, 1.0000},
{0.9800, 0.9400, 1.0000, 0.9600, 1.0000}
}}

\newcommand{\rowLabel}[1]{%
  \ifcase #1
   \strut SSS\or \strut SMA\or \strut CBF\or \strut PFM\or \strut SSA%
  \fi
}
\newcommand{\colLabel}[1]{%
  \ifcase #1
    \strut SSA\or \strut PFM\or \strut CBF\or \strut SMA\or \strut SSS%
  \fi
}

\newcommand{\drawHeatMap}[1]{%
    \begin{tikzpicture}[scale=0.6]
    \pgfmathsetmacro{\rows}{5}
    \pgfmathsetmacro{\cols}{5}
    \def\data{#1}

    \foreach \r in {1,...,\rows} {
        \foreach \c in {1,...,\cols} {
            \pgfmathparse{\data[\r-1][\c-1]}
            \let\val=\pgfmathresult

            \pgfmathparse{(\val - 0.85)/(1 - 0.85)*100}
            \pgfmathsetmacro{\colorValue}{min(100,max(0,\pgfmathresult))}

            \fill[
              myPurple!\colorValue!yellow,
              draw=black,
              thin
            ] (\c-1,\rows-\r) rectangle ++(1,1);
        }
    }
    \draw[thick] (0,0) rectangle (\cols,\rows);

    \foreach \r in {1,...,\rows} {
        \pgfmathtruncatemacro{\idx}{\r-1} 
        \node[left] at (-0.2,\r-0.5) {\scriptsize  \rowLabel{\idx}};
    }

    \foreach \c in {1,...,\cols} {
        \pgfmathtruncatemacro{\idx}{\c - 1} 
        \node[above] at (\c-0.5,\rows+0.1) {\scriptsize  \colLabel{\idx}};
    }
    \end{tikzpicture}%
}
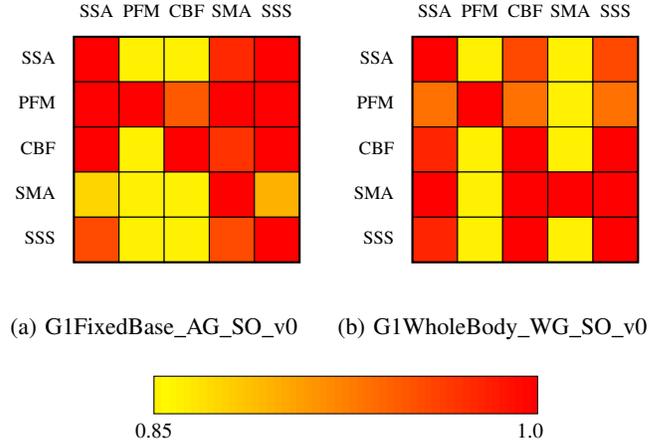
\begin{figure}[htbp]
    \centering
    \begin{tikzpicture} 
        \def\imgwidth{3cm}  
        \def\xgap{4.5}  
        \def\ygap{-2.3} 

        \node at (0, 0) {\drawHeatMap{\dataC}};
        \node[below] at (0, -2.4) {\small (a) G1FixedBase\_D1\_AG\_SO\_v1};

        \node at (\xgap, 0) {\drawHeatMap{\dataG}};
        \node[below] at (\xgap, -2.4) {\small (b) G1MobileBase\_D1\_WG\_SO\_v1};
        
        \def\barWidth{5.0}
        \def\barHeight{0.5}
        \def\steps{50}
    
        \foreach \s in {0,...,\steps} {
            \pgfmathsetmacro{\f}{\s/\steps}  
            \pgfmathparse{100*\f}            
            \let\colorValue=\pgfmathresult
    
            \pgfmathsetmacro{\xpos}{\f*\barWidth}
    
            \fill[myPurple!\colorValue!yellow,draw=none]
                 (\xpos, \ygap-1.5) rectangle
                 (\xpos + \barWidth/\steps, \barHeight+\ygap-1.5);
        }
    
        \draw[black] (0, \ygap-1.5) rectangle (\barWidth+0.1,\barHeight+\ygap-1.5);
    
        \node[below] at (0, \ygap-1.5) {\footnotesize 0.85};
        \node[below] at (\barWidth, \ygap-1.5) {\footnotesize 1.0};
    \end{tikzpicture}

    \caption{Conditional success plots for selected tasks.}
    \label{fig: conditional_success_selected}
\end{figure}

Beyond individual success rates, it is also informative to examine the relative advantages between pairs of safe control algorithms. To do this, we define the conditional success rate \( P(B|A) \) as:
\begin{align}
    P(B|A) = \frac{\#(A \cap B)}{\#A}, \\ \nonumber
\end{align}
where \( \#(B \cap A) \) represents the number of environments successfully completed by both algorithm \( A \) and algorithm \( B \), and \( \#A \) represents the number of environments successfully completed only by algorithm \( A \).  

\Cref{fig: conditional_success_selected} presents a heatmap of the conditional success rates for each task where the value of the cell $(i, j)$ is calculated by $P(algo[j]|algo[i])$. The results indicate that optimization-based methods, such as SSA, SSS, and CBF, significantly outperform PFM and SMA in success rates. Furthermore, the comparison of conditional success rates reveals that \( P(\text{SSA}|\text{CBF}) > P(\text{CBF}|\text{SSA}) \) in five out of eight tasks, indicating that SSA successfully completes most of the trajectories that CBF can achieve in these cases. In contrast, CBF does not achieve the same success rate. Detailed conditional success rates are reported in \Cref{appendix: conditional_success_rate} and \Cref{app: conditional_success_app}.

In conclusion, using \spark as a safe control benchmark highlights its composability by enabling the convenient generation of large-scale testing configurations. As a benchmarking framework, it also provides users with a structured parameter-tuning process and a comprehensive understanding of the implemented safe control algorithms, aiding in the synthesis of the most suitable safe controller for each task. The results indicate that there is still significant room for improvement in safe controller design to achieve a fully guaranteed $100\%$ safety assurance, presenting further challenges for future research.

%% file: table_template/tables/success_rate.tex
\begin{table}[h]
\centering
\captionsetup{width=0.5\textwidth}
\caption{Success rate for Different Tasks and Algorithms}
\label{tb:success_rate}
\scalebox{0.9}{
\begin{tabular}{c|ccccc}
\toprule
\textbf{Task Name} & \textbf{SSA} & \textbf{PFM} & \textbf{CBF} & \textbf{SMA} & \textbf{SSS} \\
\midrule

G1FixedBase\_D1\_AG\_SO\_v0 &  \textbf{1.0000}& 0.6667 &  \textbf{1.0000}& \textbf{1.0000} &\textbf{1.0000} \\
G1FixedBase\_D1\_AG\_SO\_v1 & 0.8667 & 0.4167 & 0.7000 & \textbf{0.9500} & 0.9000 \\
G1FixedBase\_D1\_AG\_DO\_v0 & 0.8833 & 0.6333 & 0.5500 & \textbf{0.9500} & 0.9000 \\
G1FixedBase\_D1\_AG\_DO\_v1 & 0.8167 & 0.4167 & 0.4833 & \textbf{0.9500} & 0.8000 \\
G1MobileBase\_D1\_WG\_SO\_v0 & 0.9667 & 0.9500 & \textbf{1.0000} & 0.9667 & \textbf{1.0000} \\
G1MobileBase\_D1\_WG\_SO\_v1 & \textbf{0.9333} & 0.5667 & \textbf{0.9333} & 0.6333 & \textbf{0.9333} \\
G1MobileBase\_D1\_WG\_DO\_v0 & 0.9833 & 0.9500 & 0.8833 & 0.9833 &  \textbf{1.0000} \\
G1MobileBase\_D1\_WG\_DO\_v1 & 0.7833 & 0.5667 & \textbf{0.8500} & 0.4667 & \textbf{0.8500} \\
\bottomrule
\end{tabular}
}
\label{tab: success_rate}
\end{table}

%% file: sections/8_usecase_simulation_teleoperation.tex
\section{Use Case 2: \\ Safe Teleoperation With Simulated Robot}\label{sec: usecase_safe_teleop_sim}

\begin{figure*}[htbp]
    \centering
    \vspace{1.5cm}  
    \begin{tikzpicture}[transform canvas={xshift=0cm}]
        \def\imgwidth{15cm}  
        \def\ygap{-4} 

        \node at (0, 0) {\includegraphics[width=\imgwidth]{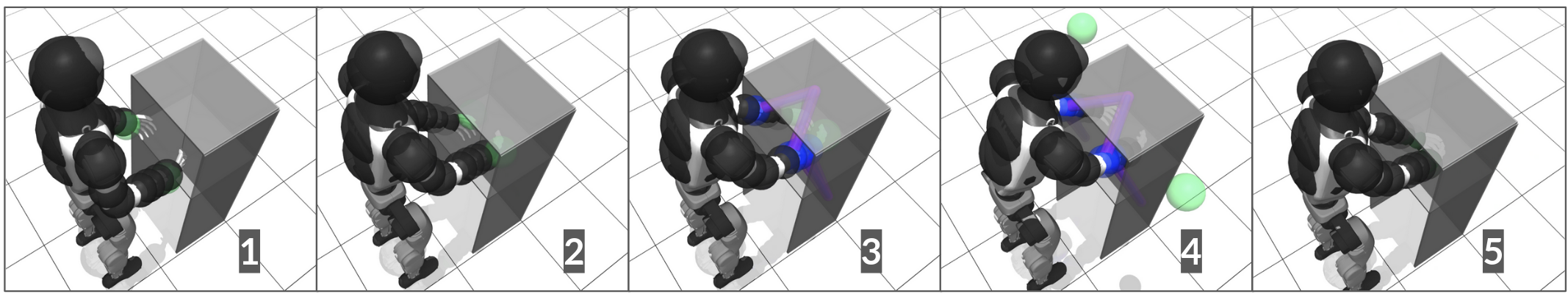}};

    \end{tikzpicture}
    \vspace{1.5cm}
    \caption{\ref{sec: usecase_safe_teleop_sim} The first two illustrate how the robot’s hands successfully reach into a confined cabinet under user teleoperation. In the fourth figure, even when the green spheres—representing the target positions for teleoperation—move outside the cabinet, the robot’s hands stay within the safe region. This demonstrates the controller’s ability to enforce safety constraints while following user commands.}
    \label{fig: safe_teleop_sim}
\end{figure*}

\begin{figure*}[htbp]
    \centering
    \vspace{1.5cm}  
    \begin{tikzpicture}[transform canvas={xshift=0cm}] 
        \def\imgwidth{15cm}  
        \def\ygap{-3} 

        \node at (0, 0) {\includegraphics[width=\imgwidth]{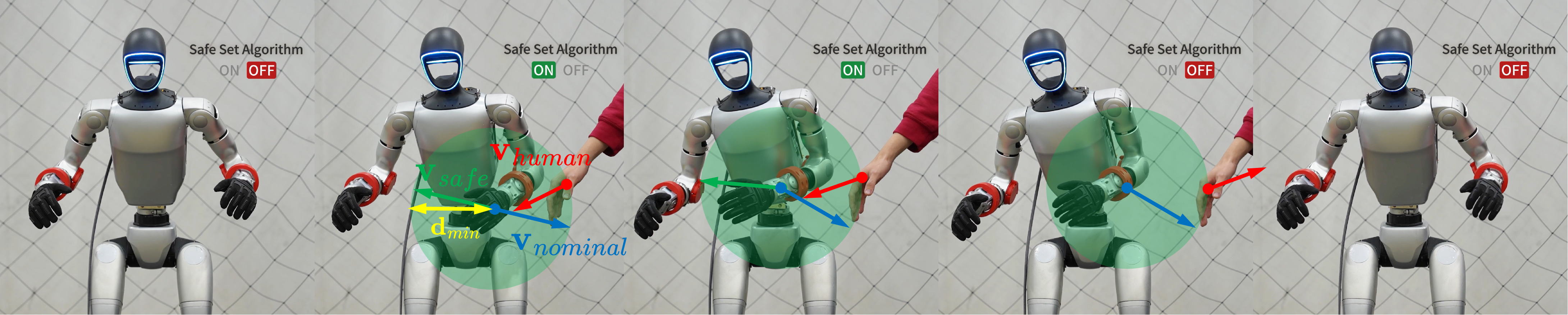}};

        \node at (0, \ygap) {\includegraphics[width=\imgwidth]{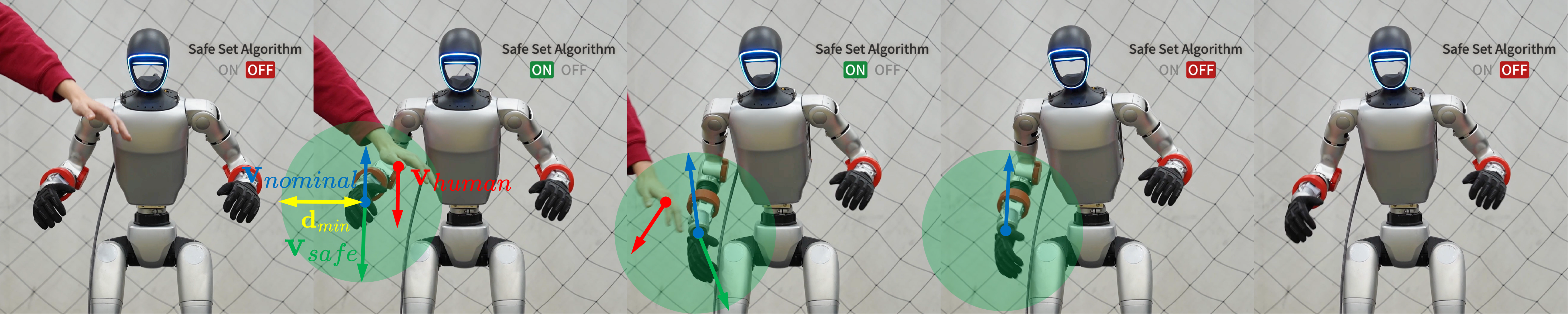}};

    \end{tikzpicture}
    \vspace{4.5cm}
    \caption{\ref{sec: usecase_safe_auto_real} Limb-level collision avoidance with static humanoid reference pose: the robot moves away when the human hand gets closer than \( d_{\min} \) and resumes its target position once the hand retreats and the environment is safe.
}
    \label{Static}
\end{figure*}

\begin{figure*}[htbp]
    \centering
    \vspace{1.5cm}  
    \begin{tikzpicture}[transform canvas={xshift=0cm}]
        \def\imgwidth{15cm}  
        \def\ygap{-4} 

        \node at (0, 0) {\includegraphics[width=\imgwidth]{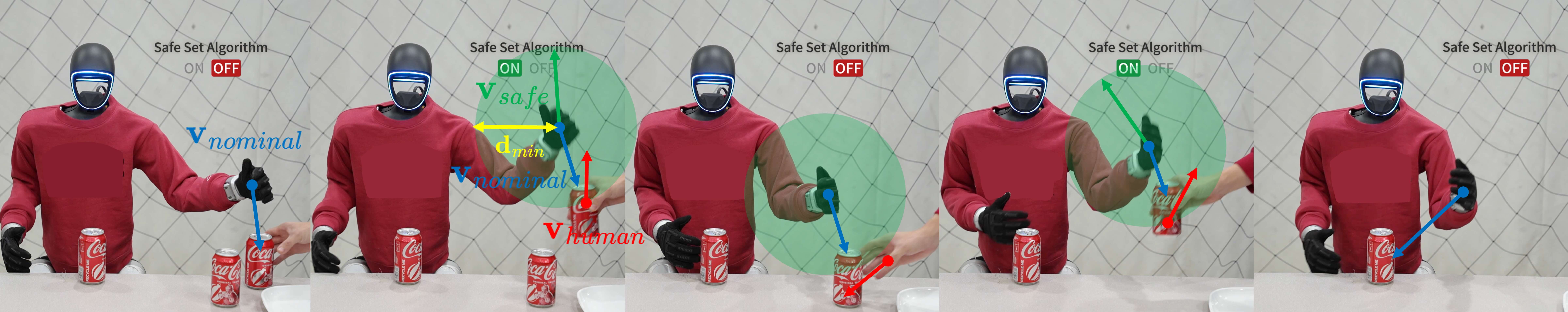}{}};
        \node at (0, -3)
        {\includegraphics[width=\imgwidth]{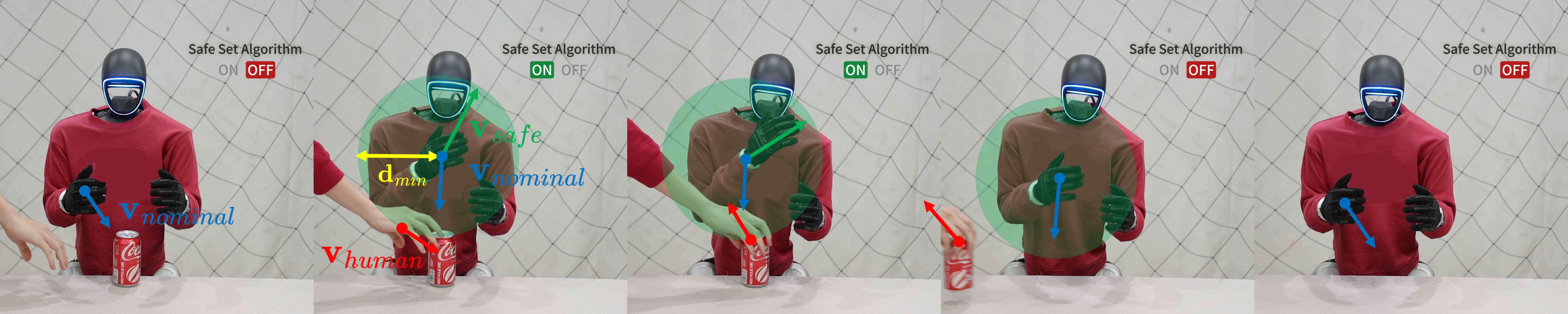}{}};

    \end{tikzpicture}
    \vspace{4.5cm}
    \caption{\ref{sec: usecase_safe_teleop_real} Limb-level collision avoidance with teleoperation commands: if the human user reaches for the same object as the robot, the safe controller is triggered, prioritizing collision avoidance over teleoperation commands to ensure safe interaction and prevent hazards from limited remote perception.
}
    \label{Teleop}
\end{figure*}

This section presents a user scenario in which teleoperation is performed within a simulation environment to collect human data in the absence of available hardware~\cite{wang2024teleophri,nechyporenko2024armadaaugmentedrealityrobot,chen2024arcapcollectinghighqualityhuman}. 

To meet this requirement, we configure the robot with \texttt{G1fixedBase} while selecting the simulation agent in \spark. 
By designing a \textbf{Task} module where a cabinet acts as an obstacle and the human teleoperation serves as the task goal—while taking human input through an external Apple Vision Pro block—we can retain the same \textbf{Policy} and \textbf{Safety} modules as in other use cases.  

From the first two plots in \Cref{fig: safe_teleop_sim}, we observe that the robot’s hands can maneuver into a confined cabinet under user teleoperation. 
In the fourth plot, we see that when the green spheres—representing the teleoperation target positions for the hands—move outside the cabinet, the robot's hands remain inside to ensure safety, demonstrating the effectiveness of the safe controller.  

This scenario highlights the flexibility of \spark in integrating human input with simulated robots. 
It not only guarantees absolute safety for algorithm verification but also provides users with richer visual feedback that cannot be directly obtained from real robot teleoperation, such as the spatial relationship between the robot's hands and obstacles. 
Moreover, it offers an intuitive and efficient way to collect human demonstration data without relying on real robot hardware.  

%% file: sections/6_usecase_hardware_autonomy.tex
\section{Use Case 3: Safe Autonomy on Real Robot}\label{sec: usecase_safe_auto_real}



To showcase the practical applicability of the \spark framework, we deployed its safe control pipelines on a real robot.
For our experiments, we utilized the Unitree G1 humanoid robot, which features 29 degrees of freedom (DOFs).
The perception module incorporated an Apple Vision Pro to capture human gestures and determine the robot’s position.
The robot’s dynamic system was modeled as a mobile dual-manipulator system.

Building on the benchmark use case introduced in \Cref{sec: usecase_benchmark}, we only need to replace the agent module with the real G1 SDK and integrate the Apple Vision Pro interface into the task module.
With these modifications, the control algorithms validated in the benchmark can be directly deployed on the real robot.

We begin with the simplest task, in which the robot remains in a fixed position while avoiding potential collisions with the human user. 
As shown in Figure~\ref{Static}, when the user attempts to approach the robot arm from various angles, the robot reacts by moving away from the human hand if the minimum distance between them becomes smaller than $d_{min}$. Once the robot detects that the human hand has moved away and the surrounding environment is safe, it resumes following the nominal controller's inputs, which commands it to remain in the target static position.

We further evaluate the performance of the \spark safe controller in tracking a dynamic target position by designing a dynamic limb-level collision avoidance test case.
Unlike the static test, where the nominal controller simply tracks a fixed target, the nominal controller in the dynamic test is tasked with following a dynamic target $\xb^R_{target}$.
Specifically, the nominal controller tracks a circular trajectory for the right hand.
In this case, the robot must track the target trajectory while simultaneously avoiding collisions with the human user using the same safety index $\phi$ as previously defined.

\begin{figure*}[htbp!]
    \centering
    \vspace{1.5cm}  
    \begin{tikzpicture}[transform canvas={xshift=0cm}]
        \def\imgwidth{15cm}  
        \def\ygap{-3} 

        \node at (0, 0) {\includegraphics[width=\imgwidth]{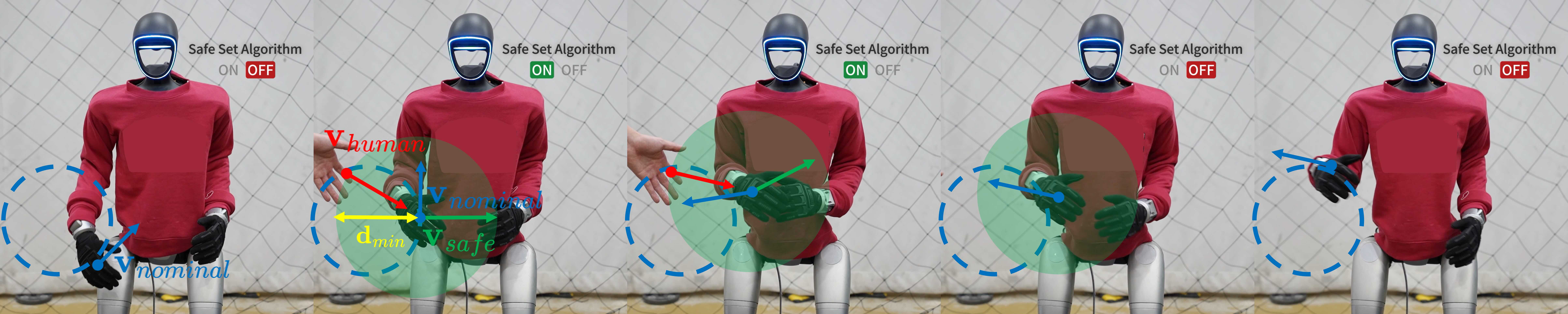}};

        \node at (0, \ygap) {\includegraphics[width=\imgwidth]{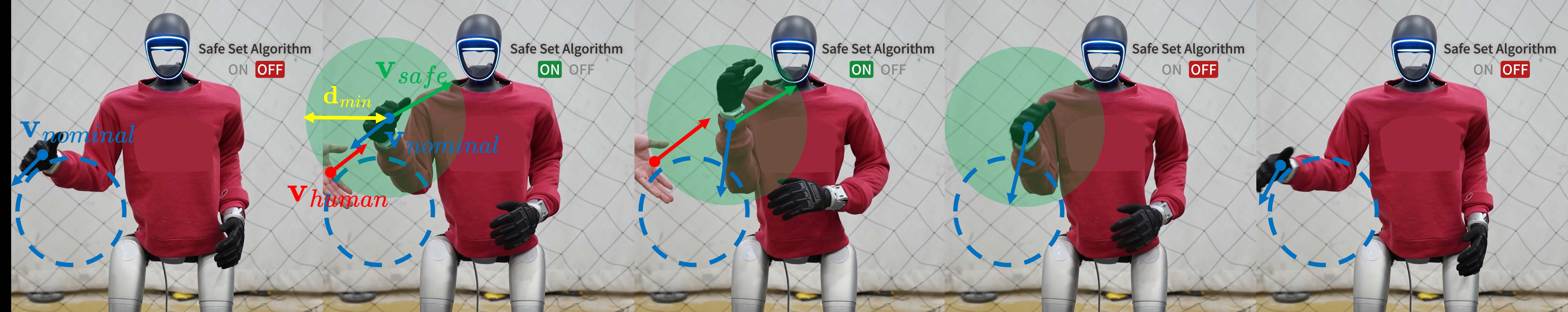}};

    \end{tikzpicture}
    \vspace{4.5cm}
    \caption{\ref{sec: usecase_safe_auto_real} Limb-level collision avoidance with dynamic humanoid reference poses: when the human hand stays outside \( d_{min} \), the robot follows the reference trajectory. If it gets too close, the humanoid adjusts its waist and arms to avoid collision, prioritizing safety. Once the hand moves away, the robot resumes its trajectory while remaining prepared.  
 }
    \label{Dynamic}
\end{figure*}

From Figure~\ref{Dynamic}, we observe that when the human hand remains outside the $d_{min}$ region, the robot follows the reference trajectory and moves along the circular path. 
If a human hand gets too close to the robot's hands, the humanoid will use both its waist and arm movements to avoid a collision, regardless of the target's motion. In other words, the safety constraint takes precedence over the normal control input, ensuring the humanoid remains safe.
Once the humans hand moves away, the robot returns to the reference trajectory while remaining prepared for potential collision avoidance.

%% file: sections/7_usecase_hardware_teleoperation.tex
\section{Use Case 4: \\ Safe Teleoperation With Real Robot}\label{sec: usecase_safe_teleop_real}

As teleoperation becomes more widely used to control humanoid robots~\cite{goodrich2013teleop,kourosh2023teleop}, it is paramount to ensure safety in this operating mode.
Thus, this section assesses the safe controller's performance in more general scenarios, in which the humanoid robot is tasked with following a human user's teleoperation commands while ensuring collision avoidance. This setup introduces the concept of ``Safe Teleoperation."

In this test, the nominal humanoid controller's target, $\xb^R_{target}$, is not pre-designed but is instead generated in real-time by the human teleoperator. This adds complexity, as the robot must generate safe motions in an unpredictable and dynamic environment.

To implement this, we only needed to make a simple modification to the task module of \spark, adding the operator's hand positions as the goal positions for the robot's arms. Meanwhile, other human participants were treated as obstacles to ensure safe human-robot interaction.

In our experiment, we created a realistic scenario where the robot attempts to retrieve objects from a table. 
\Cref{Teleop} shows if the human user reaches for the same object as the robot, the safe controller is triggered. 
The robot prioritizes collision avoidance over executing the teleoperation commands, ensuring safe interaction and protecting both the humanoid robot and the human from potential hazards caused by the limited perception of a remote teleoperator.

To provide a more comprehensive statistical evaluation, we conducted supplementary experiments on a real robot and a simulation robot under the same dynamic environment with the SSA method:  

\begin{enumerate}  
    \item \textbf{Obstacle}: Human hands act as dynamic obstacles while the robot hands attempt to follow dynamic goals moving along a circular trajectory. (Use Case 3).
    \item \textbf{Teleop}: Human hands move randomly as the goal for teleoperation, while two obstacles moves along a circular trajectory (Use Case 4).
\end{enumerate}  

300 trajectories with 100 time steps were collected. Additionally, the Pearson correlation and p-value were calculated to compare the performance of the simulated robot with the real robot (\Cref{tab:metrics_comparison}).  

\begin{table}[h]
    \caption{Comparison of key metrics between simulation and real robot.}
    \centering
    \begin{tabular}{p{0.8cm}p{1.0cm}p{1.6cm}p{1.6cm}p{1.95cm}}
        \toprule
        Metric & Exp & \makecell{Sim \\Mean $\pm$ SD} & \makecell{Real \\ Mean $\pm$ SD} & \makecell{Pearson r \\ (p-value)} \\
        \midrule
        \multirow{2}{*}{\makecell{Goal\_Dist \\ (m)}}
        & Teleop\ & $0.083 \pm 0.066$ & $0.086 \pm 0.073$ & $0.743$ ($<0.001$) \\
        & Autonomy & $0.102 \pm 0.032$ & $0.042 \pm 0.030$ & $0.538$ ($<0.001$) \\
        \midrule
        \multirow{2}{*}{\makecell{Self\_Dist \\ (m)}} 
        & Teleop\ & $0.237 \pm 0.022$ & $0.242 \pm 0.024$ & $0.706$ ($<0.001$) \\
        & Autonomy & $0.241 \pm 0.018$ & $0.241 \pm 0.022$ & $0.712$ ($<0.001$) \\
        \midrule
        \multirow{2}{*}{\makecell{Env\_Dist \\ (m)}} 
        & Teleop\ & $0.297 \pm 0.042$ & $0.299 \pm 0.046$ & $0.948$ ($<0.001$) \\
        & Autonomy & $0.310 \pm 0.038$ & $0.300 \pm 0.039$ & $0.969$ ($<0.001$) \\
        \bottomrule
    \end{tabular}
    \label{tab:metrics_comparison}
    \vspace{-10pt}
\end{table}
It illustrates the safe controller achieves performance equivalent to that of the simulation robot under the same safe autonomy and teleoperation scenarios with small p-values.

\paragraph{Computational Costs and Feasibility}  
In \Cref{tab:time_metrics_comparison}, \textbf{Loop Time} represents the computation time of each SPARK control loop in the supplementary experiments. \textbf{Trigger Safe Time} indicates the computation cost when the SSA safe controller is activated, while \textbf{No Trigger Safe} represents the opposite case. The SPARK runtime frequency reaches approximately 100 Hz without the safe controller and around 70 Hz when the safe controller is activated.
\begin{table}[h]
    \centering
    \caption{Runtime metrics of SPARK controller in simulation and real-world under different modes.}
    \begin{tabular}{c c c c}
        \toprule
        Metric & Exp & Sim Mean $\pm$ SD & Real Mean $\pm$ SD \\
        \midrule
        \multirow{2}{*}{Loop Time(s)} 
        & Teleop & $0.013 \pm 0.003$ & $0.011 \pm 0.007$ \\
        & Autonomy & $0.013 \pm 0.002$ & $0.008 \pm 0.003$ \\
        \midrule
        \multirow{2}{*}{Trigger Safe(s)} 
        & Teleop & $0.017 \pm 0.003$ & $0.015 \pm 0.007$ \\
        & Autonomy & $0.017 \pm 0.003$ & $0.011 \pm 0.002$ \\
        \midrule
        \multirow{2}{*}{No Trigger Safe(s)} 
        & Teleop & $0.012 \pm 0.002$ & $0.009 \pm 0.004$ \\
        & Autonomy & $0.012 \pm 0.002$ & $0.006 \pm 0.001$ \\
        \bottomrule
    \end{tabular}
    \label{tab:time_metrics_comparison}
    \vspace{-10pt}
\end{table}

The runtime system operates on a laptop equipped with a 12th Gen Intel i9-12900HK \(\times 20\) CPU, 31GB of memory, and running Ubuntu 20.04. No GPU is used in SPARK.

%% file: sections/9_conclusion.tex
\section{Limitations}
Aiming to be a general and user-friendly benchmark, \spark has several potential limitations slated for future improvements.


The current implementation does not distinguish between inevitable collisions from method failures (i.e., there are feasible collision-free trajectories but the method could not find one). As described in \cite{chen2025dexterous}, method failures happen when multiple safety constraints are in conflict, which does not necessarily imply a collision is inevitable. To mitigate this issue, more research is needed to either improve the safe control methods in handling multiple constraints or introduce advanced methods in detecting inevitable collisions. 

Another limitation is that the sim to real gap also exists in model-based control systems, although it is called differently as ``model mismatch". For the real deployment, the robot trajectory might be different from the simulation due to ``model mismatch". To mitigate this problem, the system model needs to be robustified and the system control needs to be aware of potential gaps, which will be left for future work.




\section{Conclusion and Future Work}
In this paper, we presented \spark, a comprehensive benchmark designed to enhance the safety of humanoid autonomy and teleoperation. We introduced a safe humanoid control framework and detailed the core safe control algorithms upon which \spark is built.

\spark offers configurable trade-offs between safety and performance, allowing it to meet diverse user requirements. Its modular design, coupled with accessible APIs, ensures compatibility with a wide range of tasks, hardware systems, and customization levels. Additionally, \spark includes a simulation environment featuring a variety of humanoid safe control tasks that serve as benchmark baselines. By utilizing \spark, researchers and practitioners can accelerate humanoid robotics development while ensuring robust hardware and environmental safety.

Beyond what was mentioned in the previous section, there are several future directions for \spark that could benefit from community collaboration. 
First of all, it is important to further lower the barrier for users to adopt state-of-the-art safety measures \cite{ji2023safety}, e.g., safe reinforcement learning approaches \cite{zhao2023absolute}\cite{zhao2023learn}\cite{zhao2023state} \cite{yao2024constraint}, by integrating them into \spark algorithm modules. 
Secondly, to enhance the robustness and reliability of the deployment pipeline of \spark, future works are needed to extend \spark compatibility with standardized safety assessment pipelines, such as stress testing tools and formal verification tools \cite{xie2024framework}. 
Thirdly, simplifying task specification is critical for usability. Enhancements to \spark's interface, such as intuitive configuration, natural language-based task definitions \cite{lin2023text2motion}, and interactive visualizations \cite{park2024dexhub}, will enable more efficient safety policy design and debugging. 
Finally, automating the selection and tuning \cite{victoria2021automatic} of safety strategies will make \spark more adaptive. Future efforts may explore automatic model-based synthesis approaches such as meta-control \cite{wei2024metacontrol} techniques to dynamically optimize safety measures based on task requirements and environmental conditions. 

%% file: sections/10_acknowledgement.tex
\section{Acknowledgement}

This work is supported by the National Science Foundation under grant No. 2144489.

%% file: sections/appendix.tex
\newpage
\onecolumn
\section{Appendix}

\subsection{Software framework}
\label{appendix: software_framework}
\begin{figure*}[htbp]
    \centering
    \vspace{2cm}  
    \begin{tikzpicture}[transform canvas={xshift=0cm}]
        \def\imgwidth{15cm}  

        \node at (0, 0) {\includegraphics[width=\imgwidth]{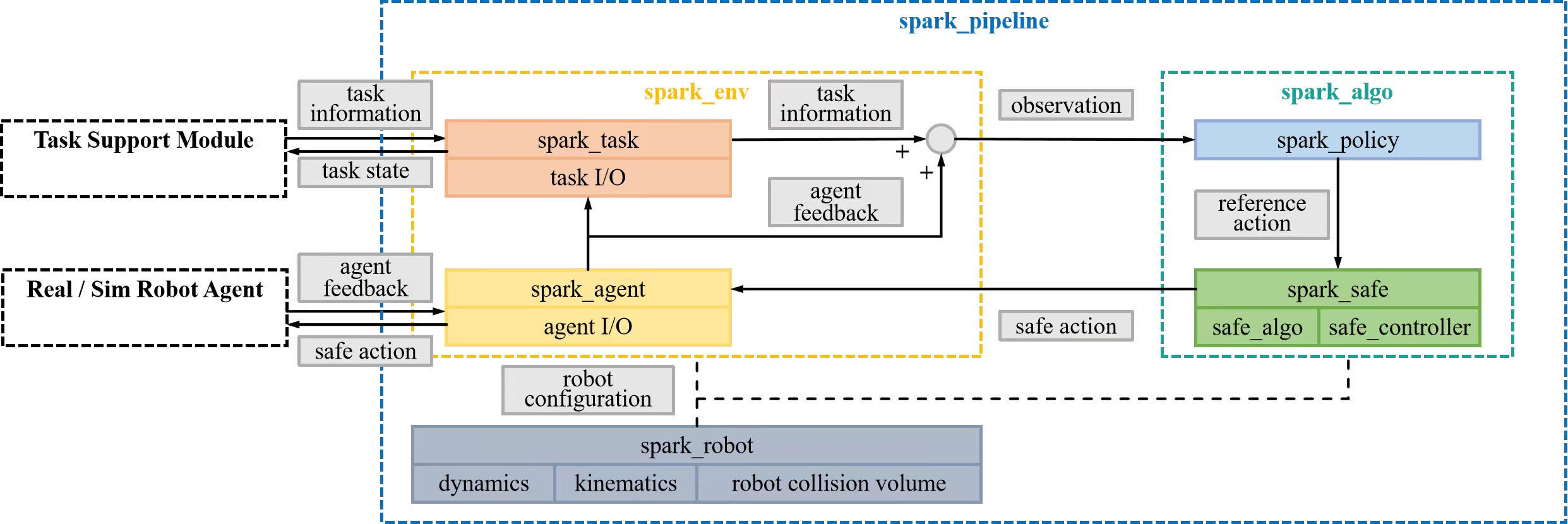}};
    \end{tikzpicture}
    \vspace{2.5cm}
    \caption{\spark system framework.}
    \label{fig:system}
\end{figure*}

\subsection{Python Example}
\label{appendix: python_example}
\input{sections/python_API}
\subsection{Robot frame}
\label{appendix: frame}
\input{sections/frame}
\clearpage
\subsection{Robot configuration}
\label{appendix: robot_config}
\input{table_template/tables/robot_config}

\clearpage
\subsection{Collision volume configuration}
\label{appendix: collision_config}
\input{table_template/tables/frame_config}
\clearpage
\subsection{Time variation of the safety index}
\label{appendix: safety_index_intro}
\input{sections/safety_index_intro}
\subsection{Task configuration}
\label{appendix: task_config}
\input{table_template/tables/task_config}
\subsection{Self-collision configuration}
\label{appendix: self_collision}
\input{table_template/tables/self_collision_pair}

\clearpage

\subsection{Trade-off curves between safety and efficiency}
\label{app: param_tuning_app}
\input{sections/param_tuning}

\subsection{Algorithm hyperparameters}
\label{appendix: hyperparameters}
\input{table_template/tables/hyperparameter}

\clearpage

\subsection{Performance comparison of the benchmark}
\label{app: benchmark_performance_app}
\input{sections/full_evaluation}

\clearpage
\subsection{Success rate}
\label{appendix: success_rate}

\input{sections/success_matrix}

\clearpage
\subsection{Algorithm comparison}
\label{appendix: conditional_success_rate}

\input{table_template/tables/algorithm_compare}
\clearpage

\clearpage
\subsection{Conditional success plots}
\label{app: conditional_success_app}
\input{sections/conditional_success_matrix}

%% file: sections/Python_API.tex
\subsubsection{Configure a pipeline}

To create a pipeline based on the benchmark configuration, you can follow the example below to test the Control Barrier Function (CBF) on a fixed-base humanoid robot:

\begin{center}
\begin{lstlisting}[language=Python]
from spark_pipeline import G1BenchmarkPipeline as Pipeline
from spark_pipeline import G1BenchmarkPipelineConfig as PipelineConfig
from spark_pipeline import generate_g1_benchmark_test_case
# Initialize the configuration
cfg = PipelineConfig()

# Load a predefined configuration for a fixed-base robot with first-order dynamics,
# targeting arm goals amidst dynamic obstacles
cfg = generate_g1_benchmark_test_case(cfg, "G1FixedBase_D1_AG_DO_v0")

# Set the safety algorithm to Control Barrier Function and configure its parameters
cfg.algo.safe_controller.safe_algo.class_name = "BasicControlBarrierFunction"
cfg.algo.safe_controller.safe_algo.lambda_cbf = 1.0
cfg.algo.safe_controller.safe_algo.control_weight = [
    1.0, 1.0, 1.0,  # Waist joints
    1.0, 1.0, 1.0, 1.0, 1.0, 1.0, 1.0,  # Left arm joints
    1.0, 1.0, 1.0, 1.0, 1.0, 1.0, 1.0,  # Right arm joints
    # Locomotion joints are omitted for fixed-base configuration
]

# Define the minimum distance to maintain from the environment
cfg.algo.safe_controller.safety_index.min_distance["environment"] = 0.1

# Select metrics to log during execution
cfg.metric_selection.dist_robot_to_env = True

# Create the pipeline instance with the specified configuration
pipeline = Pipeline(cfg)

# Run the pipeline
pipeline.run()
\end{lstlisting}
\end{center}

\subsubsection{Defining a customized safety module}

For example, to implement a controller based on SSA that applies a user-defined safety index function and safely tracks Cartesian hand goals, the users may follow the steps:
\begin{itemize}
    \item \textbf{Create a custom safety index:}  
    \begin{itemize}
        \item Navigate to \texttt{module/spark\_safe/spark\_safe\\/safe\_algo/value\_based/base/}
        \item Create \texttt{safety\_index\_nn.py} based on \texttt{basic\_collision\_safety\_index.py}.
        \item Define the user designed safety index function \texttt{phi()} of the created script.
    \end{itemize}

    \item \textbf{Develop a SSA based controller:}  
    \begin{itemize}
        \item Navigate to \texttt{module/spark\_safe/spark\_safe\\/safe\_algo/value\_based/ssa/}
        \item Create \texttt{safe\_set\_algorithm\_cartesian.py} based on \texttt{safe\_set\_algorithm.py}.
        \item Modify the function \texttt{qp\_solve()} to incorporate the Cartesian error into the objective function.
    \end{itemize}
\end{itemize}

%% file: sections/frame.tex
\begin{figure}[h]
    \centering
    \begin{subfigure}{0.45\textwidth}
        \centering
        \includegraphics[width=\linewidth]{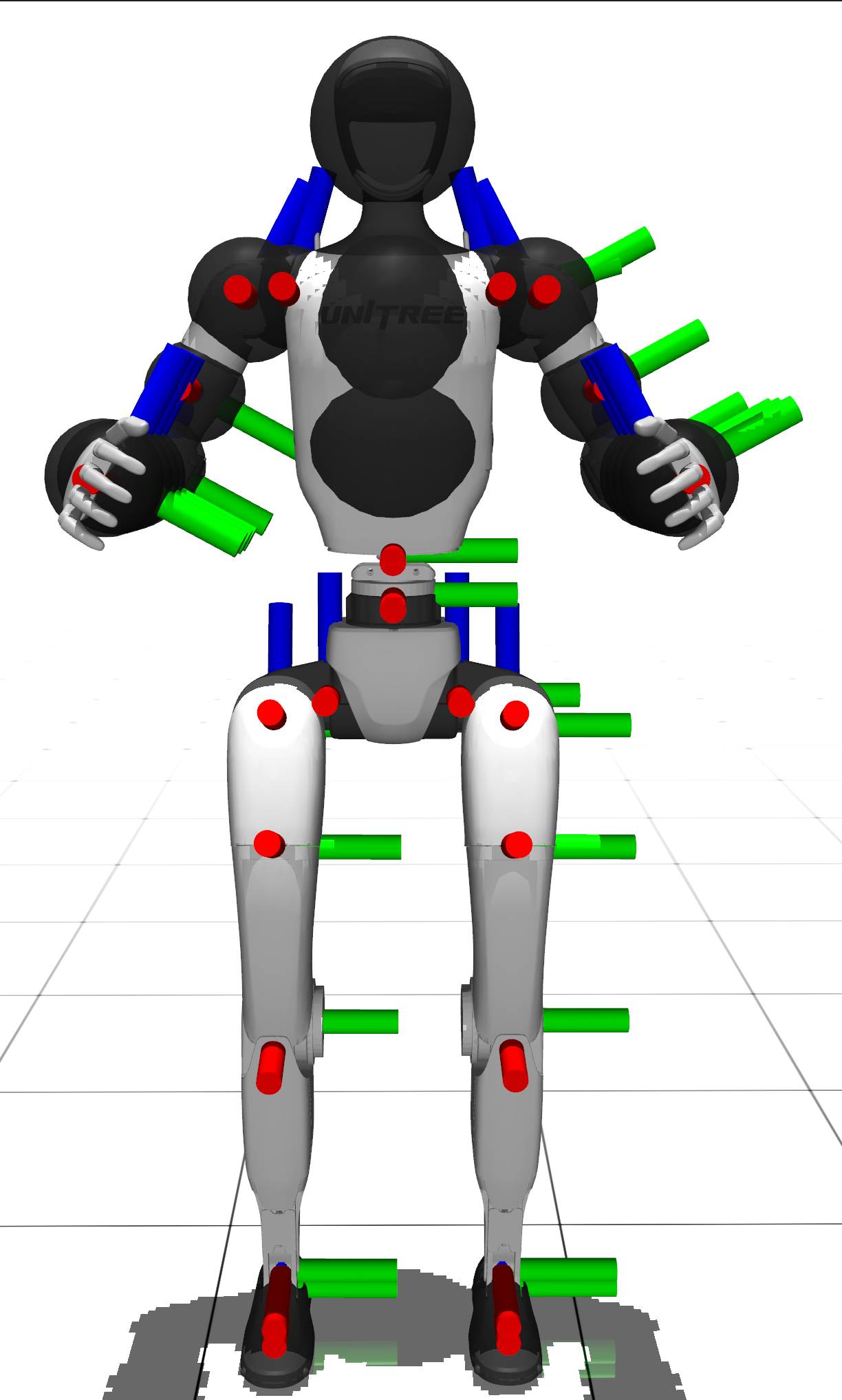}
        \caption{Robot frame coordination}
    \end{subfigure}
    \hfill
    \begin{subfigure}{0.45\textwidth}
        \centering
        \includegraphics[width=\linewidth]{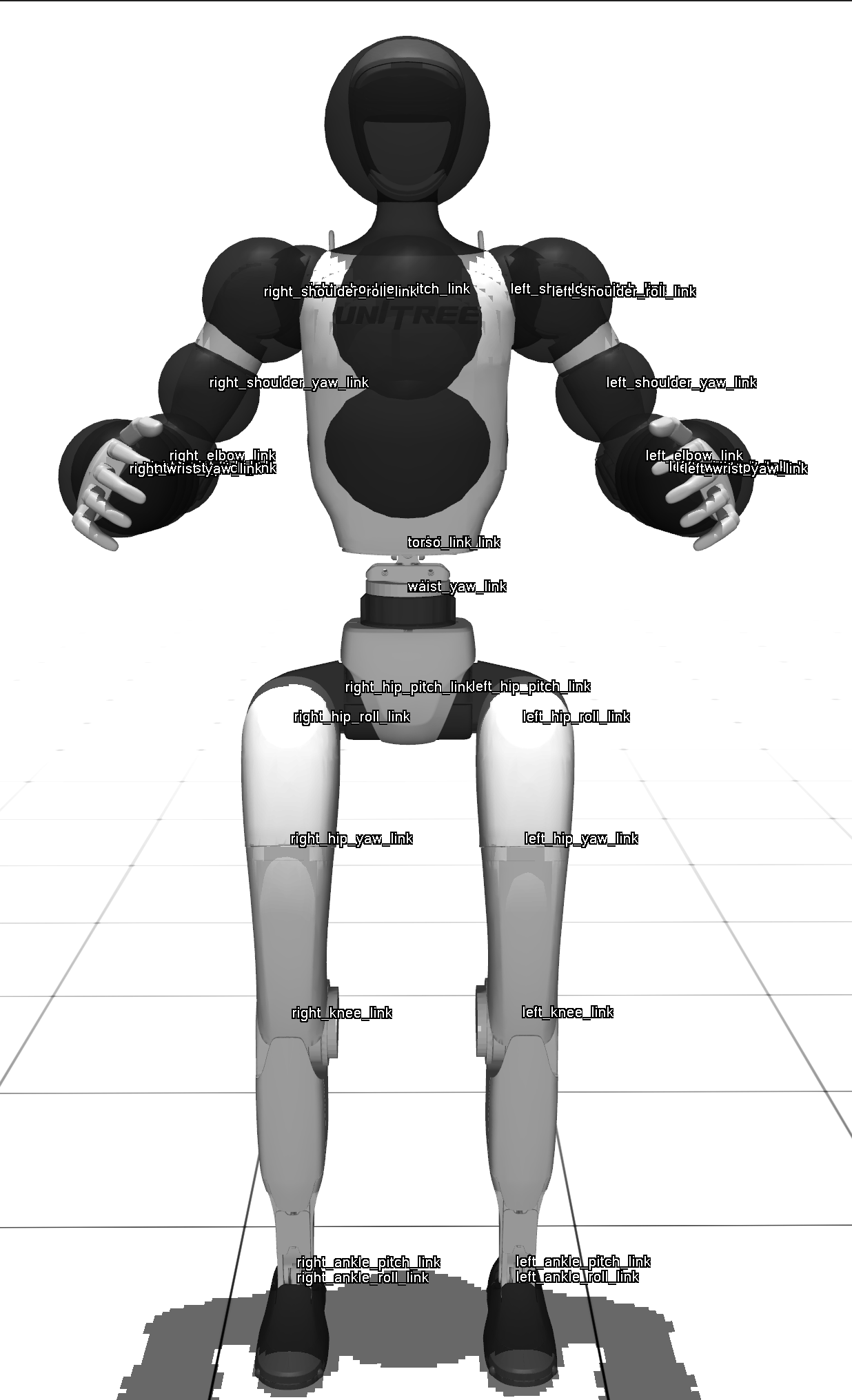}
        \caption{Robot frame name}
    \end{subfigure}
    \caption{Figure of robot frames}
    \label{fig:side_by_side}
\end{figure}

%% file: table_template/tables/robot_config.tex
\begin{table}[h]
\begin{center}
\captionsetup{width=\textwidth}
\caption{Robot Degree of Freedoms(DoFs)}
\label{tb: tasks}
\scalebox{1}{
\begin{tabular}{ccccc}
  \toprule
  Degree of Freedoms & RightArm & FixedBase & MobileBase & SportMode \\
  \midrule
WaistYaw & & \checkmark & \checkmark& \checkmark \\
WaistRoll & & \checkmark & \checkmark& \checkmark \\
WaistPitch & & \checkmark & \checkmark& \checkmark \\
LeftShoulderPitch & & \checkmark & \checkmark& \checkmark \\
LeftShoulderRoll & & \checkmark & \checkmark& \checkmark \\
LeftShoulderYaw & & \checkmark & \checkmark& \checkmark \\
LeftElbow & & \checkmark & \checkmark& \checkmark \\
LeftWristRoll & & \checkmark & \checkmark& \checkmark \\
LeftWristPitch & & \checkmark & \checkmark& \checkmark \\
LeftWristYaw & & \checkmark & \checkmark& \checkmark \\
RightShoulderPitch & \checkmark & \checkmark & \checkmark& \checkmark \\
RightShoulderRoll & \checkmark & \checkmark & \checkmark& \checkmark \\
RightShoulderYaw & \checkmark & \checkmark & \checkmark& \checkmark \\
RightElbow & \checkmark & \checkmark & \checkmark& \checkmark \\
RightWristRoll & \checkmark & \checkmark & \checkmark& \checkmark \\
RightWristPitch & \checkmark & \checkmark & \checkmark& \checkmark \\
RightWristYaw & \checkmark & \checkmark & \checkmark & \checkmark \\
LinearX & & & \checkmark & \checkmark \\
LinearY & &  & \checkmark & \checkmark \\
RotYaw & & & \checkmark & \checkmark \\
 \bottomrule
\end{tabular}
}

\end{center}
\end{table}

\begin{table}[h]
\begin{center}
\captionsetup{width=\textwidth}
\caption{Robot configurations}
\label{tb: tasks}
\scalebox{1}{
\centering
    
    \begin{tabular}{lcccc}
        \toprule
        Robot Config Class& Dynamic Order & Degree of Freedoms & State Dimension & Control Dimension \\ 
        \midrule
        G1RightArmDynamic1Config  & 1 & 7  & 7  & 7  \\
        G1RightArmDynamic2Config & 2 & 7  & 14 & 7  \\
        G1FixedBaseDynamic1Config & 1 & 17 & 17 & 17 \\
        G1FixedBaseDynamic2Config & 2 & 17 & 34 & 17 \\
        G1MobileBaseDynamic1Config & 1 & 20 & 20 & 20 \\
        G1MobileBaseDynamic2Config & 2 & 20 & 40 & 20 \\
        G1SportModeDynamic1Config & 1 & 20 & 20 & 20 \\
        G1SportModeDynamic2Config & 2 & 20 & 40 & 20 \\
        \bottomrule
    \end{tabular}
}
\end{center}
\end{table}

%% file: table_template/tables/frame_config.tex
\begin{table}[ht]
\begin{center}
\captionsetup{width=15cm}
\caption{Collision Volume Properties by Frame}
\label{tb: collision_volumes}
\scalebox{1}{
\begin{tabular}{c|cccc}
  \toprule
  Frame Name & Type & Radius & EnvCollision & SelfCollision \\
  \midrule
WaistYaw & Sphere & 0.05 &  &  \\
WaistRoll & Sphere & 0.05 &  &  \\
WaistPitch & Sphere & 0.05 &  &  \\
LeftShoulderPitch & Sphere & 0.05 & \checkmark &  \\
LeftShoulderRoll & Sphere & 0.06 & \checkmark & \checkmark \\
LeftShoulderYaw & Sphere & 0.05 & \checkmark &  \\
LeftElbow & Sphere & 0.05 & \checkmark & \checkmark \\
LeftWristRoll & Sphere & 0.05 & \checkmark &  \\
LeftWristPitch & Sphere & 0.05 & \checkmark &  \\
LeftWristYaw & Sphere & 0.05 & \checkmark &  \\
RightShoulderPitch & Sphere & 0.05 & \checkmark &  \\
RightShoulderRoll & Sphere & 0.06 & \checkmark & \checkmark \\
RightShoulderYaw & Sphere & 0.05 & \checkmark &  \\
RightElbow & Sphere & 0.05 & \checkmark & \checkmark \\
RightWristRoll & Sphere & 0.05 & \checkmark &  \\
RightWristPitch & Sphere & 0.05 & \checkmark &  \\
RightWristYaw & Sphere & 0.05 & \checkmark &  \\
L\_ee & Sphere & 0.05 & \checkmark & \checkmark \\
R\_ee & Sphere & 0.05 & \checkmark & \checkmark \\
TorsoLink1 & Sphere & 0.10 & \checkmark & \checkmark \\
TorsoLink2 & Sphere & 0.10 & \checkmark & \checkmark \\
TorsoLink3 & Sphere & 0.08 & \checkmark & \checkmark \\
PelvisLink1 & Sphere & 0.05 &  &  \\
PelvisLink2 & Sphere & 0.05 &  &  \\
PelvisLink3 & Sphere & 0.05 &  &  \\
 \bottomrule
\end{tabular}
}
\end{center}
\end{table}

%% file: sections/safety_index_intro.tex


Below is the analysis of how to derive the time variation of the safety index.

\subsubsection{Notation}
\begin{itemize}
\item $n$ Number of Degree of Freedoms (DoFs)
\item $m$ Number of robot frames.
\item $\mathbf{x}_c$ Cartesian state of the robot frames
\item $\mathbf{p} \in \mathbb{R} ^{3m}$ Cartesian position of the robot frames
\item $\mathbf{v} \in \mathbb{R} ^{3m}$ Cartesian velocity of the robot frames
\item $\mathbf{a} \in \mathbb{R} ^{3m}$ Cartesian acceleration of the robot frames
\item $\mathbf{u}_c$ Cartesian control
\item $\mathbf{\dot x}_c = \mathbf{f}_c(\mathbf{x}_c) + \mathbf{g}_c(\mathbf{x}_c)\mathbf{u}_c$ Cartesian dynamics
\item $\mathbf{x}_q$ Configuration state of the robot DoFs
\item $\mathbf{\theta} \in \mathbb{R} ^{n}$ Configuration position of the robot DoFs
\item $\mathbf{\dot\theta} \in \mathbb{R} ^{n}$ Configuration velocity of the robot DoFs
\item $\mathbf{\ddot\theta} \in \mathbb{R} ^{n}$ Configuration acceleration of the robot DoFs
\item $\mathbf{u}_q$ Configuration control
\item $\mathbf{\dot x}_q = \mathbf{f}_q(\mathbf{x}_q) + \mathbf{g}_q(\mathbf{x}_q)\mathbf{u}_q$ Configuration dynamics
\item $\mathbf{FK}(\mathbf{\theta}) = \mathbf{p} , \mathbb{R} ^{n} \rightarrow \mathbb{R} ^{3m}$ Forward kinematics
\item $\mathbf{J}(\mathbf{\theta}) = \frac{\partial\mathbf{FK}}{\partial \mathbf{\theta}}, \mathbb{R} ^{3m\times n}$ Jacobian, $\mathbf{v} = \mathbf{J}(\mathbf{\theta}) \mathbf{\dot\theta}$
\item $\phi(\mathbf{x}) \in \mathbf{R}$ Safety Index 
\item $\dot \phi(\mathbf{x}) = \frac{d\phi(\mathbf{x})}{dt} = \frac{\partial\phi(\mathbf{x})}{\partial \mathbf{x}} \mathbf{\dot x} $ Time variation of Safety Index 
\end{itemize}

\subsubsection{Single Integrator}

\[
\mathbf{x}_c = \mathbf{p}, \mathbf{u_c} = \mathbf{v}
\]

Cartesian dynamics

\[
\mathbf{\dot x}_c = \mathbf{\dot p} = \mathbf{f}_c(\mathbf{x}_c) + \mathbf{g}_c(\mathbf{x}_c)\mathbf{u}_c = \mathbf{0} + \mathbf{I}\mathbf{u_c}
\]

Configuration dynamics

\[
\mathbf{x}_q = \mathbf{\theta}, \mathbf{u_q} = \mathbf{\dot \theta}
\]
\[
\mathbf{\dot x}_q = \mathbf{f}_q(\mathbf{x}_q) + \mathbf{g}_q(\mathbf{x}_q)\mathbf{u}_q = \mathbf{0} + \mathbf{g}_q(\mathbf{x}_q)\mathbf{u_q}
\]

\begin{remark}
For a 2D mobile robot, its DoFs $\mathbf{x}_q = [p_x, p_y, \psi]$ is consists of the global 2D position and the yaw angle. The control $\mathbf{u_q} = [\hat v_x, \hat v_y, \dot \psi]$ is the local frame linear and rotation velocity. Then we have:
\begin{equation}
\mathbf{g}_q(\mathbf{x}_q) = \begin{bmatrix}
\cos\psi & -\sin\psi & 0 \\
\sin\psi & \cos\psi & 0 \\
0 & 0 & 1
\end{bmatrix} \nonumber
\end{equation}
\end{remark}

For the single integrator we have:

\[
\mathbf{\dot x}_c = \mathbf{J}(\mathbf{\theta})\mathbf{\dot x}_q
\]
\[
\mathbf{u}_c = \mathbf{J}(\mathbf{\theta})\mathbf{g}_q(\mathbf{x}_q)\mathbf{u}_q
\]

Then we can begin derivative $\dot \phi$.

Start from the Cartesian state we have:
\begin{align}
\dot \phi &= \frac{\partial\phi(\mathbf{x}_c)}{\partial \mathbf{x}_c} \mathbf{\dot x_c} = \frac{\partial\phi(\mathbf{x}_c)}{\partial \mathbf{x}_c}\mathbf{J}(\mathbf{\theta})\mathbf{\dot x}_q  \\ \nonumber
&= \frac{\partial\phi(\mathbf{x}_c)}{\partial \mathbf{x}_c}\mathbf{f}_c(\mathbf{x}_c) + \frac{\partial\phi(\mathbf{x}_c)}{\partial \mathbf{x}_c}\mathbf{g}_c(\mathbf{x}_c)\mathbf{u}_c \\ \nonumber
&=  \frac{\partial\phi(\mathbf{x}_c)}{\partial \mathbf{x}_c}\mathbf{f}_c(\mathbf{x}_c) + \frac{\partial\phi(\mathbf{x}_c)}{\partial \mathbf{x}_c}\mathbf{g}_c(\mathbf{x}_c)\mathbf{J}(\mathbf{\theta})\mathbf{g}_q(\mathbf{x}_q)\mathbf{u_q}\\ \nonumber
&= \frac{\partial\phi(\mathbf{x}_c)}{\partial \mathbf{x}_c}\mathbf{g}_c(\mathbf{x}_c)\mathbf{J}(\mathbf{\theta})\mathbf{g}_q(\mathbf{x}_q)\mathbf{u_q}\\ \nonumber
\end{align}

Start from the Configuration state we have:
\begin{align}
\dot \phi &= \frac{\partial\phi(\mathbf{x}_q)}{\partial \mathbf{x}_q} \mathbf{\dot x_q} \\ \nonumber
&=  \frac{\partial\phi(\mathbf{x}_q)}{\partial \mathbf{x}_q}\mathbf{f}_q(\mathbf{x}_q) + \frac{\partial\phi(\mathbf{x}_q)}{\partial \mathbf{x}_q}\mathbf{g}_q(\mathbf{x}_q)\mathbf{u_q}\\ \nonumber
&= \frac{\partial\phi(\mathbf{x}_q)}{\partial \mathbf{x}_q}\mathbf{g}_q(\mathbf{x}_q)\mathbf{u_q}\\ \nonumber
\end{align}

In the practice of \spark, $\frac{\partial\phi(\mathbf{x}_c)}{\partial \mathbf{x}_c}$ can be derived analytically if the safety index is defined in Cartesian space. $ \frac{\partial\phi(\mathbf{x}_q)}{\partial \mathbf{x}_q}$ can be calculated with numerical methods such as central differential method. From the above derivation we also have:

\[
 \frac{\partial\phi(\mathbf{x}_q)}{\partial \mathbf{x}_q} =  \frac{\partial\phi(\mathbf{x}_c)}{\partial \mathbf{x}_c}\mathbf{g}_c(\mathbf{x}_c)\mathbf{J}(\mathbf{\theta})
\]
which can be used to do the safety index gradient conversion between the Cartesian space and the Configuration space.
\subsubsection{Double Integrator}

\[
\mathbf{x}_c = \begin{bmatrix}
\mathbf{p} \\
\mathbf{v} \\
\end{bmatrix}, \mathbf{u_c} = \mathbf{a}
\]

Cartesian dynamics

\[
\mathbf{\dot x}_c = \begin{bmatrix}
\mathbf{\dot p} \\
\mathbf{\dot v} \\
\end{bmatrix} = \mathbf{f}_c(\mathbf{x}_c) + \mathbf{g}_c(\mathbf{x}_c)\mathbf{u}_c = 
\begin{bmatrix}
\mathbf{0} & \mathbf{I} \\
\mathbf{0} & \mathbf{0} \\
\end{bmatrix} 
\begin{bmatrix}
\mathbf{p} \\
\mathbf{v} \\
\end{bmatrix} 
+ 
\begin{bmatrix}
\mathbf{0} \\
\mathbf{I} \\
\end{bmatrix}\mathbf{u_c}
\]

Configuration dynamics

\[
\mathbf{x}_q = \begin{bmatrix}
\mathbf{\theta} \\
\mathbf{\dot \theta} \\
\end{bmatrix}, \mathbf{u_q} = \mathbf{\ddot \theta}
\]
\[
\mathbf{\dot x}_q = \begin{bmatrix}
\mathbf{\dot \theta} \\
\mathbf{\ddot \theta} \\
\end{bmatrix} = \mathbf{f}_q(\mathbf{x}_q) + \mathbf{g}_q(\mathbf{x}_q)\mathbf{u}_q 
\]

\begin{remark}
For a 2D mobile robot, its DoFs $\mathbf{x}_q = [p_x, p_y, \psi, \dot p_x, \dot p_y, \dot \psi]$ is consists of the global linear and rotation position and velocity. The control $\mathbf{u_q} = [\hat a_x, \hat a_y, \ddot \psi]$ is the local frame linear and rotation accelerations. Then we have:
\begin{align}
f(\mathbf{x}) &= \begin{bmatrix}
\dot p_x \\
\dot p_y \\
\dot \psi \\
-\hat v_x \dot \psi \sin\psi - \hat v_y \dot \psi \cos\psi \\
\hat v_x \dot \psi \cos\psi - \hat v_y \dot \psi \sin\psi \\
0
\end{bmatrix} \\ \nonumber
&= \begin{bmatrix}
\dot p_x \\
\dot p_y \\
\dot \psi \\
-(\dot p_x \cos\psi + \dot p_y \sin\psi) \dot \psi \sin\psi - (- \dot p_x \sin\psi + \dot p_y \cos\psi) \dot \psi \cos\psi \\
(\dot p_x \cos\psi + \dot p_y \sin\psi) \dot \psi \cos\psi - (- \dot p_x \sin\psi + \dot p_y \cos\psi) \dot \psi \sin\psi \\
0
\end{bmatrix}.
\end{align}

\begin{equation}
g(\mathbf{x}) = \begin{bmatrix}
0 & 0 & 0 \\
0 & 0 & 0 \\
0 & 0 & 0 \\
\cos\psi & -\sin\psi & 0 \\
\sin\psi & \cos\psi & 0 \\
0 & 0 & 1
\end{bmatrix}.
\end{equation}
\end{remark}

For the double integrator we have:

\[
\mathbf{a} = \frac{d\mathbf{v}}{dt} = \frac{d\mathbf{J}(\mathbf{\theta}) \mathbf{\dot\theta}}{dt} = \mathbf{\dot J}(\mathbf{\theta}, \mathbf{\dot\theta}) \mathbf{\dot\theta} + \mathbf{J}(\mathbf{\theta}) \mathbf{\ddot\theta}
\]

\begin{align}
\mathbf{\dot x}_c &= \begin{bmatrix}
\mathbf{v} \\
\mathbf{a} \\
\end{bmatrix} = \begin{bmatrix}
\mathbf{v} \\
\mathbf{J}(\mathbf{\theta}) \mathbf{\ddot\theta} + \mathbf{\dot J}(\mathbf{\theta}, \mathbf{\dot\theta}) \mathbf{\dot\theta} \\
\end{bmatrix} \\ \nonumber
\mathbf{u}_c &=  \mathbf{\dot J}(\mathbf{\theta}, \mathbf{\dot\theta}) \mathbf{\dot\theta} + \mathbf{J}(\mathbf{\theta}) \mathbf{\ddot\theta} \\ \nonumber
&= \mathbf{\dot J}(\mathbf{\theta}, \mathbf{\dot\theta})\mathbf{\dot\theta} + \mathbf{J}(\mathbf{\theta})\begin{bmatrix}
\mathbf{0} & \mathbf{I}
\end{bmatrix}\mathbf{f}_q(\mathbf{x}_q) + 
\mathbf{J}(\mathbf{\theta})\begin{bmatrix}
\mathbf{0} & \mathbf{I}
\end{bmatrix}\mathbf{g}_q(\mathbf{x}_q)\mathbf{u}_q
\end{align}

Then we can begin derivative $\dot \phi$.

Start from the Cartesian state we have:
\begin{align}
\dot \phi &= \frac{\partial\phi(\mathbf{x}_c)}{\partial \mathbf{x}_c} \mathbf{\dot x_c} \\ \nonumber
&= \frac{\partial\phi(\mathbf{x}_c)}{\partial \mathbf{x}_c}\mathbf{f}_c(\mathbf{x}_c) + \frac{\partial\phi(\mathbf{x}_c)}{\partial \mathbf{x}_c}\mathbf{g}_c(\mathbf{x}_c)\mathbf{u}_c\\ \nonumber
&= \frac{\partial\phi(\mathbf{x}_c)}{\partial \mathbf{x}_c}\mathbf{f}_c(\mathbf{x}_c) \\ \nonumber
&+ \frac{\partial\phi(\mathbf{x}_c)}{\partial \mathbf{x}_c}\mathbf{g}_c(\mathbf{x}_c)\mathbf{\dot J}(\mathbf{\theta}, \mathbf{\dot\theta})\mathbf{\dot\theta} \\ \nonumber
&+ 
\frac{\partial\phi(\mathbf{x}_c)}{\partial \mathbf{x}_c}\mathbf{g}_c(\mathbf{x}_c)\mathbf{J}(\mathbf{\theta})\begin{bmatrix}
\mathbf{0} & \mathbf{I}
\end{bmatrix}\mathbf{f}_q(\mathbf{x}_q) \\ \nonumber
&+ 
\frac{\partial\phi(\mathbf{x}_c)}{\partial \mathbf{x}_c}\mathbf{g}_c(\mathbf{x}_c)
\mathbf{J}(\mathbf{\theta})\begin{bmatrix}
\mathbf{0} & \mathbf{I}
\end{bmatrix}\mathbf{g}_q(\mathbf{x}_q)\mathbf{u}_q
\end{align}

where:
\[
\frac{\partial\phi(\mathbf{x}_c)}{\partial \mathbf{x}_c}\mathbf{f}_c(\mathbf{x}_c) =\frac{\partial\phi(\mathbf{x}_c)}{\partial \mathbf{p}} \mathbf{v}
\]

\[
\frac{\partial\phi(\mathbf{x}_c)}{\partial \mathbf{x}_c}\mathbf{g}_c(\mathbf{x}_c) =\frac{\partial\phi(\mathbf{x}_c)}{\partial \mathbf{v}}
\]

Start from the Configuration state we have:
\begin{align}
\dot \phi &= \frac{\partial\phi(\mathbf{x}_q)}{\partial \mathbf{x}_q} \mathbf{\dot x_q} \\ \nonumber
&=  \frac{\partial\phi(\mathbf{x}_q)}{\partial \mathbf{x}_q}\mathbf{f}_q(\mathbf{x}_q) + \frac{\partial\phi(\mathbf{x}_q)}{\partial \mathbf{x}_q}\mathbf{g}_q(\mathbf{x}_q)\mathbf{u_q}\\ \nonumber
\end{align}

In the practice of \spark, $\frac{\partial\phi(\mathbf{x}_c)}{\partial \mathbf{x}_c}$ can be derived analytically if the safety index is defined in Cartesian space. $ \frac{\partial\phi(\mathbf{x}_q)}{\partial \mathbf{x}_q}$ can be calculated with numerical methods such as central differential method. From the above derivation we also have:

\[
 \frac{\partial\phi(\mathbf{x}_q)}{\partial \mathbf{x}_q} =  \frac{\partial\phi(\mathbf{x}_c)}{\partial \mathbf{x}_c}\mathbf{g}_c(\mathbf{x}_c)
\mathbf{J}(\mathbf{\theta})\begin{bmatrix}
\mathbf{0} & \mathbf{I}
\end{bmatrix}
\]
which can be used to do the safety index gradient conversion between the Cartesian space and the Configuration space.


%% file: table_template/tables/task_config.tex
\begin{table}[htbp]
\centering
\captionsetup{width=0.95\textwidth}
\caption{Configuration Parameters for Different Test Cases}
\label{tab:config_parameters}

\resizebox{\textwidth}{!}{
\begin{tabular}{p{5cm} | C{3.5cm} C{3cm} C{2.5cm} C{2.5cm} C{2.5cm}}
\toprule
\textbf{Task Name} & \textbf{Robot Class} & \textbf{Num Obstacles} & \textbf{Obstacle Velocity} & \textbf{Arm Goal Velocity} & \textbf{Base Goal Velocity} \\
\midrule
G1FixedBase\_D1\_AG\_SO\_v0 & G1FixedBaseDynamic1Config & 10 & 0.0   & 0.0 & N/A \\
G1FixedBase\_D1\_AG\_SO\_v1 & G1FixedBaseDynamic1Config & 50 & 0.0   & 0.0 & N/A \\
G1FixedBase\_D1\_AG\_DO\_v0 & G1FixedBaseDynamic1Config & 10 & 0.005 & 0.0 & N/A \\
G1FixedBase\_D1\_AG\_DO\_v1 & G1FixedBaseDynamic1Config & 50 & 0.005 & 0.0 & N/A \\
G1MobileBase\_D1\_WG\_SO\_v0 & G1MobileBaseDynamic1Config & 10 & 0.0   & 0.0 & 0.0 \\
G1MobileBase\_D1\_WG\_SO\_v1 & G1MobileBaseDynamic1Config & 50 & 0.0   & 0.0 & 0.0 \\
G1MobileBase\_D1\_WG\_DO\_v0 & G1MobileBaseDynamic1Config & 10 & 0.005 & 0.0 & 0.0 \\
G1MobileBase\_D1\_WG\_DO\_v1 & G1MobileBaseDynamic1Config & 50 & 0.005 & 0.0 & 0.0 \\
\bottomrule
\end{tabular}
}
\end{table}

%% file: table_template/tables/self_collision_pair.tex
\begin{table}[h]
    \centering
    \caption{Self-collision joint pairs}
    \scalebox{1}{\begin{tabular}{cc}
        \hline
        \textbf{Joint 1} & \textbf{Joint 2} \\
        \hline
        left\_shoulder\_roll\_joint & left\_elbow\_joint \\
        left\_shoulder\_roll\_joint & right\_shoulder\_roll\_joint \\
        left\_shoulder\_roll\_joint & right\_elbow\_joint \\
        left\_shoulder\_roll\_joint & L\_ee \\
        left\_shoulder\_roll\_joint & R\_ee \\
        left\_shoulder\_roll\_joint & torso\_link\_3 \\
        left\_elbow\_joint & right\_shoulder\_roll\_joint \\
        left\_elbow\_joint & right\_elbow\_joint \\
        left\_elbow\_joint & L\_ee \\
        left\_elbow\_joint & R\_ee \\
        left\_elbow\_joint & torso\_link\_1 \\
        left\_elbow\_joint & torso\_link\_2 \\
        left\_elbow\_joint & torso\_link\_3 \\
        right\_shoulder\_roll\_joint & right\_elbow\_joint \\
        right\_shoulder\_roll\_joint & L\_ee \\
        right\_shoulder\_roll\_joint & R\_ee \\
        right\_shoulder\_roll\_joint & torso\_link\_3 \\
        right\_elbow\_joint & L\_ee \\
        right\_elbow\_joint & R\_ee \\
        right\_elbow\_joint & torso\_link\_1 \\
        right\_elbow\_joint & torso\_link\_2 \\
        right\_elbow\_joint & torso\_link\_3 \\
        L\_ee & R\_ee \\
        L\_ee & torso\_link\_1 \\
        L\_ee & torso\_link\_2 \\
        L\_ee & torso\_link\_3 \\
        R\_ee & torso\_link\_1 \\
        R\_ee & torso\_link\_2 \\
        R\_ee & torso\_link\_3 \\
        \hline
    \end{tabular}}
    \label{tab:self_collision_pairs}
\end{table}

%% file: sections/param_tuning.tex
\begin{figure*}[htbp]
    \centering
    \vspace{2cm}  
    \begin{tikzpicture}[transform canvas={xshift=-4.5cm}] 
        \def\imgwidth{4.5cm}  
        \def\xgap{4.5}  
        \def\ygap{-5} 
        \def\xshift{-2} 

        \node at (\xshift + 0, 0) {\includegraphics[width=\imgwidth]{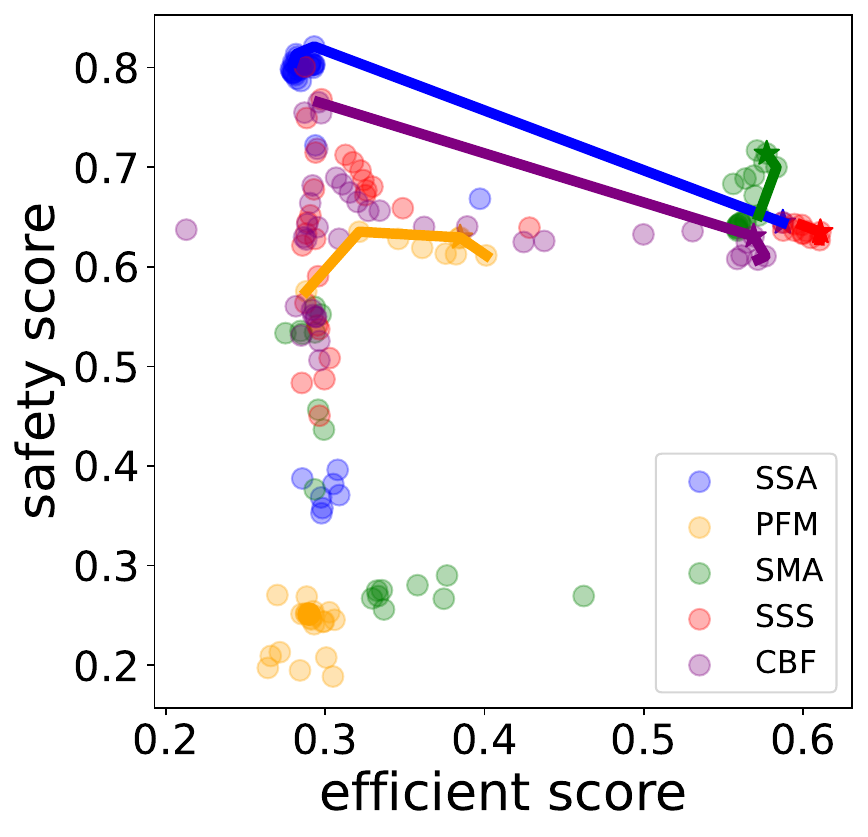}};
        \node[below] at (\xshift + 0, -2.2) {\small (a) G1FixedBase\_D1\_AG\_DO\_v0};

        \node at (\xshift + \xgap, 0) {\includegraphics[width=\imgwidth]{figure/param_tuning/G1FixedBase_SG_DO_v0_convex.pdf}};
        \node[below] at (\xshift + \xgap, -2.2) {\small (b) G1FixedBase\_D1\_AG\_DO\_v1};

        \node at (\xshift + 2*\xgap, 0) {\includegraphics[width=\imgwidth]{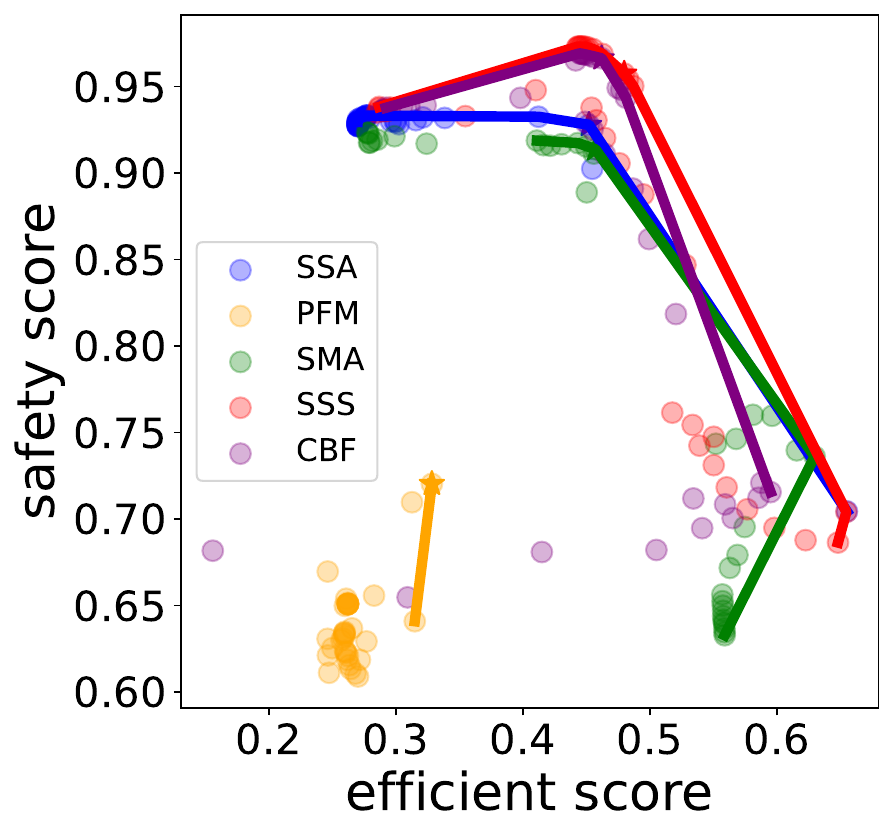}};
        \node[below] at (\xshift + 2*\xgap, -2.2) {\small (c) G1FixedBase\_D1\_AG\_SO\_v0};

        \node at (\xshift + 3*\xgap, 0) {\includegraphics[width=\imgwidth]{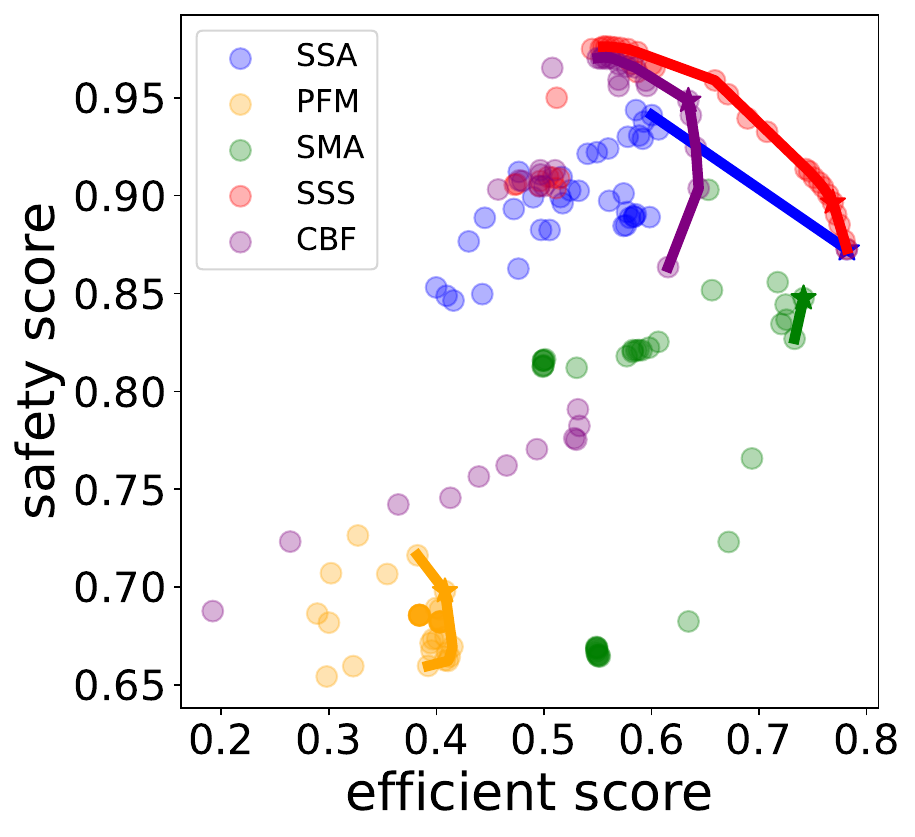}};
        \node[below] at (\xshift + 3*\xgap, -2.2) {\small (d) G1FixedBase\_D1\_AG\_SO\_v1};

        \node at (\xshift + 0, \ygap) {\includegraphics[width=\imgwidth]{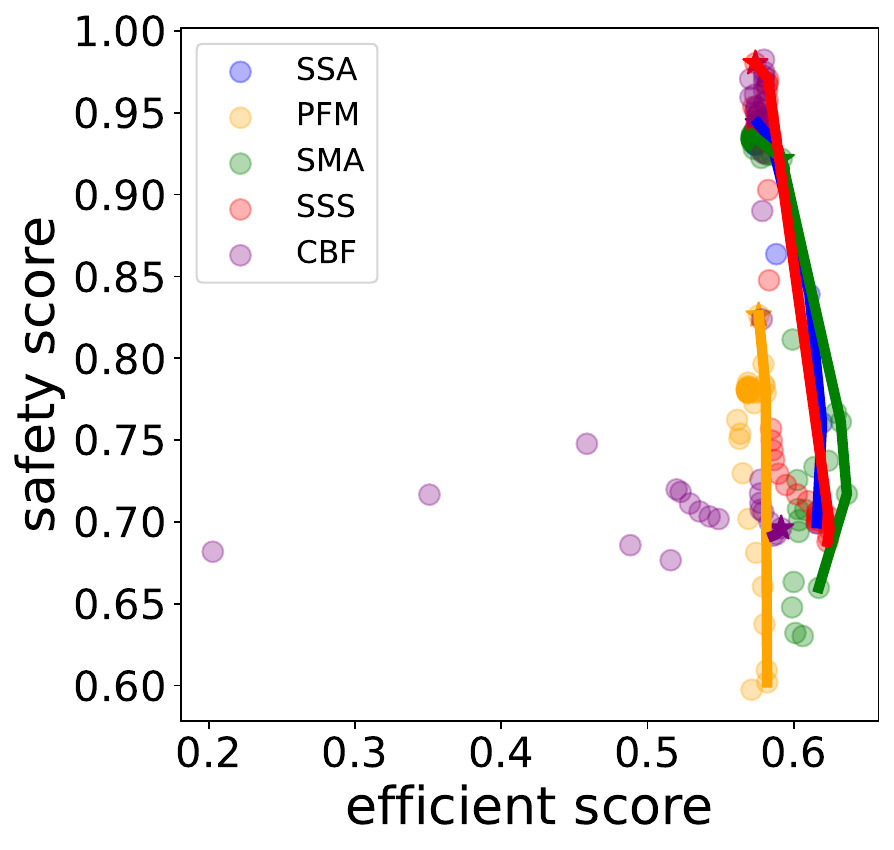}};
        \node[below, align=center, font=\small] at (\xshift + 0, \ygap-2.2) {(e) G1MobileBase\_D1\\\_WG\_DO\_v0};

        \node at (\xshift + \xgap, \ygap) {\includegraphics[width=\imgwidth]{figure/param_tuning/G1WholeBody_SG_DO_v0_convex.pdf}};
        \node[below, align=center, font=\small] at (\xshift + \xgap, \ygap-2.2) {(f) G1MobileBase\_D1\\\_WG\_DO\_v1};

        \node at (\xshift + 2*\xgap, \ygap) {\includegraphics[width=\imgwidth]{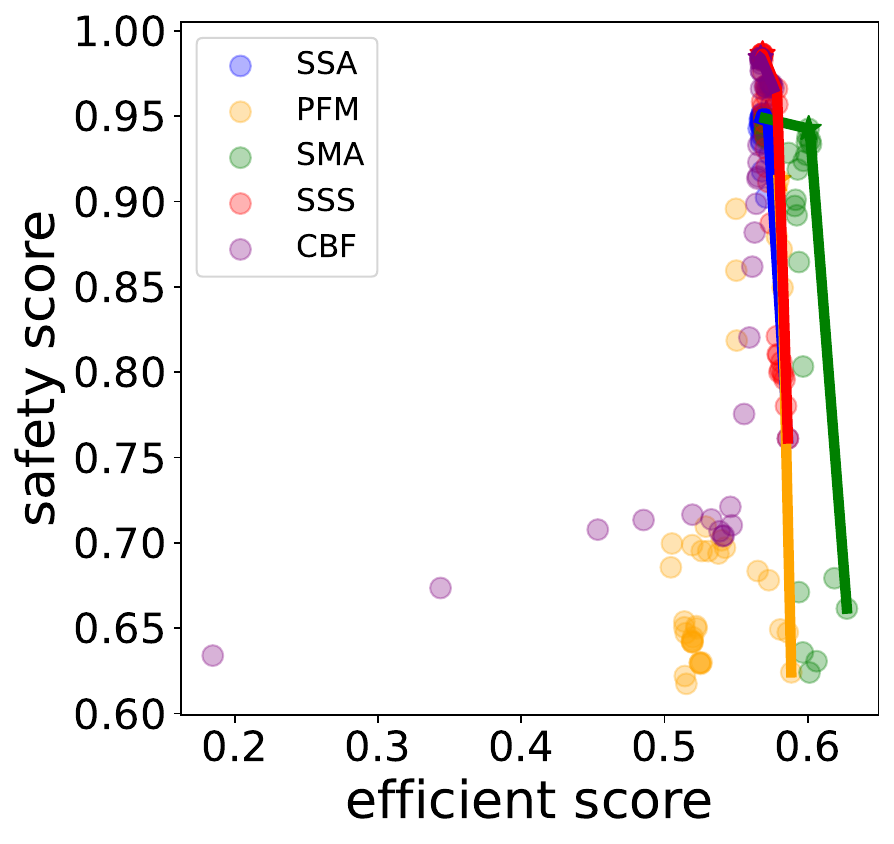}};
        \node[below, align=center, font=\small] at (\xshift + 2*\xgap, \ygap-2.2) {(g) G1MobileBase\_D1\\\_WG\_SO\_v0};

        \node at (\xshift + 3*\xgap, \ygap) {\includegraphics[width=\imgwidth]{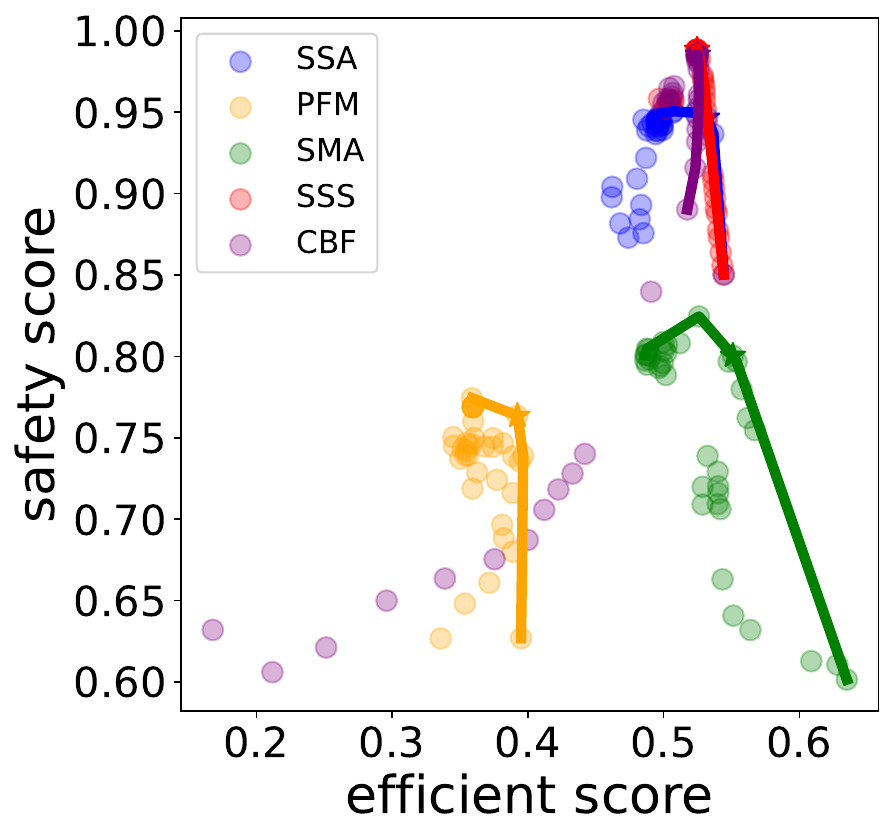}};
        \node[below, align=center, font=\small] at (\xshift + 3*\xgap, \ygap-2.2) {(h) G1MobileBase\_D1\\\_WG\_SO\_v1};

    \end{tikzpicture}
    \vspace{8cm} 
    
    \caption{Trade-off curves between safety and efficiency.}
    \label{fig: param_tuning}
\end{figure*}

%% file: table_template/tables/hyperparameter.tex
\begin{table}[h]
\centering
\captionsetup{width=0.95\textwidth}
\caption{Task and Algorithm Parameter Relationships}
\label{tb:task_params}

\begin{tabular}{c|ccccc}
\toprule
\textbf{Task} & \textbf{SSA ($\eta_{ssa}$)} & \textbf{SSS ($\lambda_{sss}$)} & \textbf{CBF ($\lambda_{cbf}$)} & \textbf{PFM ($c_{pfm}$)} & \textbf{SMA ($c_{sma}$)} \\
\midrule
G1FixedBase\_D1\_AG\_SO\_v0 & 0.2 & 10.0 & 20.0 & 0.1 & 40.0 \\
G1FixedBase\_D1\_AG\_SO\_v1 & 0.01 & 0.4 & 6.0 & 1.0 & 7.0 \\  
G1FixedBase\_D1\_AG\_DO\_v0 & 0.01 & 0.7 & 0.5 & 0.4 & 6.0 \\
G1FixedBase\_D1\_AG\_DO\_v1 & 0.01 & 0.01 & 1.0 & 0.01 & 6.0 \\  
G1MobileBase\_D1\_WG\_SO\_v0 & 0.8 & 60.0 & 80.0 & 0.9 & 6.0 \\
G1MobileBase\_D1\_WG\_SO\_v1 & 0.2 & 60.0 & 100.0 & 0.8 & 9.0 \\  
G1MobileBase\_D1\_WG\_DO\_v0 & 0.9 & 200.0 & 3.0 & 2.0 & 7.0 \\
G1MobileBase\_D1\_WG\_DO\_v1 & 0.4 & 100.0 & 100.0 & 0.7 & 0.1 \\  
\bottomrule
\end{tabular}
\end{table}


%% file: sections/full_evaluation.tex
\begin{figure*}[htbp]
    \centering

    \vspace{2cm}  
    \begin{tikzpicture}[transform canvas={xshift=-4.5cm}] 
        \def\imgwidth{4.5cm}  
        \def\xgap{4.5}  
        \def\ygap{-4} 
        \def\xshift{-2} 

        \node at (\xshift + 0, 0) {\includegraphics[width=\imgwidth]{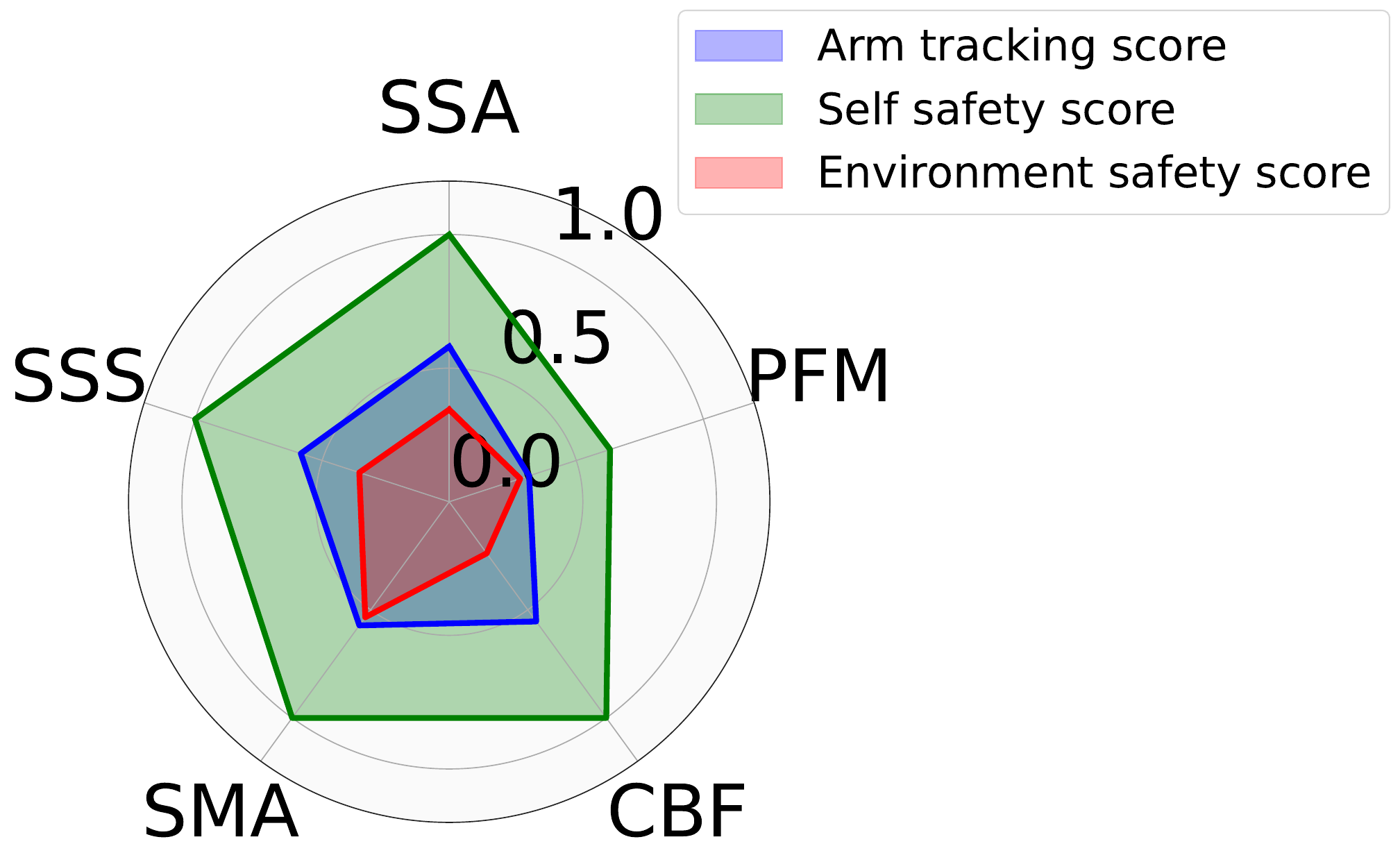}};
        \node[below] at (\xshift + 0, -1.7) {\small (a) G1FixedBase\_D1\_AG\_DO\_v0};

        \node at (\xshift + \xgap, 0) {\includegraphics[width=\imgwidth]{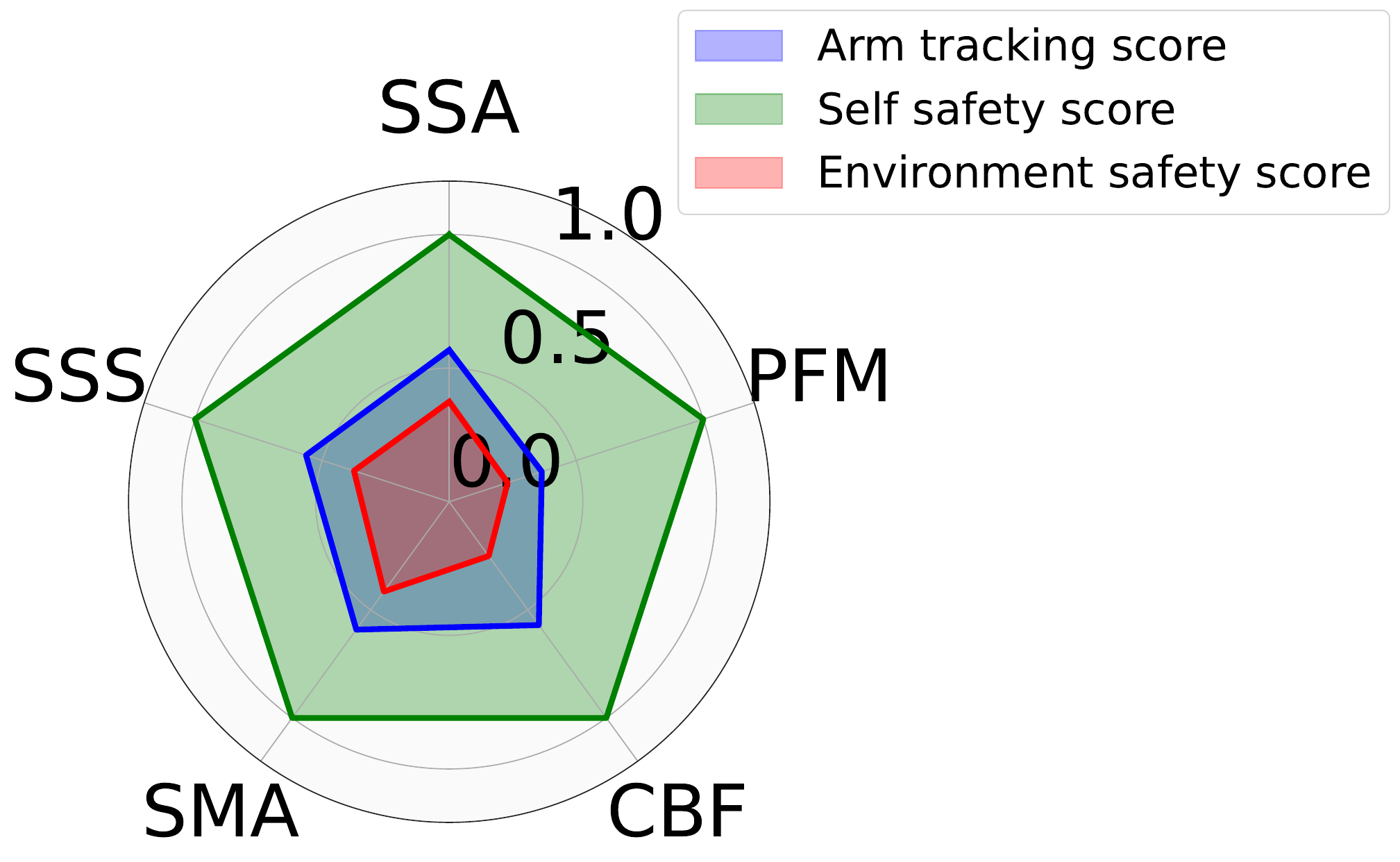}};
        \node[below] at (\xshift + \xgap, -1.7) {\small (b) G1FixedBase\_D1\_AG\_DO\_v1};

        \node at (\xshift + 2*\xgap, 0) {\includegraphics[width=\imgwidth]{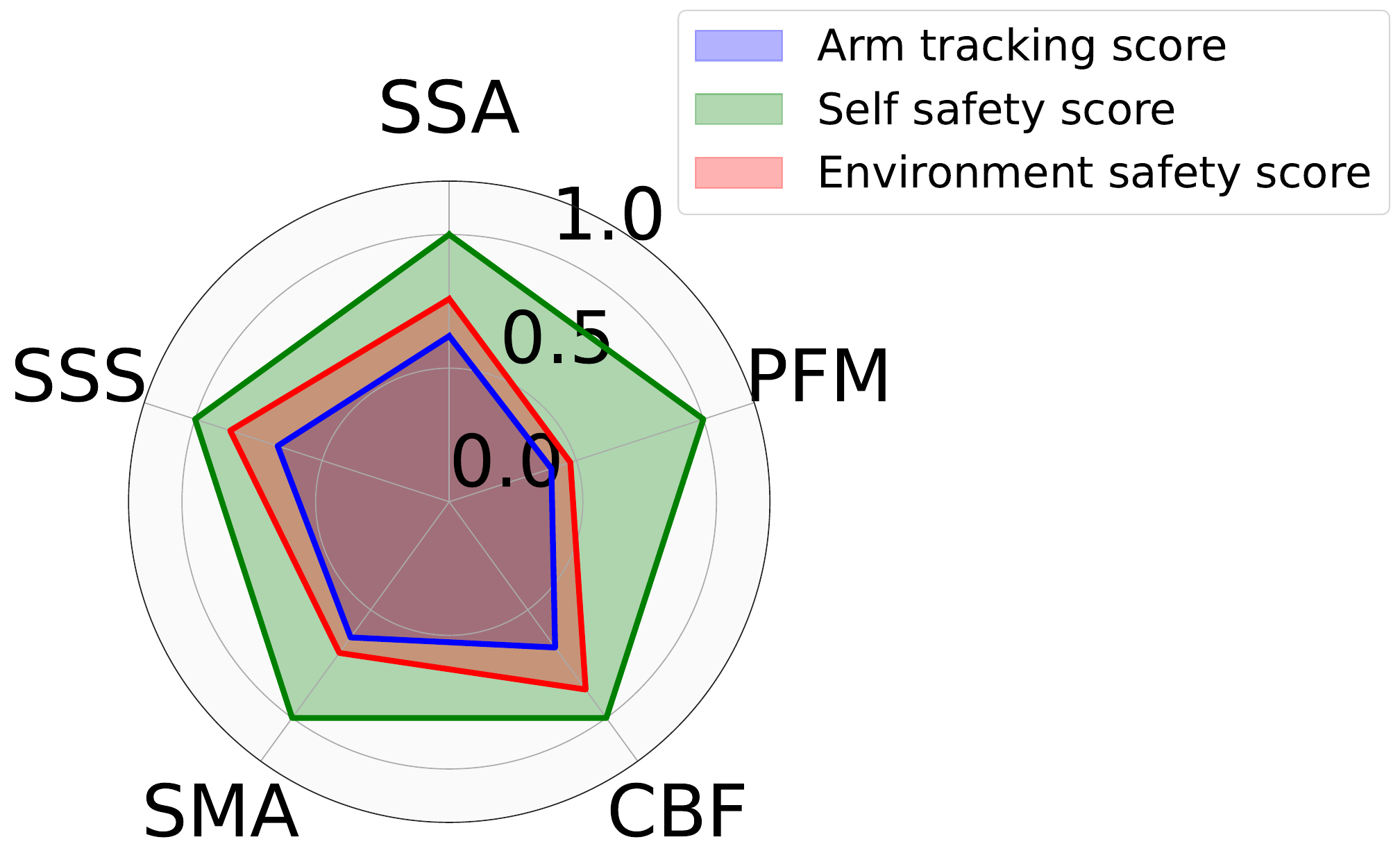}};
        \node[below] at (\xshift + 2*\xgap, -1.7) {\small (c) G1FixedBase\_D1\_AG\_SO\_v0};

        \node at (\xshift + 3*\xgap, 0) {\includegraphics[width=\imgwidth]{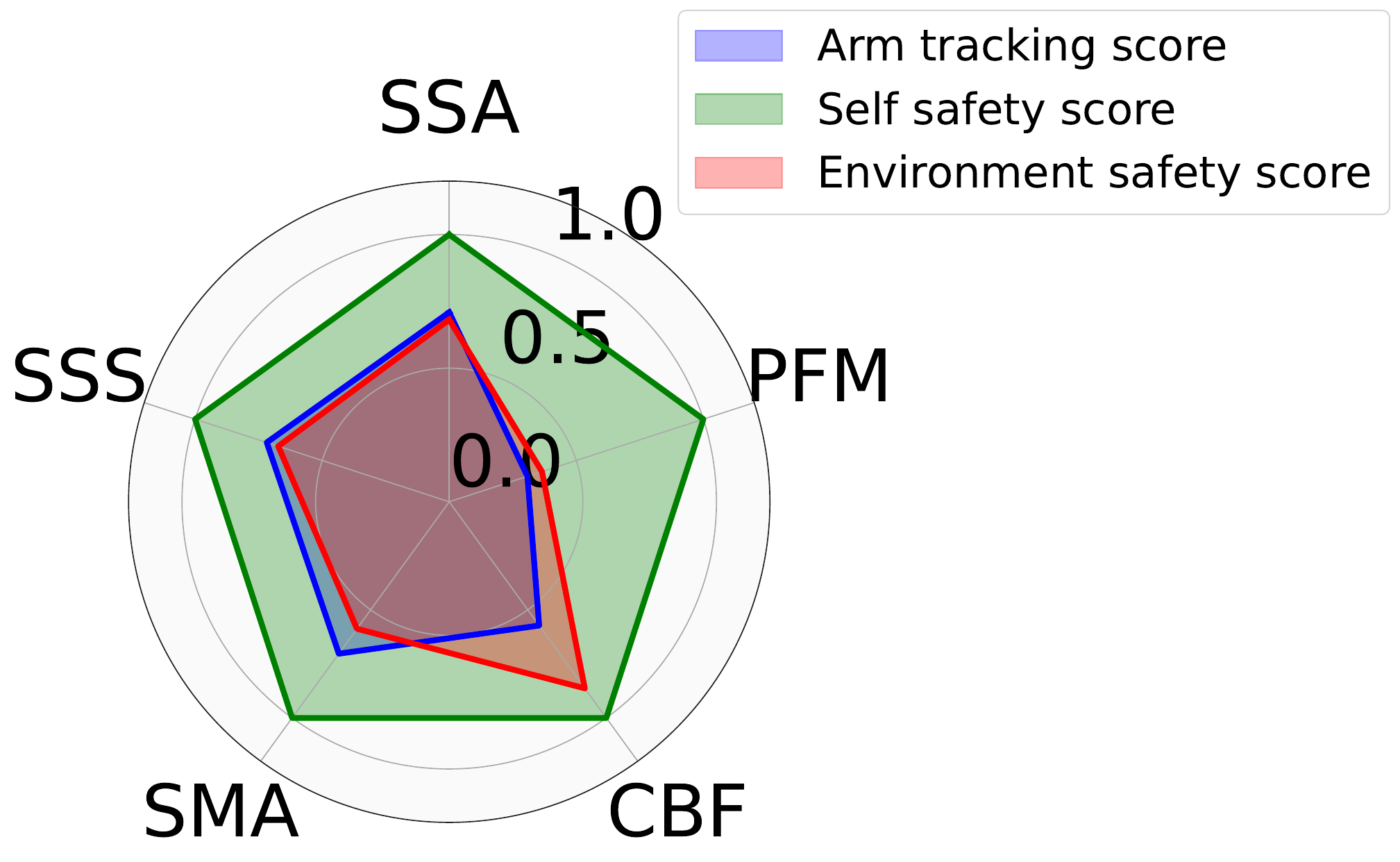}};
        \node[below] at (\xshift + 3*\xgap, -1.7) {\small (d) G1FixedBase\_D1\_AG\_SO\_v1};

        \node at (\xshift + 0, \ygap) {\includegraphics[width=\imgwidth]{figure/full_evaluation/G1WholeBody_SG_DO_v1_radar.pdf}};
        \node[below, align=center, font=\small] at (\xshift + 0, \ygap-1.7) {(e) G1MobileBase\_D1\\\_WG\_DO\_v0};

        \node at (\xshift + \xgap, \ygap) {\includegraphics[width=\imgwidth]{figure/full_evaluation/G1WholeBody_SG_DO_v0_radar.pdf}};
        \node[below, align=center, font=\small] at (\xshift + \xgap, \ygap-1.7) {(f) G1MobileBase\_D1\\\_WG\_DO\_v1};

        \node at (\xshift + 2*\xgap, \ygap) {\includegraphics[width=\imgwidth]{figure/full_evaluation/G1WholeBody_SG_SO_v1_radar.pdf}};
        \node[below, align=center, font=\small] at (\xshift + 2*\xgap, \ygap-1.7) {(g) G1MobileBase\_D1\\\_WG\_SO\_v0};

        \node at (\xshift + 3*\xgap, \ygap) {\includegraphics[width=\imgwidth]{figure/full_evaluation/G1WholeBody_SG_SO_v0_radar.pdf}};
        \node[below, align=center, font=\small] at (\xshift + 3*\xgap, \ygap-1.7) {(h) G1MobileBase\_D1\\\_WG\_SO\_v1};

    \end{tikzpicture}
    \vspace{6.5cm} 
    
    \label{fig: benchmark_performance}
    \caption{Performance comparison of the benchmark}
\end{figure*}

%% file: sections/success_matrix.tex
\begin{figure*}[htbp]
    \centering
    \vspace{0.7cm}  
    \begin{tikzpicture}[transform canvas={xshift=0cm}]
        \def\imgwidth{15cm}  
        \def\ygap{-2.5}  
        
        \node at (0, 0) {\includegraphics[width=\imgwidth]{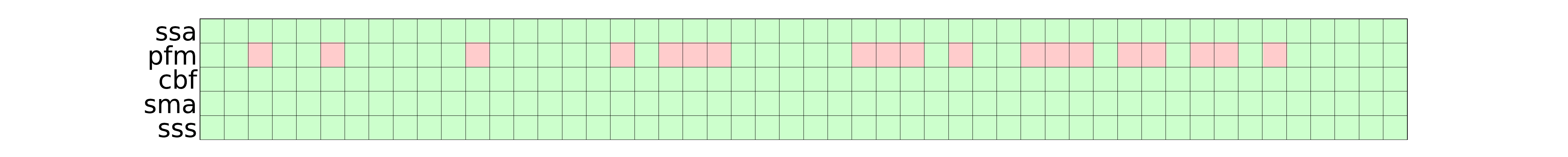}};
        \node[below] at (0, -1) {\small (a) G1FixedBase\_D1\_AG\_SO\_v0};
        
        \node at (0, \ygap) {\includegraphics[width=\imgwidth]{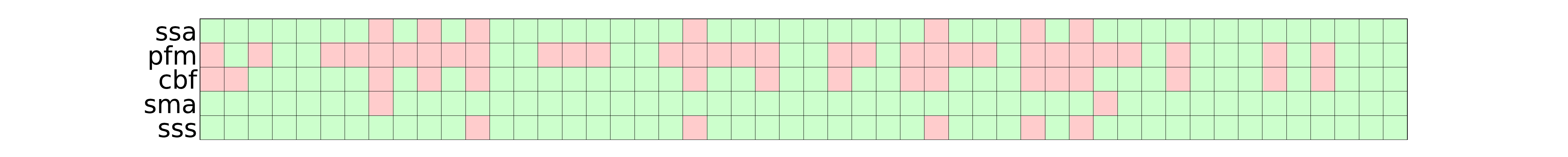}};
        \node[below] at (0, \ygap-1) {\small (b) G1FixedBase\_D1\_AG\_SO\_v1};
        
        \node at (0, 2*\ygap) {\includegraphics[width=\imgwidth]{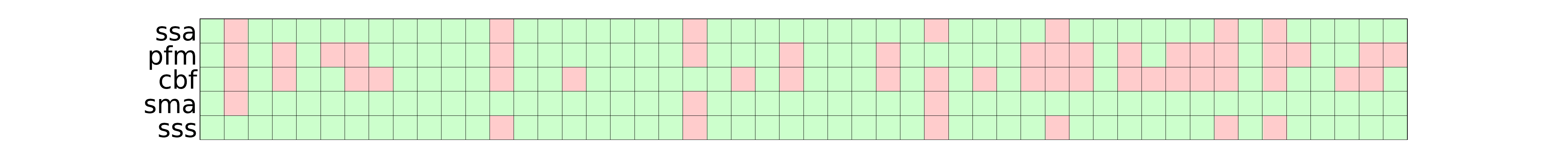}};
        \node[below] at (0, 2*\ygap-1) {\small (c) G1FixedBase\_D1\_AG\_DO\_v0};
        
        \node at (0, 3*\ygap) {\includegraphics[width=\imgwidth]{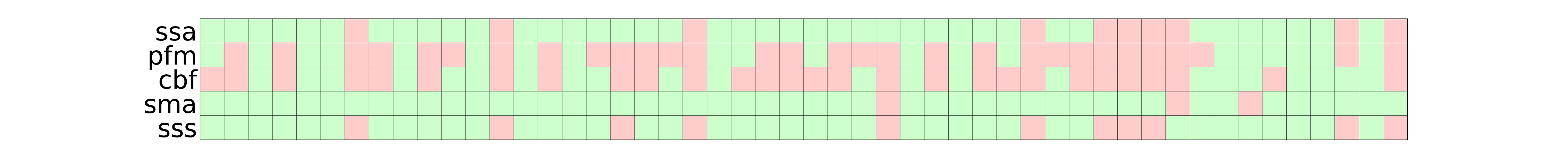}};
        \node[below] at (0, 3*\ygap-1) {\small (d) G1FixedBase\_D1\_AG\_DO\_v1};
        
        \node at (0, 4*\ygap) {\includegraphics[width=\imgwidth]{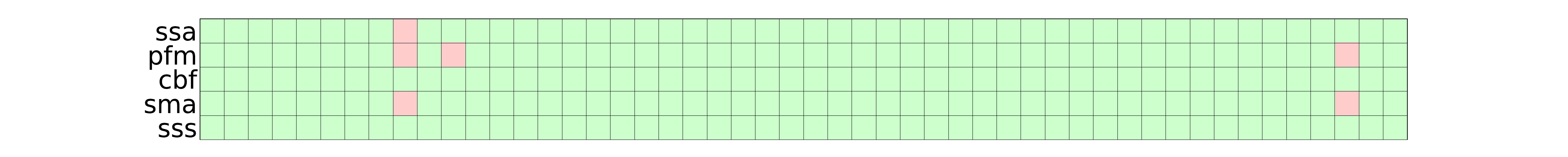}};
        \node[below] at (0, 4*\ygap-1) {\small (e) G1MobileBase\_D1\_WG\_SO\_v0};
        
        \node at (0, 5*\ygap) {\includegraphics[width=\imgwidth]{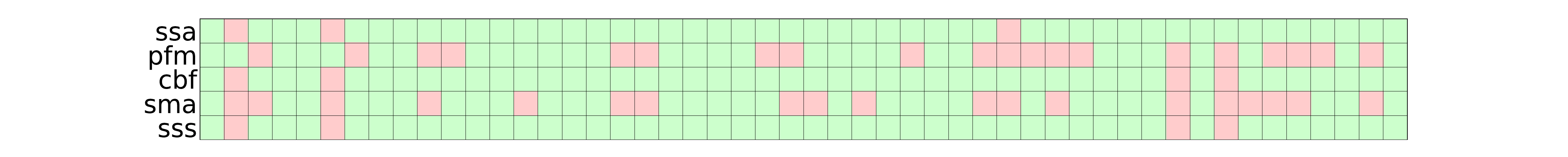}};
        \node[below] at (0, 5*\ygap-1) {\small (f) G1MobileBase\_D1\_WG\_SO\_v1};
        
        \node at (0, 6*\ygap) {\includegraphics[width=\imgwidth]{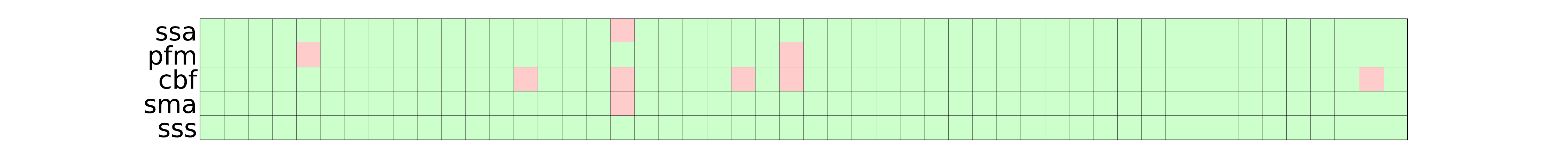}};
        \node[below] at (0, 6*\ygap-1) {\small (g) G1MobileBase\_D1\_WG\_DO\_v0};
        
        \node at (0, 7*\ygap) {\includegraphics[width=\imgwidth]{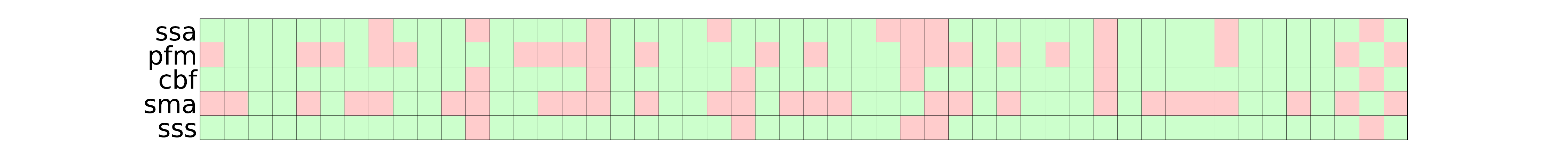}};
        \node[below] at (0, 7*\ygap-1) {\small (h) G1MobileBase\_D1\_WG\_DO\_v1};

    \end{tikzpicture}
    \vspace{19.3cm}
    \caption{Success matrices for different tasks}
\end{figure*}


%% file: table_template/tables/algorithm_compare.tex
\begin{table}[h]
\centering
\caption{Success Matrices for Different Scenarios}
\label{tb:algorithm_matrices}

\begin{subtable}[t]{0.48\textwidth}
    \caption{G1FixedBase\_D1\_AG\_DO\_v0}
    \centering
    \scalebox{1.0}{
    \begin{tabular}{c|ccccc}
    \toprule
    \textbf{Algorithm} & \textbf{SSA} & \textbf{PFM} & \textbf{CBF} & \textbf{SMA} & \textbf{SSS} \\
    \midrule
    \textbf{SSA} & 1.0000 & 0.6977 & 0.6279 & 1.0000 & 1.0000 \\
    \textbf{PFM} & 0.9677 & 1.0000 & 0.7742 & 0.9677 & 0.9677 \\
    \textbf{CBF} & 0.9643 & 0.8571 & 1.0000 & 0.9643 & 0.9643 \\
    \textbf{SMA} & 0.9149 & 0.6383 & 0.5745 & 1.0000 & 0.9149 \\
    \textbf{SSS} & 0.9773 & 0.6818 & 0.6136 & 0.9773 & 1.0000 \\
    \bottomrule
    \end{tabular}
    }
\end{subtable}
\hfill
\begin{subtable}[t]{0.48\textwidth}
    \caption{G1FixedBase\_D1\_AG\_DO\_v1}
    \centering
    \scalebox{1.0}{
    \begin{tabular}{c|ccccc}
    \toprule
    \textbf{Algorithm} & \textbf{SSA} & \textbf{PFM} & \textbf{CBF} & \textbf{SMA} & \textbf{SSS} \\
    \midrule
    \textbf{SSA} & 1.0000 & 0.5000 & 0.5250 & 0.9500 & 0.9500 \\
    \textbf{PFM} & 1.0000 & 1.0000 & 0.7500 & 0.9500 & 1.0000 \\
    \textbf{CBF} & 0.9545 & 0.6818 & 1.0000 & 0.9545 & 0.9545 \\
    \textbf{SMA} & 0.8085 & 0.4043 & 0.4468 & 1.0000 & 0.7872 \\
    \textbf{SSS} & 0.9744 & 0.5128 & 0.5385 & 0.9487 & 1.0000 \\
    \bottomrule
    \end{tabular}
    }
\end{subtable}

\vspace{0.5cm}

\begin{subtable}[t]{0.48\textwidth}
    \caption{G1FixedBase\_D1\_AG\_SO\_v0}
    \centering
    \scalebox{1.0}{
    \begin{tabular}{c|ccccc}
    \toprule
    \textbf{Algorithm} & \textbf{SSA} & \textbf{PFM} & \textbf{CBF} & \textbf{SMA} & \textbf{SSS} \\
    \midrule
    \textbf{SSA} & 1.0000 & 0.6200 & 1.0000 & 1.0000 & 1.0000 \\
    \textbf{PFM} & 1.0000 & 1.0000 & 1.0000 & 1.0000 & 1.0000 \\
    \textbf{CBF} & 1.0000 & 0.6200 & 1.0000 & 1.0000 & 1.0000 \\
    \textbf{SMA} & 1.0000 & 0.6200 & 1.0000 & 1.0000 & 1.0000 \\
    \textbf{SSS} & 1.0000 & 0.6200 & 1.0000 & 1.0000 & 1.0000 \\
    \bottomrule
    \end{tabular}
    }
\end{subtable}
\hfill
\begin{subtable}[t]{0.48\textwidth}
    \caption{G1FixedBase\_D1\_AG\_SO\_v1}
    \centering
    \scalebox{1.0}{
    \begin{tabular}{c|ccccc}
    \toprule
    \textbf{Algorithm} & \textbf{SSA} & \textbf{PFM} & \textbf{CBF} & \textbf{SMA} & \textbf{SSS} \\
    \midrule
    \textbf{SSA} & 1.0000 & 0.4419 & 0.7907 & 0.9767 & 1.0000 \\
    \textbf{PFM} & 1.0000 & 1.0000 & 0.9474 & 1.0000 & 1.0000 \\
    \textbf{CBF} & 1.0000 & 0.5294 & 1.0000 & 0.9706 & 1.0000 \\
    \textbf{SMA} & 0.8750 & 0.3958 & 0.6875 & 1.0000 & 0.8958 \\
    \textbf{SSS} & 0.9556 & 0.4222 & 0.7556 & 0.9556 & 1.0000 \\
    \bottomrule
    \end{tabular}
    }
\end{subtable}
\vspace{0.5cm}

\begin{subtable}[t]{0.48\textwidth}
    \caption{G1MobileBase\_D1\_WG\_DO\_v0}
    \centering
    \scalebox{1.0}{
    \begin{tabular}{c|ccccc}
    \toprule
    \textbf{Algorithm} & \textbf{SSA} & \textbf{PFM} & \textbf{CBF} & \textbf{SMA} & \textbf{SSS} \\
    \midrule
    \textbf{SSA} & 1.0000 & 0.9592 & 0.9184 & 1.0000 & 1.0000 \\
    \textbf{PFM} & 0.9792 & 1.0000 & 0.9167 & 0.9792 & 1.0000 \\
    \textbf{CBF} & 1.0000 & 0.9778 & 1.0000 & 1.0000 & 1.0000 \\
    \textbf{SMA} & 1.0000 & 0.9592 & 0.9184 & 1.0000 & 1.0000 \\
    \textbf{SSS} & 0.9800 & 0.9600 & 0.9000 & 0.9800 & 1.0000 \\
    \bottomrule
    \end{tabular}
    }
\end{subtable}
\hfill
\begin{subtable}[t]{0.48\textwidth}
 \caption{G1MobileBase\_D1\_WG\_DO\_v1}
    \centering
    \scalebox{1.0}{
    \begin{tabular}{c|ccccc}
    \toprule
    \textbf{Algorithm} & \textbf{SSA} & \textbf{PFM} & \textbf{CBF} & \textbf{SMA} & \textbf{SSS} \\
    \midrule
    \textbf{SSA} & 1.0000 & 0.6250 & 0.9750 & 0.5000 & 0.9750 \\
    \textbf{PFM} & 0.8621 & 1.0000 & 0.8966 & 0.5862 & 0.8966 \\
    \textbf{CBF} & 0.8864 & 0.5909 & 1.0000 & 0.4773 & 0.9773 \\
    \textbf{SMA} & 0.8696 & 0.7391 & 0.9130 & 1.0000 & 0.9130 \\
    \textbf{SSS} & 0.8667 & 0.5778 & 0.9556 & 0.4667 & 1.0000 \\
    \bottomrule
    \end{tabular}
    }
\end{subtable}

\vspace{0.5cm}

\begin{subtable}[t]{0.48\textwidth}
    \caption{G1MobileBase\_D1\_WG\_SO\_v0}
    \centering
    \scalebox{1.0}{
    \begin{tabular}{c|ccccc}
    \toprule
    \textbf{Algorithm} & \textbf{SSA} & \textbf{PFM} & \textbf{CBF} & \textbf{SMA} & \textbf{SSS} \\
    \midrule
    \textbf{SSA} & 1.0000 & 0.9592 & 1.0000 & 0.9796 & 1.0000 \\
    \textbf{PFM} & 1.0000 & 1.0000 & 1.0000 & 1.0000 & 1.0000 \\
    \textbf{CBF} & 0.9800 & 0.9400 & 1.0000 & 0.9600 & 1.0000 \\
    \textbf{SMA} & 1.0000 & 0.9792 & 1.0000 & 1.0000 & 1.0000 \\
    \textbf{SSS} & 0.9800 & 0.9400 & 1.0000 & 0.9600 & 1.0000 \\
    \bottomrule
    \end{tabular}
    }
\end{subtable}
\hfill
\begin{subtable}[t]{0.48\textwidth}
    \caption{G1MobileBase\_D1\_WG\_SO\_v1}
    \centering
    \scalebox{1.0}{
    \begin{tabular}{c|ccccc}
    \toprule
    \textbf{Algorithm} & \textbf{SSA} & \textbf{PFM} & \textbf{CBF} & \textbf{SMA} & \textbf{SSS} \\
    \midrule
    \textbf{SSA} & 1.0000 & 0.5957 & 0.9574 & 0.6596 & 0.9574 \\
    \textbf{PFM} & 0.9333 & 1.0000 & 0.9333 & 0.8000 & 0.9333 \\
    \textbf{CBF} & 0.9783 & 0.6087 & 1.0000 & 0.6739 & 1.0000 \\
    \textbf{SMA} & 1.0000 & 0.7742 & 1.0000 & 1.0000 & 1.0000 \\
    \textbf{SSS} & 0.9783 & 0.6087 & 1.0000 & 0.6739 & 1.0000 \\
    \bottomrule
    \end{tabular}
    }
\end{subtable}

\end{table}

%% file: sections/conditional_success_matrix.tex
\definecolor{myPurple}{rgb}{1.0, 0.0, 0.0}
\def\dataA{{
{1.0000, 0.5000, 0.5250, 0.9500, 0.9500},
{1.0000, 1.0000, 0.7500, 0.9500, 1.0000},
{0.9545, 0.6818, 1.0000, 0.9545, 0.9545},
{0.8085, 0.4043, 0.4468, 1.0000, 0.7872},
{0.9744, 0.5128, 0.5385, 0.9487, 1.0000}
}}

\def\dataB{{
{1.0000, 0.6977, 0.6279, 1.0000, 1.0000},
{0.9677, 1.0000, 0.7742, 0.9677, 0.9677},
{0.9643, 0.8571, 1.0000, 0.9643, 0.9643},
{0.9149, 0.6383, 0.5745, 1.0000, 0.9149},
{0.9773, 0.6818, 0.6136, 0.9773, 1.0000}
}}

\def\dataC{{
{1.0000, 0.4419, 0.7907, 0.9767, 1.0000},
{1.0000, 1.0000, 0.9474, 1.0000, 1.0000},
{1.0000, 0.5294, 1.0000, 0.9706, 1.0000},
{0.8750, 0.3958, 0.6875, 1.0000, 0.8958},
{0.9556, 0.4222, 0.7556, 0.9556, 1.0000}
}}

\def\dataD{{
{1.0000, 0.6200, 1.0000, 1.0000, 1.0000},
{1.0000, 1.0000, 1.0000, 1.0000, 1.0000},
{1.0000, 0.6200, 1.0000, 1.0000, 1.0000},
{1.0000, 0.6200, 1.0000, 1.0000, 1.0000},
{1.0000, 0.6200, 1.0000, 1.0000, 1.0000}
}}

\def\dataE{{
{1.0000, 0.6250, 0.9750, 0.5000, 0.9750},
{0.8621, 1.0000, 0.8966, 0.5862, 0.8966},
{0.8864, 0.5909, 1.0000, 0.4773, 0.9773},
{0.8696, 0.7391, 0.9130, 1.0000, 0.9130},
{0.8667, 0.5778, 0.9556, 0.4667, 1.0000}
}}

\def\dataF{{
{1.0000, 0.9592, 0.9184, 1.0000, 1.0000},
{0.9792, 1.0000, 0.9167, 0.9792, 1.0000},
{1.0000, 0.9778, 1.0000, 1.0000, 1.0000},
{1.0000, 0.9592, 0.9184, 1.0000, 1.0000},
{0.9800, 0.9600, 0.9000, 0.9800, 1.0000}
}}

\def\dataG{{
{1.0000, 0.5957, 0.9574, 0.6596, 0.9574},
{0.9333, 1.0000, 0.9333, 0.8000, 0.9333},
{0.9783, 0.6087, 1.0000, 0.6739, 1.0000},
{1.0000, 0.7742, 1.0000, 1.0000, 1.0000},
{0.9783, 0.6087, 1.0000, 0.6739, 1.0000}
}}

\def\dataH{{
{1.0000, 0.9592, 1.0000, 0.9796, 1.0000},
{1.0000, 1.0000, 1.0000, 1.0000, 1.0000},
{0.9800, 0.9400, 1.0000, 0.9600, 1.0000},
{1.0000, 0.9792, 1.0000, 1.0000, 1.0000},
{0.9800, 0.9400, 1.0000, 0.9600, 1.0000}
}}

\begin{figure*}[htbp]
    \centering
    \vspace{2cm}  
    \begin{tikzpicture}[transform canvas={xshift=-4.5cm}] 
        \def\imgwidth{4.5cm}  
        \def\xgap{4.5}  
        \def\ygap{-5.2} 
        \def\xshift{-2} 

        \node at (\xshift + 0, 0) {\drawHeatMap{\dataB}};
        \node[below] at (\xshift + 0, -2.4) {\small (a) G1FixedBase\_D1\_AG\_DO\_v0};

        \node at (\xshift + \xgap, 0) {\drawHeatMap{\dataA}};
        \node[below] at (\xshift + \xgap, -2.4) {\small (b) G1FixedBase\_D1\_AG\_DO\_v1};

        \node at (\xshift + 2*\xgap, 0) {\drawHeatMap{\dataD}};
        \node[below] at (\xshift + 2*\xgap, -2.4) {\small (c) G1FixedBase\_D1\_AG\_SO\_v0};

        \node at (\xshift + 3*\xgap, 0) {\drawHeatMap{\dataC}};
        \node[below] at (\xshift + 3*\xgap, -2.4) {\small (d) G1FixedBase\_D1\_AG\_SO\_v1};

        \node at (\xshift + 0, \ygap) {\drawHeatMap{\dataF}};
        \node[below, align=center, font=\small] at (\xshift + 0, \ygap-2.4) {(e) G1MobileBase\_D1\\\_WG\_DO\_v0};

        \node at (\xshift + \xgap, \ygap) {\drawHeatMap{\dataE}};
        \node[below, align=center, font=\small] at (\xshift + \xgap, \ygap-2.4) {(f) G1MobileBase\_D1\\\_WG\_DO\_v1};

        \node at (\xshift + 2*\xgap, \ygap) {\drawHeatMap{\dataH}};
        \node[below, align=center, font=\small] at (\xshift + 2*\xgap, \ygap-2.4) {(g) G1MobileBase\_D1\\\_WG\_SO\_v0};

        \node at (\xshift + 3*\xgap, \ygap) {\drawHeatMap{\dataG}};
        \node[below, align=center, font=\small] at (\xshift + 3*\xgap, \ygap-2.4) {(h) G1MobileBase\_D1\\\_WG\_SO\_v1};
        
        \def\barWidth{8.0}
        \def\barHeight{0.5}
        \def\steps{50}
    
        \foreach \s in {0,...,\steps} {
            \pgfmathsetmacro{\f}{\s/\steps}  
            \pgfmathparse{100*\f}            
            \let\colorValue=\pgfmathresult
    
            \pgfmathsetmacro{\xpos}{\f*\barWidth}
    
            \fill[myPurple!\colorValue!yellow,draw=none]
                 (\xpos, 1.75*\ygap) rectangle
                 (\xpos + \barWidth/\steps, \barHeight+ 1.75*\ygap);
        }
    
        \draw[black] (0, 1.75*\ygap) rectangle (\barWidth+0.15,\barHeight+ 1.75*\ygap);
    
        \node[below] at (0, 1.75*\ygap) {\footnotesize 0.85};
        \node[below] at (\barWidth, 1.75*\ygap) {\footnotesize 1.0};
    \end{tikzpicture}
    \vspace{9.5cm} 
    
    \caption{Conditional success plots for different tasks. The value at cell $(i, j)$ represents the proportion of environment settings successfully completed by the $i$-th algorithm that are also successfully completed by the $j$-th algorithm.}
    \label{fig: conditional_success}
\end{figure*}